\documentclass[10pt]{article} 
\usepackage[accepted]{tmlr}

\usepackage{hyperref}
\usepackage{url}

\usepackage{amsmath}
\usepackage{amssymb}
\usepackage{mathtools}
\usepackage{amsthm}

\usepackage{algorithm}
\usepackage{algpseudocode}

\newtheorem{thm}{Theorem}[]
\newtheorem*{thm*}{Theorem}
\newtheorem{m-thm}[thm]{Meta-Theorem}
\newtheorem*{m-thm*}{Meta-Theorem}

\newtheorem{remark}{Remark}[]

\newtheorem{prop}[thm]{Proposition}
\newtheorem*{prop*}{Proposition}
\newtheorem{Definition}{Definition}

\newtheorem{Corollary}[thm]{Corollary}

\newtheorem{Example}[thm]{Example}

\newtheorem{algor}[thm]{Method}

\newtheorem{Condition}[thm]{Condition}

\newtheorem{sett}[]{Setting}

\renewcommand{\phi}{\varphi}

\newcommand{\X}{\mathcal{X}}

\newcommand{\E}[2][]{\mathbb{E}_{#1}\!\left[#2\right]}
\renewcommand{\P}[1]{\mathbb{P}\!\left(#1\right)}

\newcommand{\R}{\mathbb{R}}

\newcommand{\ve}{\epsilon}

\renewcommand{\S}{\mathcal{S}}

\usepackage{bm,bbm}
\usepackage{nicematrix}
\usepackage{comment}

\newcommand{\mask}{\text{\ttfamily[MASK]}}
\usepackage{booktabs}
\newcommand{\argtop}[1]{\operatorname{argtop\hspace{0.25\fontdimen2\font}{\it#1}}}
\usepackage{enumitem}
\usepackage{subfigure}
\usepackage{wrapfig}
\usepackage{float}
\usepackage{multirow}

\newcommand{\red}[1]{{#1}}
\newcommand{\blue}[1]{{#1}}
\newcommand{\rev}[1]{{#1}}
\newcommand{\bt}[1]{{\bf\texttt{#1}}}

\newcommand{\mtabcaption}[2]{
    \caption{#2}
}
\newcommand{\mfigcaption}[2]{
    \caption{#2}
}

\newcommand{\nfigcaption}[2]{
    \caption{#2}
}

\usepackage{tcolorbox}
\newcommand{\eot}{\textless\textbar endoftext\textbar\textgreater}
\usepackage{listings}
\usepackage{lstlinebgrd}
\lstdefinestyle{samplerpython}{%
    language=Python,
    basicstyle=\ttfamily\footnotesize,
    numbers=left,
    numberstyle=\tiny\color{black},
    numbersep=6pt,
    keepspaces=true,
    showstringspaces=false,
    columns=fullflexible,
    keywordstyle=\color{green!50!black}\bfseries,
    commentstyle=\color{teal!70!black}\itshape,
    stringstyle=\color{red!75!black}\itshape,
}
\newcommand{\maskgitlinebackground}{%
    \ifnum\value{lstnumber}>27\relax
        \ifnum\value{lstnumber}<40\relax\color{cyan!18}\fi
    \fi
    \ifnum\value{lstnumber}>48\relax
        \ifnum\value{lstnumber}<55\relax\color{cyan!18}\fi
    \fi
}
\newcommand{\momentlinebackground}{%
    \ifnum\value{lstnumber}>7\relax
        \ifnum\value{lstnumber}<11\relax\color{cyan!18}\fi
    \fi
    \ifnum\value{lstnumber}>26\relax
        \ifnum\value{lstnumber}<35\relax\color{cyan!18}\fi
    \fi
    \ifnum\value{lstnumber}>43\relax
        \ifnum\value{lstnumber}<57\relax\color{cyan!18}\fi
    \fi
}
\usepackage{csquotes}

\title{Demystifying MaskGIT Sampler and Beyond:\\ Adaptive Order Selection in Masked Diffusion}

\author{\name Satoshi Hayakawa
\email hayakawa@mist.i.u-tokyo.ac.jp \\
\addr Sony Group Corporation
\AND
\name Yuhta Takida
\email yuta.takida@sony.com \\
\addr Sony AI
\AND
\name Masaaki Imaizumi
\email imaizumi@g.ecc.u-tokyo.ac.jp \\
\addr The University of Tokyo
\AND
\name Hiromi Wakaki
\email hiromi.wakaki@sony.com \\
\addr Sony Group Corporation
\AND
\name Yuki Mitsufuji
\email yuhki.mitsufuji@sony.com \\
\addr Sony AI, Sony Group Corporation
}

\def\month{04}  
\def\year{2026} 
\def\openreview{\url{https://openreview.net/forum?id=mKlW68i2Ig}} 

\renewcommand{\S}{\mathcal{S}}
\renewcommand{\mask}{\texttt{M}}

\begin{document}

\maketitle

\begin{abstract}
Masked diffusion models have shown promising performance in generating high-quality samples in a wide range of domains,
but accelerating their sampling process remains relatively underexplored.
To investigate efficient samplers for masked diffusion, this paper theoretically analyzes
the MaskGIT sampler for image modeling,
revealing its implicit temperature sampling mechanism.
\blue{Through this analysis, we show that MaskGIT is asymptotically equivalent to a {\it choose-then-sample} (CTS) formulation,
instantiated as the \enquote{moment sampler,}
which explicitly separates index selection from token sampling.
}
\blue{This CTS reformulation is essential: it yields unbiased token sampling and exposes an algorithmic design space for index selection,
both of which are inaccessible in MaskGIT’s original formulation.}
\blue{Regarding token sampling, we reveal that MaskGIT implicitly adopts a low-temperature sampler, which explains why MaskGIT often degrades with more sampling steps. The CTS reformulation of MaskGIT allows us to fix the temperature sampling to ensure unbiasedness.}
\blue{We also improve the index selection in CTS} through two key innovations: a partial caching technique for transformers that approximates longer sampling trajectories without proportional computational cost, and a hybrid approach formalizing the exploration-exploitation trade-off in adaptive unmasking.
Experiments in image and text domains demonstrate our theory as well as the efficiency of our proposed methods,
advancing both theoretical understanding and practical implementation of masked diffusion samplers.
\end{abstract}

\section{Introduction}
Generative models have witnessed remarkable progress in recent years, with diffusion models \citep{sohl2015deep} emerging as a dominant paradigm across various domains including images \citep{ho2020denoising,dhariwal2021diffusion}, audio \citep{kong2021diffwave} and video \citep{ho2022video}.
While continuous diffusion models have garnered significant attention, discrete diffusion models
\citep{austin2021structured,campbell2022continuous,gu2022vector,loudiscrete}
offer compelling advantages for inherently discrete data, such as tokenized representations in all the above domains.

Among discrete diffusion approaches, masked diffusion models \citep{sahoo2024simple,shi2024simplified,ou2024your} have demonstrated exceptional performance, particularly in generating high-quality samples in language domains.
However, their sampling process remains computationally intensive, requiring hundreds of function evaluations.
Recent efforts to accelerate sampling in discrete/masked diffusion models have explored various directions, including optimized scheduling \citep{jys}, distillation \citep{deschenaux2024beyond,hayakawa2025distillation,zhu2025di},
and adaptive token selection \citep{kim2025train,ben2025accelerated}.

One of the most notable masked diffusion samplers is given by MaskGIT \citep{chang2022maskgit}, which has shown that impressive image generation can be achieved with as few as 8-12 sampling steps, significantly fewer than the hundreds typically required by discrete diffusion models.
This efficiency has inspired subsequent work in image \citep{lezama2022discrete,besnier2025halton},
audio \citep{garcia2023vampnet,comunita2024specmaskgit}, 
and language domains \citep{zhengreparameterized,nie2025large}.
However, the method is heuristic and not well understood in theory,
and it often shows degraded performance when increasing sampling steps \rev{in image generation} (\citealp[e.g.,][]{gat2024discrete}; this paper's Figure~\ref{fig:mage})
\red{or severely reduced diversity in language generation (Figure~\ref{fig:sdtt}).}
\red{Moreover, MaskGIT's sampling process entangles its token sampling and index selection,
thereby limiting its compatibility with various existing index-selection criteria
\citep{xiang2022mimt,zhengreparameterized,besnier2025halton,kim2025train}.
In summary, although the MaskGIT sampler has a promising performance in image generation,
there are two problems to solve in the MaskGIT sampler, especially when applying it to language and other domains:
(i) reduced diversity in many-step and/or low-temperature sampling, and (ii)~limited flexibility of the sampler to incorporate other index-selection criteria.
}

\begin{figure}[t]
    \centering
    \includegraphics[width=\linewidth]{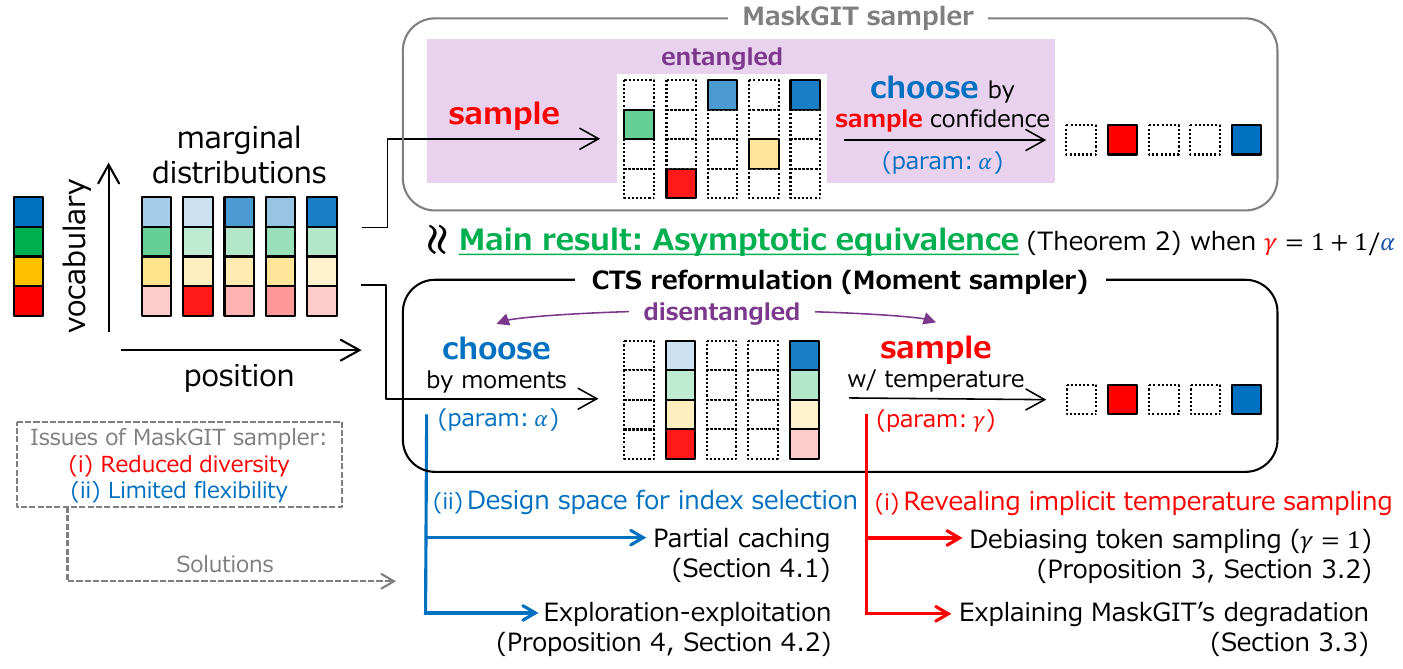}
    \vspace{-6mm}
    \caption{Overview of our contributions.
    \blue{We show that the MaskGIT sampler is asymptotically equivalent to a choose-then-sample (CTS) formulation (Theorem~\ref{thm:main}),
    instantiated as the moment sampler, which decouples index selection and token sampling.
    This CTS reformulation yields an unbiased sampler when temperature is removed (Proposition~\ref{prop:one-by-one-cts})
    and provides a principled explanation for MaskGIT’s degradation at large step counts (Section~\ref{sec:theory}).
    Moreover, it opens an algorithmic design space for index selection that enables partial caching and hybrid bi-criteria methods
    (Section~\ref{sec:method}).}
    }
    \label{fig:paper}
\end{figure}

\red{In this paper, we address these problems through a theoretical analysis of MaskGIT.
We show that MaskGIT is asymptotically equivalent to a {\it choose-then-sample} (CTS) formulation
(instantiated as ``moment sampler''),
which explicitly separates index selection from token sampling.
This reformulation allows us to address problems (i) and (ii) independently:
for (i), we reveal that MaskGIT implicitly performs temperature-based token sampling, which accounts for the diversity degradation, and the CTS formulation enables unbiased sampling by simply removing the temperature;
for (ii), the decoupled structure naturally accommodates various index-selection criteria that were incompatible with MaskGIT's entangled process
(see Figure~\ref{fig:paper}).}


For practical improvement of masked diffusion samplers \red{concerning point (ii)},
we then introduce two key techniques
specifically tailored for choose-then-sample methods.
First, we propose a partial caching technique for transformer-based models that effectively approximates the sampling trajectories with more steps without proportionally increasing computational cost, unlike the MaskGIT sampler that requires recomputation of all positions at each step.
Second, we formalize the exploration-exploitation trade-off in adaptive unmasking of masked diffusion sampling, leading to a hybrid approach that combines the strengths of exploitation-focused methods \citep{kim2025train,ben2025accelerated} with exploration-oriented techniques such as Halton scheduling \citep{besnier2025halton}.

Our contributions are illustrated in Figure~\ref{fig:paper}.
They can be summarized as follows:

\begin{itemize}
    \item In Section~\ref{sec:theory}, we show that MaskGIT implicitly performs temperature sampling
    \blue{and is asymptotically equivalent to a CTS formulation instantiated as the moment sampler
    (Theorem~\ref{thm:main}),
    providing a precise decomposition into index selection and token sampling.
    }
    \blue{The CTS reformulation yields an insight that MaskGIT's degradation at large step counts
    can be attributed to its implicit temperature sampling; removing temperature leads to unbiased sampling
    in the many-step limit (Proposition~\ref{prop:one-by-one-cts}).}
    \item In Section~\ref{sec:method},
    \blue{leveraging the CTS formulation,
    we introduce two algorithmic improvements in index selection that are not expressible
    in MaskGIT's original sample-then-choose form.
    }
    One is a partial caching technique for transformer-based models that approximates more steps without proportionally increasing computational cost. The other is through our formalization of the exploration-exploitation trade-off in masked diffusion sampling, developing a hybrid approach that balances these competing objectives.
    \item In Section~\ref{sec:experiments},
    \blue{experiments on image and language generation
    validate the theoretical equivalence, isolate the dominant role of temperature,
    and demonstrate improved efficiency-quality trade-offs for CTS-based samplers.}
\end{itemize}

Our work not only advances the theoretical understanding of masked diffusion samplers but also provides practical techniques to enhance their efficiency, toward better modeling for discrete tokens.

\section{Preliminaries}
Let $q_\mathrm{data}$ be the data distribution over the product space $\X=\S^D$,
where $\S$ is a finite set of tokens and $D$ is the dimensionality/length of the data.
Each state $\bm{x} \in \mathcal{X}$ can be represented as a sequence of tokens
$\bm{x}=(x_i)_{i=1}^D$ with $x_i \in \S$ for each $i\in[D]$, where $[D]:=\{1,\ldots,D\}$.
For any set of indices $I\subset [D]$, let us write $\bm{x}_I=(x_i)_{i\in I}$.
For $I,J\subset [D]$,
let $q_{I|J}(\bm{x}_I|\bm{x}_J)=\P{\bm{y}_I=\bm{x}_I\mid \bm{y}_J=\bm{x}_J}$
with $\bm{y}\sim q_\mathrm{data}$.
Let us also slightly abuse the notation to 
represent the case $I=\{i\}$ to write
$q_{i|J}(x_i|\bm{x}_J)$.
Finally, $\argtop{k}_{i\in I}\{a_i\}$ is the top-$k$ indices of a sequence $(a_i)_{i\in I}$
rearranged in descending order.

\subsection{Masked diffusion models}\label{sec:mdm}
In masked diffusion models, we augment the vocabulary with a special mask token $\mask\not\in\S$. The forward process $\bm{x}(t)\in (\S\cup\{\mask\})^D$
with $\S^D\ni \bm{x}(0)=\bm{x} \sim q_\mathrm{data}$ 
gradually replaces tokens with mask tokens according to a predefined schedule,
eventually reaching $\bm{x}(1) = (\mask,\ldots,\mask)$.
For generation,
we simulate its backward process starting from a fully masked sequence.
While the time-dependent formulation is a natural consequence of general discrete diffusion modeling,
recent studies \citep{ou2024your,zheng2024masked} pointed out that, under commonly used forward masking models,
the conditional probability $\P{\bm{x}(0) = \bm{z}\mid \bm{x}(t) = \bm{y}}$
does not depend on $t$.
Thus, for simplicity, we assume that a given pretrained model is time-independent
and tries to approximate the conditional distribution of
$q_{I|J}(\bm{x}_I|\bm{x}_J)$
for $I,J\subset[D]$, and $\bm{x}\sim q_\mathrm{data}$.
Then, our unmasking process is given by $\emptyset =J_0 \subset\cdots\subset J_n=[D]$
to iteratively sample $\bm{x}_{J_\ell\setminus J_{\ell-1}}$ approximately
from $q_{J_\ell\setminus J_{\ell-1}|J_{\ell-1}}(\cdot|\bm{x}_{J_{\ell-1}})$ for $\ell=1,\ldots,n$.
For efficient sampling, we prefer using smaller $n$, the number of unmasking steps,
which requires unmasking of multiple token positions at each step.

Since the probability distributions over $\S^{\lvert I\rvert}$ with large $\lvert\S\rvert$ or $\lvert I\rvert$ are intractable,
directly modeling $q_{I|J}$ for $I$ containing multiple token positions is inefficient.
A common workaround is the \textit{product modeling}
$p_{I|J}(\bm{x}_I|\bm{x}_J):=\prod_{i\in I}p_{i|J}(x_i|\bm{x}_J)$,
which only approximates one-token marginals as $p_{i|J}\approx q_{i|J}$.
In this paper, we investigate efficient samplers for such product models
without any retraining or additional components.

\subsection{MaskGIT sampler}
We review MaskGIT, which is a pioneering method of post-hoc efficient sampling for masked diffusion \citep{chang2022maskgit}.
To simplify the analysis, let us just consider {\it one step} of unmasking.
Suppose we are given $N$ probability distributions $p_1,\ldots, p_N$
(corresponding to $p_{i|J}$ for $i\not\in J$ in the previous section,
and so $N=D-\lvert J\rvert$).
We would like to unmask $k$ token positions from $[N]$.
Given the temperature parameter
\footnote{While not mentioned \citet{chang2022maskgit},
this ``Gumbel temperature'' $\alpha$
is used in their official implementation (\url{https://github.com/google-research/maskgit/}).
Its typical value ranges around $1.0$ to $10.0$
with additional step-dependent scheduling
\citep{besnier2023pytorch,li2023mage,comunita2024specmaskgit}.
}
$\alpha>0$,
the MaskGIT sampler \citep{chang2022maskgit} is given as follows:
\begin{enumerate}[label=(MG\arabic*)]
    \item Independently sample $x_i\sim p_i$
        and a standard Gumbel noise $\xi_i$ for each $i\in[N]$.
    \item \label{mg-step-2}
        Choose $(i_1,\ldots,i_k)=\argtop{k}_{i\in[N]}\{\log p_i(x_i)+\alpha\xi_i\}$.
    \item Return the indices $i_1,\ldots, i_k$ and samples $x_{i_1},\ldots,x_{i_k}$.
\end{enumerate}
We formalize this sampling scheme in Algorithm~\ref{algo:maskgit-one-round} in the appendix.
This sampler works surprisingly well in image modeling
when the number of sampling steps is small \citep{chang2022maskgit,li2023mage}.
However, it is also observed that its performance degrades as we increase the number of steps (\citealp[][]{gat2024discrete,ren2025fast}; see also Figure~\ref{fig:mage}).
In the following section, we investigate this algorithm and derive an asymptotically equivalent sampler that is more flexible and interpretable;
the analysis also partially explains the above performance decay.

\section{\blue{Choose-then-sample reformulation of MaskGIT sampler}}\label{sec:theory}
We analyze the MaskGIT sampler in this section.
Let us start from a general result on the Gumbel-top-$k$ sampling,
which is used in \ref{mg-step-2} of the MaskGIT sampler.
As a top-$k$ generalization of the Gumbel-max trick \citep{maddison2014sampling},
the following result is known:
\begin{prop}[Gumbel-top-$k$ trick, {\citealp[][Eq.~18]{kool2019stochastic}}]
\label{thm:gumbel-topk}
    Suppose we are given $\mu_1,\ldots,\mu_N\in\R$ and i.i.d.~standard Gumbel noise $\xi_1,\ldots,\xi_N$.
    Let $(i^*_1,\ldots,i^*_k) = \argtop{k}_{i\in[N]}\{\mu_i+\xi_i\}$.
    Then, for distinct indices $i_1,\ldots,i_\ell \in [N]$ with $\ell\le k$, we have
    $
        \P{i^*_\ell = i_\ell \mid i^*_1=i_1,\ldots,i^*_{\ell-1}=i_{\ell-1}}
        ={\exp(\mu_{i_\ell})}
        /{\sum_{i\in [N]\setminus I_{\ell-1}}\exp(\mu_i)},
    $
    where $I_{\ell-1}:=\{i_1,\ldots,i_{\ell-1}\}$.
\end{prop}
As mentioned in the original paper,
it is mathematically equivalent to the size-$k$ sampling without replacement with logits $\mu_i$.
While it also reveals the value of $\P{i^*_1=i_1,\ldots,i^*_{k}=i_{k}}$,
we are particularly interested in the conditional form in the proposition.

Let us consider the case $I=[N]$ for simplicity.
Let $(i^*_1,\ldots, i^*_k)=\argtop{k}_{i\in[N]}\{\log p_i(x_i)+\alpha\xi_i\}$
in the MaskGIT sampler (Algorithm~\ref{algo:maskgit-one-round}, line 3).
Since it is equivalent to consider the Gumbel-top-$k$ sampling for
$\mu_i = \alpha^{-1}\log p_i(x_i)$
from Proposition~\ref{thm:gumbel-topk}, we have
\begin{equation}
    \P{i_\ell^* = i_\ell \mid i_1^*=i_1,\ldots,i_{\ell-1}^*=i_{\ell-1}, (x_i)_{i=1}^N}
    = \frac{p_{i_\ell}(x_{i_\ell})^{1/\alpha}}{\sum_{i\in [N]\setminus I_{\ell-1}} p_i(x_i)^{1/\alpha}}
    \label{eq:maskgit-topk-eq}
\end{equation}
for each $\ell\le k$.
When $N - k$ is large (i.e., there are many masked positions),
the sum of independent terms $p_i(x_i)^{1/\alpha}$
should be approximated by its expectation because of probability concentration:
\begin{equation}
    \sum_{i\in [N]\setminus I_{\ell-1}}p_i(x_i)^{1/\alpha}
    \approx \sum_{i\in [N]\setminus I_{\ell-1}}\E[x_i\sim p_i]{p_i(x_i)^{1/\alpha}}
    =\sum_{i\in [N]\setminus I_{\ell-1}}\sum_{x\in\S}{p_i(x)^{1+1/\alpha}}.
    \label{eq:large-sample-approxiamtion}
\end{equation}
{To see its quantitative version, applying Bernstein's inequality (e.g., \citealt[Corollary 2.11]{boucheron2013concentration}) to the independent random variables $p_i(x_i)^{1/\alpha}\in [0, 1]$ yields,
for $t>0$,
\begin{equation*}
    \P{\left\lvert\sum_{i\in [N]\setminus I_{\ell-1}} p_i(x_i)^{1/\alpha}-\sum_{i\in [N]\setminus I_{\ell-1}}\E{p_i(x_i)^{1/\alpha}}\right\rvert \ge t}
    \le 2\exp\left( -\frac{t^2}{2(\sigma^2 + t/3)}
    \right),
    \label{eq:bernstein-reb}
\end{equation*}
where $\sigma^2$ is the variance of $\sum_{i\in [N]\setminus I_{\ell-1}} p_i(x_i)^{1/\alpha}$.
Since $0\le p_i(x_i)\le 1$, it is upper-bounded as
$
    \sigma^2\le \sum_{i\in [N]\setminus I_{\ell-1}} \E{(p_i(x_i)^{1/\alpha})^2}
    \le \sum_{i\in [N]\setminus I_{\ell-1}} \E{p_i(x_i)^{1/\alpha}}.
$
By substituting this upper bound and letting $t=\ve \sum_{i\in [N]\setminus I_{\ell-1}} \E{p_i(x_i)^{1/\alpha}}$
for an $\epsilon>0$ in the Bernstein estimate,
we can quantify the concentration of \eqref{eq:large-sample-approxiamtion} as follows:
\begin{align*}
    \P{\left\lvert\frac{\sum_{i\in [N]\setminus I_{\ell-1}} p_i(x_i)^{1/\alpha}}{\sum_{i\in [N]\setminus I_{\ell-1}}\E{p_i(x_i)^{1/\alpha}}}-1\right\rvert \ge \epsilon}
    &\le 2\exp\biggl( -\frac{3\ve}{8} \sum_{i\in [N]\setminus I_{\ell-1}}\E{p_i(x_i)^{1/\alpha}}\biggr)\\
    &\le 2\exp\left(-\frac{3(N-k+1)}{8\lvert \S\rvert^{1/\alpha}}\ve\right),
\end{align*}
where the second inequality follows from $N-k+1\le N-(\ell-1)$ and the lower bound of $\E{p_i(x_i)^{1/\alpha}}$
using the vocabulary size $\lvert\S\rvert$, shown by \eqref{eq:app:lower-bound-moment} in the appendix.
Although the actual proof is more involved,
this concentration is at the heart of proving our main theorem (Theorem~\ref{thm:main}).

}

Since $\sum_{x\in\S}p_i(x)^{1+1/\alpha}$ is the {($1+1/\alpha$)}-th power of {($1+1/\alpha$)}-norm of $p_i$
when it is regarded as a vector in $\R^{\lvert\S\rvert}$,
we can simply write
$
    \sum_{i\in [N]\setminus I_{\ell-1}}p_i(x_i)^{1/\alpha}
    \approx \sum_{i\in [N]\setminus I_{\ell-1}} \lVert p_i\rVert_{{1+1/\alpha}}^{{1+1/\alpha}}.
$
Under this approximation (including the same approximation for smaller $\ell$), we can derive
\begin{equation}
    \P{x_{i_\ell}, i_\ell^* = i_\ell \mid i_1^*=i_1,\ldots,i_{\ell-1}^*=i_{\ell-1}}
    \approx \frac{p_{i_\ell}(x_{i_\ell})^{{1+1/\alpha}}}{\sum_{i\in[N]\setminus I_{\ell-1}}\lVert p_i\rVert^{{1+1/\alpha}}_{{1+1/\alpha}}},
    \label{eq:maskgit-cond-prob}
\end{equation}
where the right-hand side makes a probability distribution over $\S\times([N]\setminus I_{\ell-1})$.
The derivation of \eqref{eq:maskgit-cond-prob}
is deferred to Appendix~\ref{sec:derivation-moment-maskgit}.
From Proposition~\ref{thm:gumbel-topk},
choosing indices $i_1,\ldots,i_\ell$ with the right-hand side of \eqref{eq:maskgit-cond-prob}
(summed over $x_{i_\ell}\in\S$)
is equivalent to Gumbel-top-$k$ sampling with $\mu_i=\log\lVert p_i\rVert_{{1+1/\alpha}}^{{1+1/\alpha}}$.
Then, given the index $i$, the sampling distribution for $x_i$ is proportional to ${p_i^{{1+1/\alpha}}}$.

\subsection{Moment sampler}
Based on the above analysis,
the \textit{moment sampler},
our alternative sampler for approximating MaskGIT,
is formulated as follows:
\begin{enumerate}[label=(MM\arabic*)]
    \item\label{moment-step-1} Let $\smash{(i_1, \ldots, i_k) = \argtop{k}_{i\in[N]}
    \{\log\lVert p_i\rVert_{{1+1/\alpha}}^{{1+1/\alpha}} + \xi_i\}}$ with i.i.d~standard Gumbel $\smash{(\xi_i)_{i=1}^N}$.
    \item\label{moment-step-2} Independently sample
    {$x_i\sim \tilde{p}_{i}$ with ${\tilde{p}_i\propto p_i^{1+1/\alpha}}$} for each $i\in\{i_1,\ldots,i_k\}$.
    \item Return the indices $i_1,\ldots, i_k$ and samples $x_{i_1},\ldots,x_{i_k}$.
\end{enumerate}

{
Since the exponent $1+1/\alpha$ in \ref{moment-step-2} is a potential source of sampling error as we discuss later in this section,
we also consider the ``unbiased'' version of second step,
i.e., simply sampling from $p_i$.
To cover both of these options of sampling from $\propto p_i^\gamma$ with exponent $\gamma=1+1/\alpha$ and $1$, we formalize the moment sampler in Algorithm~\ref{algo:moment-one-round} with a general exponent $\gamma$ in \ref{moment-step-2}.
In the experiments in Section~\ref{sec:experiments}, $\gamma=1+1/\alpha$ defines \bt{Moment}, whereas the choice $\gamma=1$ provides \bt{U-Moment}.
See also Figures~\ref{fig:algo-maskgit} and~\ref{fig:algo-moment} in the appendix for comparing the MaskGIT and moment samplers in actual implementation using PyTorch.
}

Mathematically, we can prove the following:
\begin{thm}[{Moment sampler approximates MaskGIT in the $N\gg k^2$ regime}]\label{thm:main}
    Let $p_1,\ldots, p_N$ be probability distributions over a finite set $\S$.
    For $k\in [N]$ and $\alpha>0$,
    let $p_\mathrm{MaskGIT}$ be the output distribution of
    the corresponding
    MaskGIT sampler (Algorithm~\ref{algo:maskgit-one-round}),
    i.e., the distribution of $(i_1,\ldots, i_k)$ and $(x_{i_\ell})_{\ell=1}^k$ over $[N]^k\times\S^k$.
    Similarly, let $p_\mathrm{moment}$ be the output distribution of the moment sampler
    (Algorithm~\ref{algo:moment-one-round} with $\gamma=1+1/\alpha$).
    Then, we have
    \begin{equation}
        d_\mathrm{TV}({p}_\mathrm{moment}, {p}_\mathrm{MaskGIT})
        \le 5\sqrt{\frac{k^2\lvert\S\rvert^{1/\alpha}}{N}}\left(
            1 + \sqrt{
                    \log^+\!\left(\frac{N}{ k^2\lvert\S\rvert^{1/\alpha}}
                    \right)
                }
            \right),
        \nonumber
    \end{equation}
    where $d_\mathrm{TV}$ denotes the total variation distance
    and $\log^+(x):=\log(\max\{1, x\})$ for $x\in\R$.
\end{thm}
For a formal argument, see Appendix~\ref{sec:proof-main} and Theorem~\ref{thm:main-formal}.
Since $\lvert\S\rvert^{1/\alpha}$ is constant
and $N/k$ is approximately the number of steps,
it gets tighter when $\#(\text{steps})\gg \#(\text{unmasked tokens per step})$.
In this regime,
MaskGIT is approximated well by the moment sampler,
which samples from a different (or biased) distribution compared
to the true marginal because of the exponent ${{1+1/\alpha}}$ in \ref{moment-step-2}.
It thus partially explains the degraded behavior of the MaskGIT sampler as we increase the number of steps 
\blue{(see also Section~\ref{sec:cts-importance})}.
Moreover, even outside the above regime, 
we empirically observe that the moment sampler shows
fairly close behavior with MaskGIT in terms of evaluation metrics 
(Figures~\ref{fig:mage} and \ref{fig:sdtt}(Left)).

From Theorem~\ref{thm:main},
we can approximately decompose the MaskGIT sampler
into index selection and temperature sampling (Figure~\ref{fig:paper}).
This allows better understanding:
for instance, in our experiments,
temperature sampling is the dominant factor as
we can mostly replicate the performance of MaskGIT
even if we omit the index selection of moment sampler (Section~\ref{sec:experiments}).
Also, from a practical viewpoint,
we can improve the moment sampler
by techniques not applicable to MaskGIT (Section~\ref{sec:method}).

\subsection{Choose-then-sample algorithms}
Arguably the most notable difference of the moment sampler from MaskGIT
is that we choose the indices to unmask {\it before} sampling a token in each position.
Let us call this strategy \textit{choose-then-sample} (CTS) in general,
which we formalize and analyze below.


Given the set of currently unmasked indices $I$ and sample $\bm{x}_I$,
a step of CTS algorithms over $\S^D$ with inverse temperature $\gamma>0$
can be formalized as follows:
\begin{enumerate}[label=(CTS\arabic*)]
    \item\label{step:cts-1} Sample $J\subset[D]\setminus I$ for where to unmask.
        Its distribution is denoted as $\pi(\cdot|I, \bm{x}_I)$.
    \item\label{step:cts-2} $x_j\sim p_j$ with $p_j\propto p_{j|I}(\cdot|\bm{x}_I)^\gamma$ for each $j\in J$.
\end{enumerate}
We also present the whole algorithm as pseudocode in Algorithm~\ref{algo:choose-then-sample} in the appendix.
The unmasking position distribution $\pi$ is usually determined from
the information regarding $(p_{j|I}(\cdot|\bm{x}_I))_{j\in [D]\setminus I}$
such as entropy and probability margin \citep{xiang2022mimt,kim2025train}.

It is worth noting that, unlike the ``sample-then-choose'' MaskGIT,
CTS algorithms without temperature (i.e., $\gamma=1$)
return the correct distribution as we increase the number of steps,
provided that its marginals are accurate.
We formally prove this in Proposition~\ref{prop:one-by-one-cts}.
It is also mentioned by \citet[Section 5]{ben2025accelerated}
without a formal proof, and
we prove it for completeness in Appendix~\ref{sec:proof-cts}.
\begin{prop}[One-by-one CTS algorithm is unbiased]\label{prop:one-by-one-cts}
    In a CTS algorithm with $\gamma=1$, let us further assume that
    $J\sim\pi(\cdot|I, \bm{x}_I)$ in \ref{step:cts-1} is always a singleton set (i.e., $\lvert J\rvert=1$)
    and $p_{i|I} = q_{i|I}$ holds for each $I\subset[D]$ and $i\in[D]\setminus I$.
    Then, we have $\bm{x}\sim q_\text{\rm data}$ for the generated sample $\bm{x}$.
\end{prop}

\blue{
We next discuss the significance of CTS reformulation beyond providing an alternative to MaskGIT.

\subsection{Why CTS reformulation is essential}\label{sec:cts-importance}
Theorem~\ref{thm:main} establishes an asymptotic equivalence between MaskGIT and the moment sampler.
Importantly, this result goes beyond providing an alternative implementation: it converts MaskGIT’s sample-then-choose mechanism into an explicit CTS formulation, in which index selection and token sampling are decoupled.
This separation allows us to analyze correctness independently of ordering: in Proposition~\ref{prop:one-by-one-cts},
we show that CTS algorithms without temperature ($\gamma=1$) are unbiased in the one-by-one extreme, provided that one-token marginals are accurate.
This observation is difficult to express in MaskGIT’s original formulation, where token sampling and index selection are intrinsically entangled.

The CTS view also provides a principled explanation for MaskGIT’s degradation as the number of steps increases
(\citealp[][]{gat2024discrete,ren2025fast}).
Under the moment-sampler approximation, MaskGIT virtually samples from distributions proportional to $p_i^{1+1/\alpha}$ rather than the estimated marginal $p_i$, introducing a persistent bias that accumulates across refinement steps.
This interpretation isolates temperature as the dominant factor governing MaskGIT behavior, which we verify empirically in Section~\ref{sec:experiments}.

Beyond these theoretical implications, the CTS reformulation opens an algorithmic design space that is inaccessible in the original MaskGIT formulation.
Because index selection is explicitly decoupled from token sampling,
one can introduce ordering-level improvements without interfering with token sampling,
such as the partial caching approximation (Section~\ref{sec:caching})
and the hybrid ordering strategy (Section~\ref{sec:explo}).
}

\section{Towards more efficient choose-then-sample algorithms}\label{sec:method}
One benefit of CTS algorithms pointed out by \citet{zheng2024masked} is that
we can avoid sampling from $N$ categorical distributions
over $\S$, which can be costly if $N$ and $\lvert\S\rvert$ are large.
In this section,
we further provide two techniques for enhancing CTS algorithms.

\subsection{Partial caching approximation with transformers}\label{sec:caching}

\begin{figure}[htbp]
    \centering
    \includegraphics[width=\linewidth]{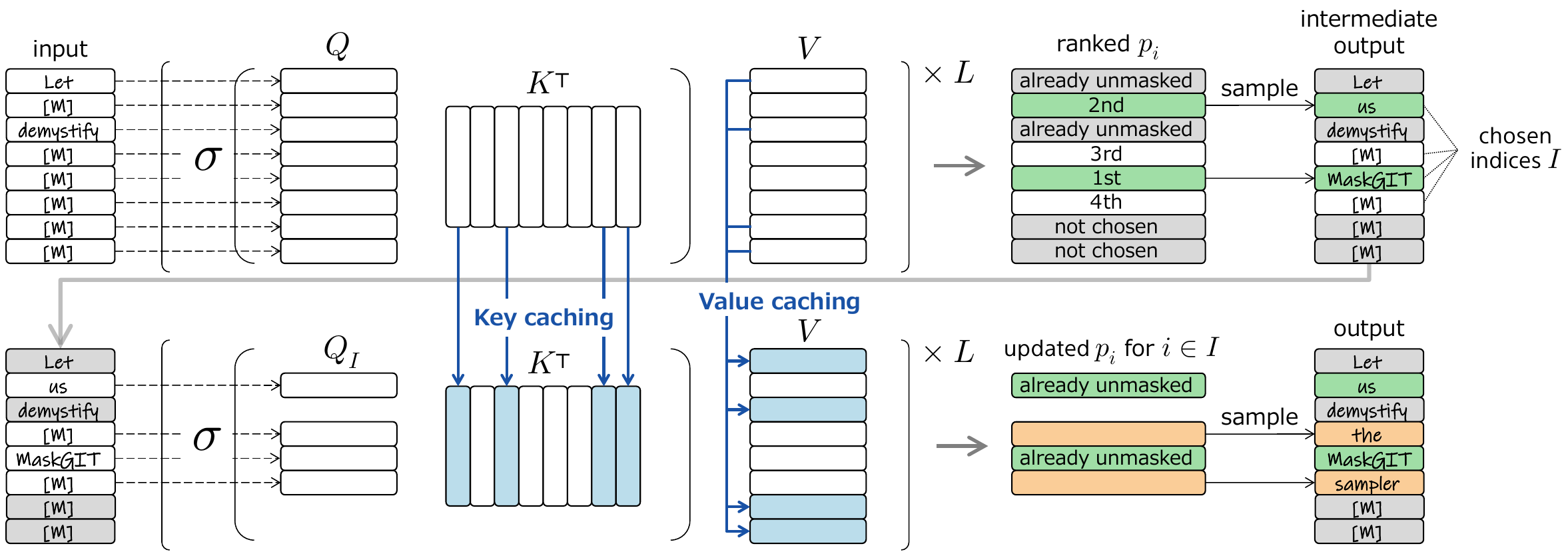}
    \vspace{-6mm}
    \caption{Illustration of partial caching approximation
    applied to an $L$-layer transformer,
where
$\sigma=\mathop\mathrm{softmax}(\cdot/\sqrt{d_k})$,
with $d_k$ being the dimension of key and query vectors.}
    \label{fig:caching}
\end{figure}
In masked diffusion modeling,
it is common to use a bidirectional transformer for computing logits at each position,
where the standard key-value (KV) caching for transformers with causal masking \citep{katharopoulos2020transformers}
is not applicable.
However, in CTS methods,
we can use a similar idea to KV-caching for virtually increasing the sampling steps
without proportionally increasing the cost.

Consider one CTS unmasking step.
Let $[D]$ be the set of all the indices and
$U\subset [D]$ be the set of already unmasked positions with values $\bm{x}_U$.
To compute $p_{i|U}(\cdot|\bm{x}_U)$
for $i\in [D]\setminus U$,
we input $D$ tokens to the transformer:
$x_i$ at each position $i\in U$ and $\mask$ at all the other positions.
In ordinary CTS methods, we then sample $I\sim \pi(U, \bm{x}_U)$,
the set of indices to unmask at this step,
where $\pi$ can depend on the inferred $(p_{i|U}(\cdot|\bm{x}_U))_{i\in[D]\setminus U}$.
Finally, we sample $x_i\sim p_{i|U}(\cdot|\bm{x}_U)$ for each $i\in I$.

Our {\it partial caching} method
can incorporate arguably finer-grained token dependencies within $I$.
First, we cache the key-value vectors as in the standard KV-caching.
Then, divide $I$ into two as $A,B\subset I$ with $A\cup B=I$ and $A\cap B=\emptyset$.
For $i\in A$, we just sample $x_i\sim p_{i|U}(\cdot|\bm{x}_U)$.
We {\it approximate} sampling from $p_{i|U\cup A}(\cdot|\bm{x}_{U\cup A})$
for remaining indices $i\in B$
with low cost as follows:
\begin{itemize}
    \item Run the transformer only at positions $i\in I$,
        with the input $x_i$ for $i\in A$ and $\mask$ for $i\in B$.
    \item As key-value vectors at positions $i\not\in I$, we use the cached ones.
\end{itemize}
It does not fully compute
the effect of input change (from $\mask$ to $x_i$ at each position $i\in A$)
over positions other than $A\cup B$.
However, it approximates the attention change for positions $i\in B$
that directly comes from $\bm{x}_A$, thus boosting the empirical performance
(Figure~\ref{fig:latency-perf}).
The second step operates with $\lvert I\rvert / D$
times computation of the original inference,
and the empirical performance gain outweighs its additional computation.
In practice, we can determine the partition $A\cup B=I$ by introducing an ordering.
For instance, we let $A$ as the first half of $I$ in the order of confidence or preference.
Figure~\ref{fig:caching} shows an example of such a two-stage sampling process.
See Section~\ref{sec:survey} for comparison with other caching techniques for discrete diffusion.

\subsection{Balancing exploration and exploitation}\label{sec:explo}
In this section, we decompose the error of a CTS algorithm
and see there are two types of existing algorithms depending on which part of the error to optimize.
\blue{Crucially, the CTS formulation makes such a decomposition possible at the level of index selection:
since ordering is explicit, we can reason about how the choice of $I$ trades off different sources of error without entangling it with token sampling.
Motivated by the error bound, we design a hybrid (bi-criteria) ordering that explicitly balances terms corresponding to exploitation and exploration.}

\blue{\begin{prop}[KL decomposition for two-round CTS]\label{prop:hybrid}
Consider a two-round CTS procedure on the index set $[N]$ with the ground-truth distribution $q$.
In the first round, a size-$k$ subset $I\subset [N]$ is chosen,
and tokens at $i\in I$ are generated as $x_i\sim q_i$, independently to each other.
Then, the remaining tokens are generated independently as $x_j\sim q_{j|I}(\cdot|\bm{x}_I)$ for $j\in [N]\setminus I$.
Let $p$ be the resulting distribution of this two-round procedure.
Then, we have
\begin{align}
    &D_\mathrm{KL}(q\,\Vert\,p)=
    D_\mathrm{KL}\Biggl(q_I\,\Bigg\Vert\,\prod_{i\in I}q_i\Biggr) + \E[\bm{x}_I\sim q_I]{
        D_\mathrm{KL}\Biggl(q_{I^c|I}(\cdot|\bm{x}_I)\, \Bigg\Vert \,
        \prod_{i\not\in I} q_{i|I}(\cdot|\bm{x}_I)\Biggr)} \nonumber\\
    &\le
    \underbrace{\sum_{i\in I}H(q_i)}_{\text{(a) exploitation}} -
    \underbrace{\sum_{i\in I}
    \E[J\sim \phi(\cdot|I\setminus\{i\}), \bm{x}_J\sim q_J]{H(q_{i|J}(\cdot|\bm{x}_J))}}_{\text{(b) spatial dispersion}}
    + \underbrace{\E[\bm{x}_I\sim q_I]{\sum_{i\in[N]\setminus I} H(q_{i|I}(\cdot|\bm{x}_I))}}_{
        \text{(c) exploration}},
    \label{eq:kl-decomposition}
\end{align}
where $H(\cdot)$ denotes the entropy of a probability distribution and
$\phi(\cdot|I^\prime)$ for $I^\prime\subset[N]$ is a distribution
over the subsets of $I^\prime$ defined as
$\phi(J|I^\prime) = (\lvert I^\prime\rvert + 1)^{-1}\binom{\lvert I^\prime\rvert}{\lvert J\rvert}$.
\end{prop}}
The formal proof of Proposition~\ref{prop:hybrid} is given in Appendix~\ref{sec:proof-kl-decomposition}.
After surveying existing approaches, we propose a ``hybrid'' approach
balancing the optimization of different terms in \eqref{eq:kl-decomposition}.

\paragraph{Exploitation.}
Most of the existing adaptive unmasking schemes
\citep{xiang2022mimt,zhengreparameterized,kim2025train}
just focus on greedy minimization of (\ref{eq:kl-decomposition}.a),
if not explicitly based on entropy.
Let us call this strategy \textit{exploitation}.
Term (\ref{eq:kl-decomposition}.b) is utilized by \citet{ben2025accelerated}
for justifying adaptive selection of $k$, through its lower bound
$\max_{i\in I}H(q_i)$, but it is not used for selecting $I$ itself,
which is chosen through exploitation given $k$.

\paragraph{Exploration.}
In contrast to exploitation methods,
Halton-MaskGIT \citep{besnier2025halton} tries to maximize (\ref{eq:kl-decomposition}.b) by using the two-dimensional
Halton sequence \citep{halton1960efficiency} for image modeling,
which prevents the conditional entropy $\E[\bm{x}_J\sim q_J]{H(q_{i|J}(\cdot|\bm{x}_J))}$
from becoming small
(see \citealp[Section B]{besnier2025halton}).
While not discussed, the use of low-discrepancy sequence including Halton sequence
also aligns with the minimization of (\ref{eq:kl-decomposition}.c).
Let us call \textit{exploration}
the strategy of using $I$ whose indices are non-informative to each other
and informative as a set to the rest of indices.

\blue{Proposition~\ref{prop:hybrid} suggests a natural trade-off between exploration and exploitation:
selecting $I$ greedily by confidence as in exploitation methods
tends to reduce (\ref{eq:kl-decomposition}.a) but may worsen (\ref{eq:kl-decomposition}.b) and (\ref{eq:kl-decomposition}.c), while spatially dispersed choices (exploration) do the opposite.
Our hybrid (bi-criteria) ordering is a natural consequence from the above observations:
we allocate a fraction of the budget to an exploitation-oriented ordering and use the remainder for an exploration-oriented ordering.}

\paragraph{Proposed method: Hybrid approach.}
Let $\bm{i}=(i_1, i_2,\ldots)$ and $\bm{j}=(j_1,j_2,\ldots)$ be two orderings of indices given by different strategies.
To determine $n$ indices to unmask in the next step,
we simply take the first $m$ $(<n)$ indices from $\bm{i}$
and follow the ordering of $\bm{j}$ for the rest to make a merged ordering $\bm{k}$.
Here is an example in the case $n=4$ and $m=2$:
\[
    \bm{i}=(\underline{\textbf{2}}, \underline{\textbf{3}}, 6, 5, 1, 4),\quad
    \bm{j}=(\underline{4}, 3, \underline{1}, 5, 6, 2)
    \quad \Longrightarrow \quad \bm{k}=(\underline{\textbf{2}}, \underline{\textbf{3}}, \underline{4}, \underline{1}, 5, 6),
\]
where underlined indices are chosen for the next unmasking step.
In our experiment in Section~\ref{sec:exp-language}, we use the Halton sequence (exploration) for $\bm{i}$ and moment-based ordering (exploitation) for $\bm{j}$.
We can also apply the caching in the previous section by using the merged ordering.

\section{Experiments}\label{sec:experiments}
We evaluate our theory and proposed methods on unconditional generation tasks in image and text domains.
Additional experimental details are provided in Appendix~\ref{app:experiment}.
\blue{The primary objectives of our experiments differ across domains.
In the image domain (Section~\ref{sec:exp-image}),
where MaskGIT already demonstrates strong performance,
our goal is to validate our theoretical analysis (that the moment sampler approximates MaskGIT; Theorem~\ref{thm:main}) and to isolate the role of implicit temperature sampling,
rather than to pursue further performance improvements.
In contrast, in the language domain (Section~\ref{sec:exp-language}),
we evaluate the practical impact of CTS-based design choices on efficiency and quality-diversity trade-offs.
}

\subsection{Image modeling}\label{sec:exp-image}
In our experiment in the image domain,
we adopted MAGE \citep[ViT-B model]{li2023mage}
as the pretrained masked diffusion model.
MAGE can be regarded as a masked diffusion model
over a VQGAN \citep{esser2021taming} tokenizer space,
trained on ImageNet $256\times256$ \citep{deng2009imagenet}
for \textit{un}conditional generation.
Following the original implementation of MAGE,
we employ a Gumbel temperature of
$\alpha(1 - n/N)$
for the $n$-th step out of $N$ steps,
where $\alpha$ is the global temperature parameter,
and we adopt the cosine unmasking schedule \citep{chang2022maskgit}.
The schedules are shared within the tested methods.
We compared {\bf\texttt{MaskGIT}} (Algorithm~\ref{algo:maskgit-one-round}) and \bt{Moment} (with $\gamma={{1+1/\alpha}}$ in Algorithm~\ref{algo:moment-one-round}, without temperature sampling in the final step)
along with the following:
\begin{itemize}
    \item \bt{Temp}: We only conduct temperature sampling in \bt{Moment}, i.e.,
        replace the step \ref{moment-step-1} with uniformly random selection of indices.
    \item \bt{Random}: Vanilla discrete diffusion sampler conditioned with fixed number of unmasking indices.
        It is equivalent to \bt{MaskGIT} with $\alpha\to\infty$ or at-random index selection.
    \item \bt{Halton}: Index ordering is given by a fixed two-dimensional Halton sequence as in \citep{besnier2025halton}.
        There is no temperature involved in this sampler.
\end{itemize}

\begin{figure}[htbp]
    \centering
    \subfigure{
        \includegraphics[width=0.475\linewidth]{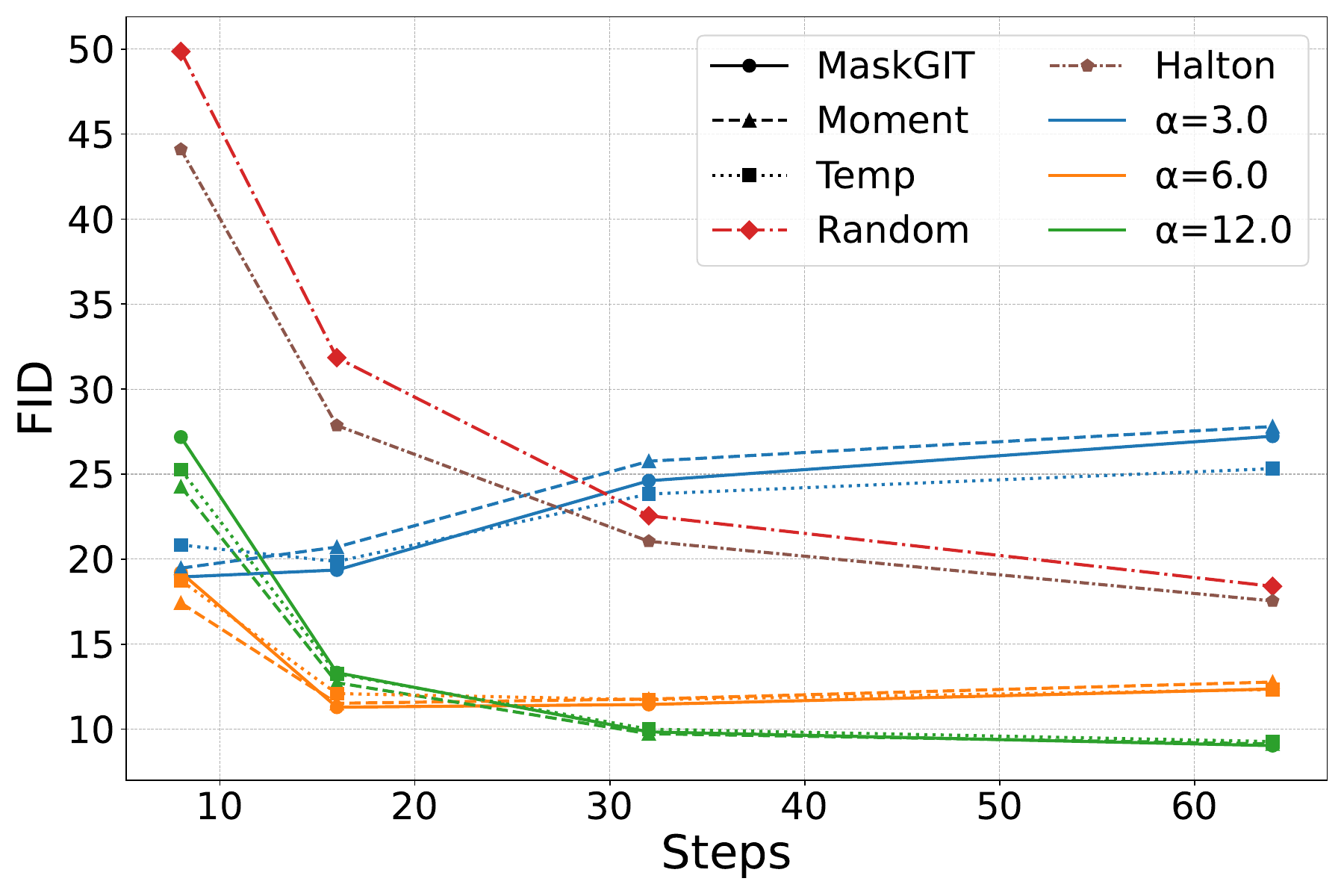}
        \label{fig:fid-temp}
    }
    \subfigure{
        \includegraphics[width=0.475\linewidth]{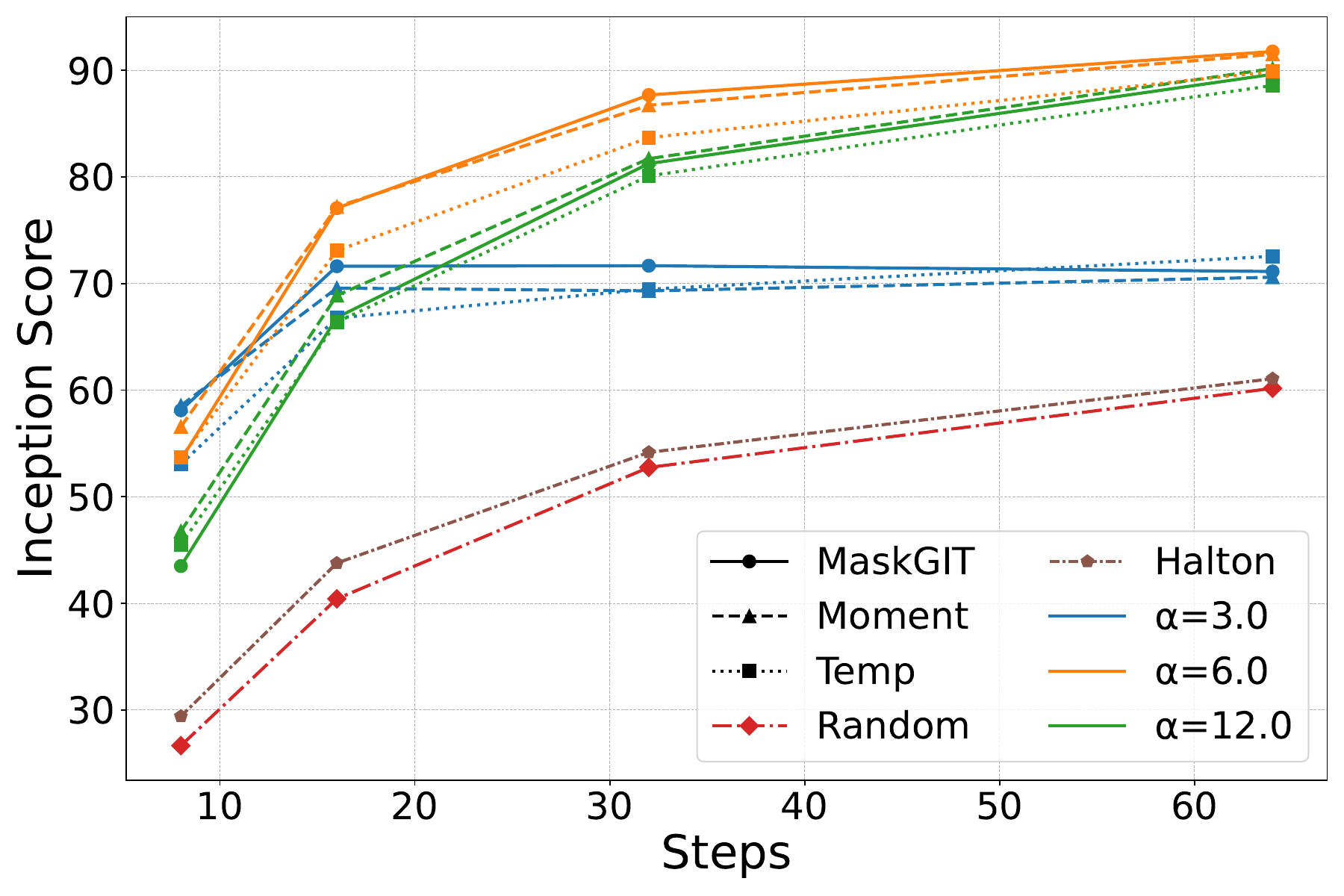}
        \label{fig:is-temp}
    }
    \vspace{-4mm}
    \caption{Fr{\'e}chet Inception Distance (FID, $\downarrow$) and Inception Score ($\uparrow$) against the number of steps for various samplers with MAGE.
        Both metrics were computed by 50,000 generated images.
        {We can see that \bt{Moment} closely approximates \bt{MaskGIT} with the same temperature in both metrics; their quantitative approximation precision is given in Table~\ref{tab:error-mage}.}
        }
    \label{fig:mage}
    \vspace{2mm}
    \centering
    \subfigure{
        \includegraphics[width=0.475\linewidth]{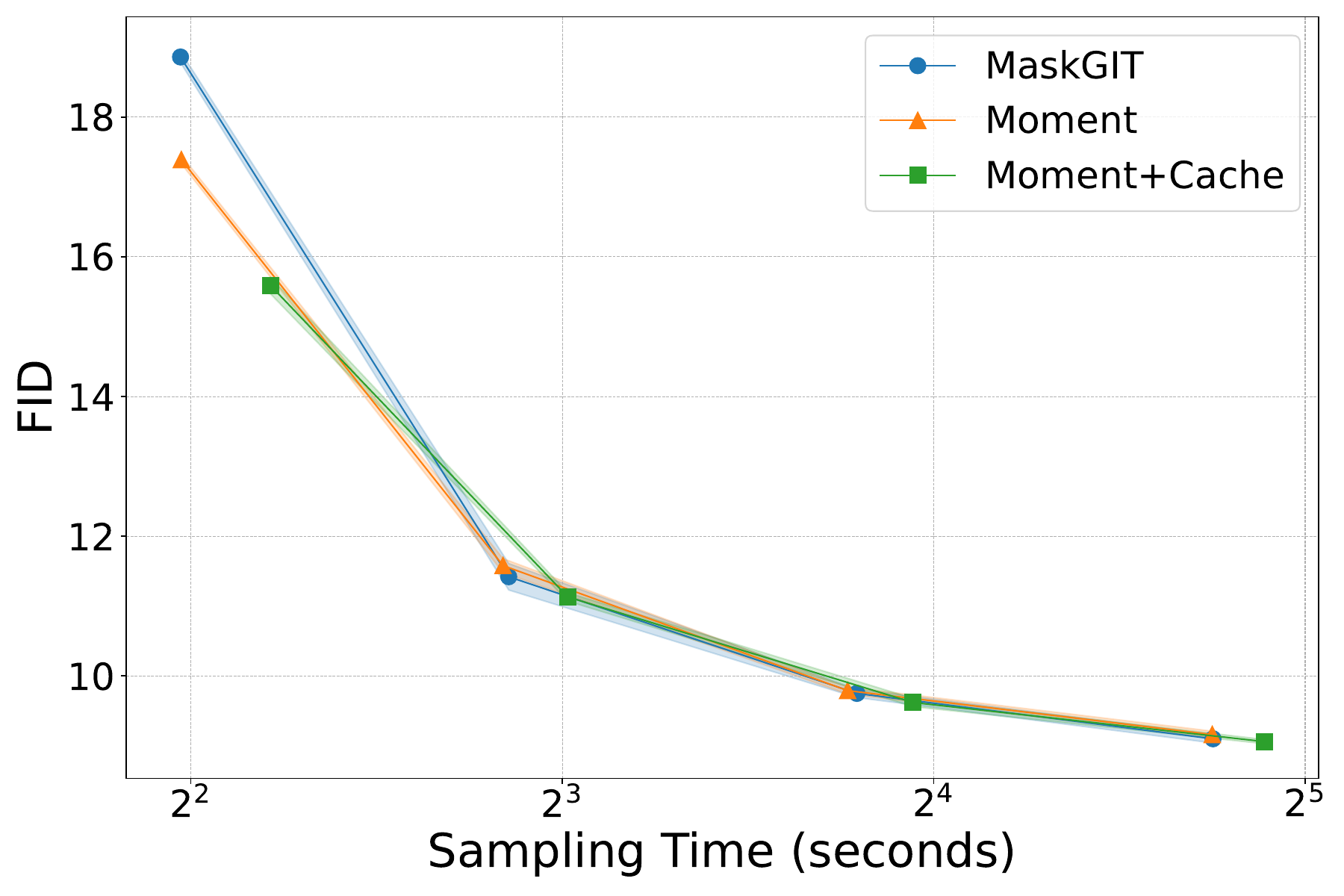}
    }
    \subfigure{
        \includegraphics[width=0.475\linewidth]{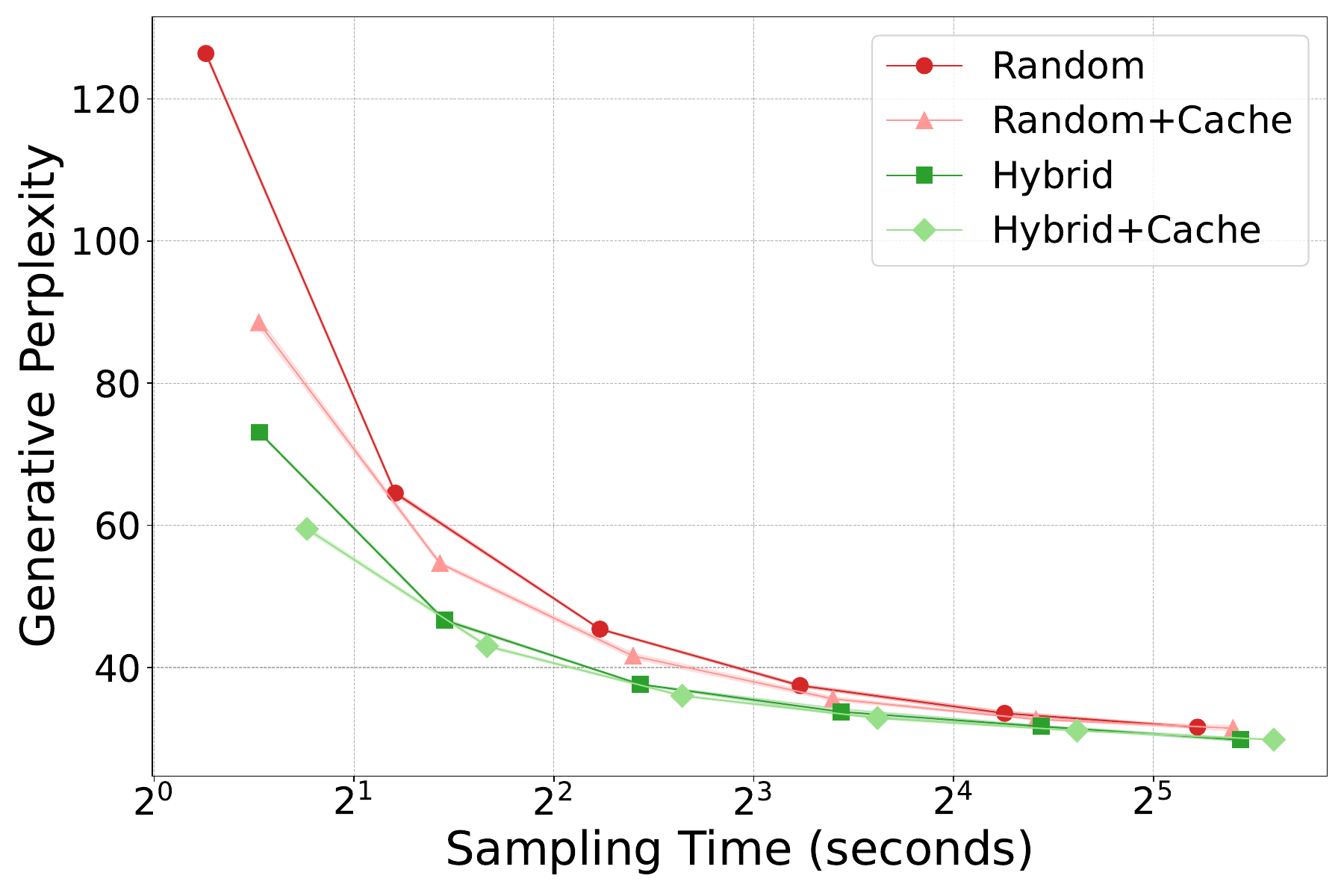}
    }
    \vspace{-4mm}
    \caption{Average performance gains of our proposed samplers against sampling time per batch (on A6000 GPU).
        Each shaded region shows the standard deviation over three trials.
        (\textit{Left}) FID of samplers applied to MAGE.
        (\textit{Right}) Generative Perplexity ($\downarrow$) of samplers applied to SDTT.
        {Especially, \bt{Hybrid} achieves approximately 2x speedup to achieve the same Generative Perplexity as \bt{Random}.}
    }
    \label{fig:latency-perf}
    \vspace{2mm}
    \centering
    \subfigure{
        \includegraphics[width=0.475\linewidth]{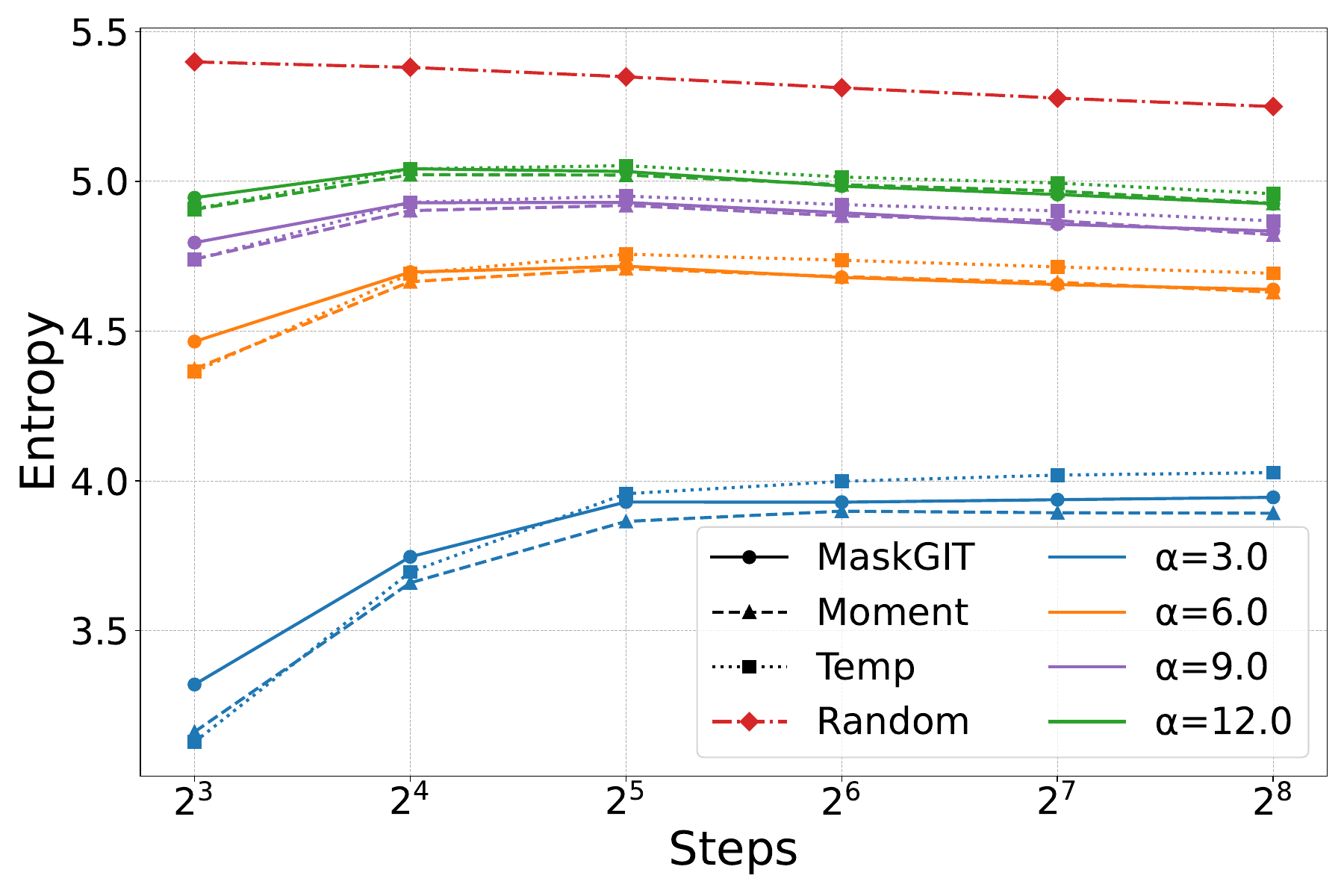}
    }
    \subfigure{
        \includegraphics[width=0.475\linewidth]{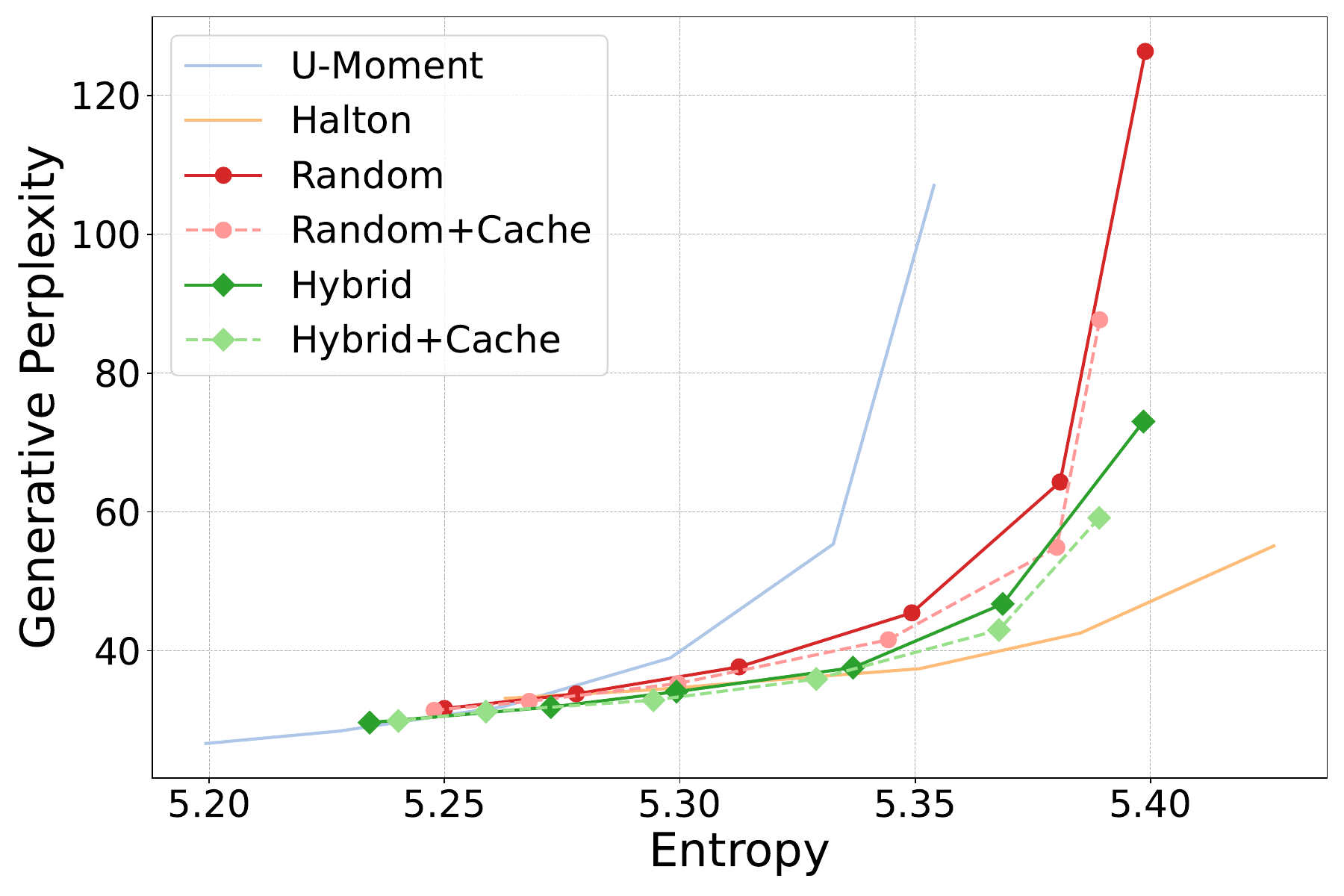}
    }
    \vspace{-4mm}
    \caption{Language experiments. Each plot was computed by 1,024 generated sentences with 1,024 tokens with
    number of steps in $\{8, 16, 32, 64, 128, 256\}$.
    (\textit{Left}) Entropy ($\uparrow$) against the number of steps for temperature-based methods.
    {We can see that temperature sampling significantly harms Entropy, as one can also see qualitatively in Appendix~\ref{sec:samples-sdtt}.}
    (\textit{Right}) Trade-off between Generative Perplexity and Entropy.
    {\bt{Hybrid} samplers improve the trade-off uniformly over \bt{Random} across different numbers of steps.}
    }
    \label{fig:sdtt}
\end{figure}

Figure~\ref{fig:mage} shows the comparison of these five samplers combined with three choices of
global Gumbel temperature: $\alpha\in\{3.0, 6.0, 12.0\}$ ($\alpha=9.0$ is omitted to avoid making the figures too dense).
{Additionally, Table~\ref{tab:error-mage} shows the relative difference of \bt{MaskGIT} and other samplers, in terms of the ImageNet FID performance.
}
We can see that \bt{Moment} shows similar performance to \bt{MaskGIT}, supporting our theory.
A surprising finding is that we can almost replicate the performance of \bt{MaskGIT} using \bt{Temp} without adaptive ordering, which suggests that the performance of the MaskGIT sampler {with MAGE} primarily stems from its implicit temperature sampling rather than the confidence-based ordering.
\blue{We emphasize that these image results are not intended to demonstrate improvements over MaskGIT.
Instead, they serve to validate our analysis by confirming that
(i) the moment sampler clearly approximates MaskGIT and
(ii) sampling temperature largely determines MaskGIT’s behavior in this regime.
Consequently, we focus on practical gains in the language domain.}

\begin{table}[htbp]
\centering
{
\mtabcaption{A}{Mean relative difference (\%) of ImageNet FIDs against reference MaskGIT samplers across 8, 16, 32, 64 steps in unconditional experiment with MAGE.  Given reference sequence $(x_i)_i$,
mean relative difference of sequence $(y_i)_i$ was computed as average of $\lvert 1 - y_i/x_i\rvert$.
Difference under 10\% is bolded.
\bt{Moment} and \bt{Temp} closely approximate \bt{MaskGIT} with the same temperature.
}
\vspace{2mm}
\begin{NiceTabular}
{ccccccccccc}
\toprule
\Block{2-1}{Ref. \bt{MaskGIT} \\ temperature} & \Block{2-1}{\bt{Random}} & \multicolumn{3}{c}{\bt{MaskGIT}} & \multicolumn{3}{c}{\bt{Moment}} & \multicolumn{3}{c}{\bt{Temp}} \\
\cmidrule(lr){3-5} \cmidrule(lr){6-8} \cmidrule(lr){9-11}
 & & $\alpha=3.0$ & 6.0 & 12.0 & 3.0 & 6.0 & 12.0 & 3.0 & 6.0 & 12.0 \\
\midrule
$\alpha=3.0$ & 67.1 & - & 37.8 & 50.4 & \textbf{4.1} & 38.5 & 47.3 & \textbf{5.7} & 36.4 & 47.5 \\
$\alpha=6.0$ & 121.9 & 77.0 & - & 25.1 & 83.7 & \textbf{4.4} & 20.0 & 74.3 & \textbf{3.1} & 21.6 \\
$\alpha=12.0$ & 113.9 & 106.8 & 24.5 & - & 113.4 & 27.6 & \textbf{4.3} & 98.8 & 24.0 & \textbf{3.0} \\
\bottomrule
\end{NiceTabular}
\label{tab:error-mage}
}
\end{table}

In unconditional experiments with MAGE,
we also tested our caching method in Figure~\ref{fig:latency-perf}(Left).
\bt{Moment+Cache} applies the caching technique in Section~\ref{sec:caching} to \bt{Moment}
for creating an ``intermediate'' step
at each sampling step (see Appendix~\ref{app:caching} for details).
For each sampler,
we plotted the best FID among $\alpha\in\{3.0, 6.0, 9.0, 12.0\}$ for each number of sampling steps $N\in\{8,16,32,64\}$,
against the average sampling time per batch.
The experimental result suggests that,
while the performance gain might not be large enough to make the latency overhead negligible,
the partial caching indeed gives some performance boost.

As a supplementary experiment, we compared these samplers in class-conditional setting, using MaskGIT-PyTorch \citep{besnier2023pytorch} as a pretrained masked model.
The relative differences are reported in Table~\ref{tab:error-cond-maskgit} in the appendix.
In this setting, \bt{Moment} again well approximates \bt{MaskGIT} with the same temperature, while \bt{Temp} is not necessarily aligned well with \bt{MaskGIT}.

\subsection{Language modeling}\label{sec:exp-language}
\blue{In the language domain, we focus on evaluating the practical consequences of the CTS reformulation,
since diversity degradation caused by temperature sampling is more pronounced and efficiency gains are more meaningful in this setting.}
We used SDTT \citep[small model with KL divergence target]{deschenaux2024beyond}
as the pretrained model, which was obtained after seven rounds of distillation of the MDLM model~\citep{sahoo2024simple}
trained on the OpenWebText dataset \citep{gokaslan2019openwebtext}.
SDTT is a masked diffusion model over the space of GPT-2 tokenizer \citep{radford2019language}.
In the experiments, we adopted the linear unmasking schedule (i.e., unmasking the same number of positions at each step)
and the same Gumbel temperature schedule as in Section~\ref{sec:exp-image}.
As evaluation metrics, we adopted Generative Perplexity (with GPT-2-large)
and Entropy,
which are in a trade-off relationship, following recent literature \citep{gat2024discrete,zheng2024masked}.

We first compared \bt{MaskGIT}, \bt{Moment}, \bt{Temp}, and \bt{Random}
as described in the previous section.
We confirmed the tendency that the temperature roughly determines the performance;
while the methods with temperature sampling attain lower (= better) Generative Perplexity (Figure~\ref{fig:sdtt-omitted}, Left),
their generated sentences have extremely lower Entropy (Figure~\ref{fig:sdtt}, Left), thus harming the overall quality.
To mitigate this issue, we introduced \bt{U-Moment} (\bt{U} for ``unbiased''),
where we only conduct index selection and omit the temperature sampling in \bt{Moment}, i.e.,
the case of $\gamma=1$ in Algorithm~\ref{algo:moment-one-round}.
In the experiment with ``unbiased'' methods (Figure~\ref{fig:sdtt}, Right),
\bt{Halton} and \bt{U-Moment} showed contrastive behaviors,
where the former gave better trade-off with fewer steps, and the latter worked well with more steps.
By merging these two into \bt{Hybrid} as explained in Section~\ref{sec:explo},
we obtain uniformly better trade-off compared to \bt{Random}
(see Appendix~\ref{sec:hybrid-details} for the details of \bt{Hybrid}).

By using \bt{Hybrid} and/or \bt{+Cache}, we can improve not only the trade-off (Figure~\ref{fig:sdtt}, Right)
but also sampling efficiency in terms of computational time (Figure~\ref{fig:latency-perf}, Right),
leading to 1.5-2x acceleration.
It shows the effectiveness of our approaches in Section~\ref{sec:method}.
Note that,
while the improvement of \bt{Hybrid} seems consistent,
the efficiency gain of \bt{+Cache} can depend on GPUs (Appendix~\ref{sec:caching-compute}).

\blue{
To examine whether these trends persist beyond the SDTT(-small) model,
we further evaluate Hybrid and related CTS variants on
(i) a larger SDTT model,
(ii) an SDTT model augmented with an expanded architecture by the Di4C distillation~\citep{hayakawa2025distillation},
\rev{and (iii) LLaDA-8B~\citep{nie2025large}} in the appendix.
As reported in Appendix~\ref{sec:language-different-models},
\bt{Hybrid} consistently improves the perplexity-entropy trade-off over \bt{Random} and \bt{U-Moment} in both settings,
with comparable or larger efficiency gains.
These results suggest that the benefits of our proposed methods
are not limited to a particular model size or training procedure.}

\section{Related work}\label{sec:survey}
We overview relevant studies on efficient sampling/modeling \rev{and theoretical analyses} for discrete diffusion.
\paragraph{Efficient sampling schemes for discrete diffusion.}
Several efficient sampling schemes have been proposed for speeding up
discrete diffusion.
For general discrete diffusion models, not limited to masked diffusion,
\citet{jys} proposed non-adaptive time-schedule optimization,
and \citet{ren2025fast} proposed discrete diffusion analogues of high-order ODE solvers.
They can be combined with our method in principle.

For masked diffusion setting,
\citet{chang2022maskgit} proposed the MaskGIT sampler,
which has been theoretically analyzed in this paper.
LLaDA \citep{nie2025large} utilizes the MaskGIT sampler in sampling from diffusion LLMs.
\citet{besnier2025halton} adopts the two-dimensional Halton sequence
for selecting unmasking positions in masked image modeling (\textit{exploration}).
There have been also a few top-$k$ sampling (\textit{exploitation}) methods including
entropy \citep{xiang2022mimt},
confidence \citep{zhengreparameterized},
and probability margin \citep{kim2025train}.
Additionally, \citet{ben2025accelerated} recently proposed 
an ``entropy bounded'' unmasking procedure that
adaptively determines the number of unmasked indices in combination with any top-$k$ sampling criteria.
See also Section~\ref{sec:explo} for how these methods can be understood from
the viewpoints of exploration and exploitation.
Finally, the first-hitting sampler \citep{zheng2024masked} demonstrated that
the usual masked diffusion sampler can be accelerated by replacing it with an equivalent
CTS sampler.

\rev{Regarding the exploration-exploitation trade-off, \citet{wei2025accelerating} and \citet{lee2025lookahead}
propose dynamic sampling approaches for diffusion language models
that go beyond simple exploitation by incorporating token-level and trajectory-level uncertainty, respectively.
\citet{fu2025bits} independently provide an information-theoretic lower bound on the number of unmasking steps
for confidence-based parallel sampling and propose Explore-Then-Exploit 
(ETE), which targets high-entropy tokens through beam search to trigger 
``cascades'' of confident predictions; they address the exploration-exploitation 
trade-off in adaptive unmasking of block diffusion \citep{arriola2025block}
differently from ours in both theory and method, complementing our formalization in Section~\ref{sec:explo}.}

\paragraph{Efficient modeling of dimensional correlation in discrete diffusion.}
In contrast to the previous section,
there are also methods that require additional training of models
for efficient sampling in discrete diffusion.
For general discrete diffusion,
\citet{hayakawa2025distillation} proposed ``mixture'' modeling to distill
dimensional correlations learned by many-step teacher models,
and \citet{chen2025overcoming} pointed out that
quantum circuits can realize a one-step diffusion sampler.

For masked diffusion, \citet{lezama2022discrete} proposed the discrete predictor-corrector sampling,
which is essentially based on learning an additional model that determines which token to unmask \citep[see also][]{peng2025path}.
There are two methods exploiting a pretrained autoregressive model
to recover dimensional correlations:
one is based on copula \citep{liu2024discrete},
and the other conducts importance sampling \citep{xu2024energy},
similar to speculative decoding in autoregressive models \citep{guo2025reviving}.
Finally, \citet{zhu2025di} distill pretrained masked image generation models into a one-step sampler.

\paragraph{Caching methods for masked diffusion.}
There are a couple of works on applying KV-cache to masked diffusion models
\citep{arriola2025block,ma2025dkv}.
They both focus on caching the KV information from past tokens for long-term efficiency,
whereas our method in Section~\ref{sec:caching} caches past (already unmasked) and future (not unmasked in this round) tokens for stepwise efficiency.

\rev{
\paragraph{Theoretical analyses of discrete diffusion.}
While this paper primarily studies the theoretical approximation of MaskGIT with a CTS variant, there exist theoretical studies on discrete diffusion samplers that are particularly relevant to our Proposition~\ref{prop:hybrid}.

\citet[][Theorem~3.3 \& Corollary~3.4]{ren2025discrete} derive a Girsanov transform for discrete-state Markov processes and a subsequent KL-divergence estimate, which can be seen as a generalization of Proposition~\ref{prop:hybrid} in  a continuous-time regime,
whereas the main focus of our bound \eqref{eq:kl-decomposition} is an interpretable error decomposition into exploration and exploitation rather than a tight convergence analysis.

Convergence guarantees for discrete/masked diffusion models have been studied for various samplers including $\tau$-leaping \citep{campbell2022continuous,ren2025discrete,liang2025absorb},
score-based denoising~\citep{zhang2025convergence,conforti2025non},
Euler~\citep{liang2025discrete,liang2026sharp} and its higher-order variants~\citep{ren2025fast},
uniformization and its variant~\citep{chen2024convergence,huang2025complexity},
the first-hitting sampler~\citep{zheng2024masked,liang2026sharp},
as well as other sampling strategies \citep{huang2025almost,dmitriev2026efficient}.
\citet{wan2025error} also study convergence within the framework of discrete flow matching \citep{gat2024discrete}.

Another line of work focuses on the statistical error of discrete diffusion models in the context of learning theory \citep{srikanth2025discrete,wakasugi2025state}.

While these studies generally discuss the convergence of samplers or models to the data distribution, our aim is to demystify and approximate the MaskGIT sampler, which is inherently biased (as discussed in Section~\ref{sec:cts-importance}) but useful in practice.
That said, their theoretical rigor can be informative for designing better sampling strategies; tackling the exploration-exploitation tradeoff (Section~\ref{sec:explo}) using such tools could be an interesting future direction.}

\section{Conclusion}
In this paper, we provided a theoretical analysis of the MaskGIT sampler for masked diffusion models, revealing its implicit temperature sampling mechanism.
\blue{Through this analysis, we showed that MaskGIT is asymptotically equivalent to a choose-then-sample (CTS) formulation, instantiated as the moment sampler, which explicitly separates index selection from token sampling.
This CTS reformulation yields interpretability and flexibility that enables efficiency techniques that do not arise from MaskGIT’s original sample-then-choose formulation.}
We further enhanced the efficiency of general CTS algorithms through two techniques: a partial caching approximation for transformer-based models and a hybrid approach that formalizes the exploration-exploitation trade-off in adaptive unmasking.
\blue{Our experiments validate the theory in the image domain and demonstrate the practical benefits of CTS-based design choices in the language domain.}

While our methods are post-hoc and can easily boost the efficiency of pretrained masked diffusion models whose outputs
consist only of \textit{one-token marginals},
more fundamental ``modeling'' approaches as in Section~\ref{sec:survey} are needed to accurately capture
the complex distributions of high-dimensional discrete data.
Theoretical understanding and practical improvement of such methods
will be important future work toward realization of more efficient
discrete generative models.

\subsubsection*{Broader Impact Statement}
While this work mainly presents theoretical contributions, the proposed methods could be applied to generate misleading or harmful content if used with inappropriately trained models.
Furthermore, as our methods modify the sampling distribution, existing biases in pretrained models may be amplified or altered in unpredictable ways.
We are committed to responsible use of these techniques and recommend implementing appropriate content filtering mechanisms in practical applications.

\subsubsection*{Acknowledgments}
We are grateful to Naoki Murata, Joan Serrà, and the anonymous reviewers for their constructive feedback, which significantly improved the paper.

\bibliography{cite}

@inproceedings{loudiscrete,
  title={Discrete Diffusion Modeling by Estimating the Ratios of the Data Distribution},
  author={Lou, Aaron and Meng, Chenlin and Ermon, Stefano},
  booktitle={Proceedings of the 41st International Conference on Machine Learning},
  year={2024}
}

@inproceedings{austin2021structured,
  title={Structured denoising diffusion models in discrete state-spaces},
  author={Austin, Jacob and Johnson, Daniel D and Ho, Jonathan and Tarlow, Daniel and Van Den Berg, Rianne},
  booktitle={Advances in Neural Information Processing Systems},
  volume={34},
  pages={17981--17993},
  year={2021}
}

@inproceedings{gu2022vector,
  title={Vector quantized diffusion model for text-to-image synthesis},
  author={Gu, Shuyang and Chen, Dong and Bao, Jianmin and Wen, Fang and Zhang, Bo and Chen, Dongdong and Yuan, Lu and Guo, Baining},
  booktitle={Proceedings of the IEEE/CVF Conference on Computer Vision and Pattern Recognition},
  pages={10696--10706},
  year={2022}
}

@inproceedings{campbell2022continuous,
  title={A continuous time framework for discrete denoising models},
  author={Campbell, Andrew and Benton, Joe and De Bortoli, Valentin and Rainforth, Thomas and Deligiannidis, George and Doucet, Arnaud},
  booktitle={Advances in Neural Information Processing Systems},
  volume={35},
  pages={28266--28279},
  year={2022}
}

@article{chen2024convergence,
  title={Convergence analysis of discrete diffusion model: Exact implementation through uniformization},
  author={Chen, Hongrui and Ying, Lexing},
  journal={arXiv preprint arXiv:2402.08095},
  year={2024}
}

@inproceedings{sohl2015deep,
  title={Deep unsupervised learning using nonequilibrium thermodynamics},
  author={Sohl-Dickstein, Jascha and Weiss, Eric and Maheswaranathan, Niru and Ganguli, Surya},
  booktitle={Proceedings of the 36th International Conference on Machine Learning},
  pages={2256--2265},
  year={2015},
}

@inproceedings{ho2020denoising,
  title={Denoising diffusion probabilistic models},
  author={Ho, Jonathan and Jain, Ajay and Abbeel, Pieter},
  booktitle={Advances in Neural Information Processing Systems},
  volume={33},
  pages={6840--6851},
  year={2020}
}

@book{cover2006elements,
  title={Elements of information theory},
  author={Thomas M.~Cover and Joy A.~Thomas},
  year={2006},
  edition={Second},
  publisher={John Wiley \& Sons}
}

@inproceedings{ho2022video,
  title={Video diffusion models},
  author={Ho, Jonathan and Salimans, Tim and Gritsenko, Alexey and Chan, William and Norouzi, Mohammad and Fleet, David J},
  booktitle={Advances in Neural Information Processing Systems},
  volume={35},
  pages={8633--8646},
  year={2022}
}

@inproceedings{ou2024your,
  title={Your Absorbing Discrete Diffusion Secretly Models the Conditional Distributions of Clean Data},
  author={Ou, Jingyang and Nie, Shen and Xue, Kaiwen and Zhu, Fengqi and Sun, Jiacheng and Li, Zhenguo and Li, Chongxuan},
  booktitle={Proceedings of the 13th International Conference on Learning Representations},
  year={2025}
}

@inproceedings{liu2024discrete,
  title={Discrete Copula Diffusion},
  author={Liu, Anji and Broadrick, Oliver and Niepert, Mathias and Broeck, Guy Van den},
  booktitle={Proceedings of the 13th International Conference on Learning Representations},
  year={2025}
}

@inproceedings{jys,
  title={Jump Your Steps: Optimizing Sampling Schedule of Discrete Diffusion Models},
  author={Yong-Hyun Park and Chieh-Hsin Lai and Satoshi Hayakawa and Yuhta Takida and Yuki Mitsufuji},
  booktitle={Proceedings of the 13th International Conference on Learning Representations},
  year={2025}
}

@article{besnier2023pytorch,
  title={A Pytorch Reproduction of Masked Generative Image Transformer},
  author={Besnier, Victor and Chen, Mickael},
  journal={arXiv preprint arXiv:2310.14400},
  year={2023}
}

@inproceedings{chang2022maskgit,
  title={Mask{GIT}: Masked generative image transformer},
  author={Chang, Huiwen and Zhang, Han and Jiang, Lu and Liu, Ce and Freeman, William T},
  booktitle={Proceedings of the IEEE/CVF Conference on Computer Vision and Pattern Recognition},
  pages={11315--11325},
  year={2022}
}

@inproceedings{esser2021taming,
  title={Taming transformers for high-resolution image synthesis},
  author={Esser, Patrick and Rombach, Robin and Ommer, Bjorn},
  booktitle={Proceedings of the IEEE/CVF Conference on Computer Vision and Pattern Recognition},
  pages={12873--12883},
  year={2021}
}

@inproceedings{deng2009imagenet,
  title={Imagenet: A large-scale hierarchical image database},
  author={Deng, Jia and Dong, Wei and Socher, Richard and Li, Li-Jia and Li, Kai and Fei-Fei, Li},
  booktitle={2009 IEEE Conference on Computer Vision and Pattern Recognition},
  pages={248--255},
  year={2009},
  organization={Ieee}
}

@inproceedings{deschenaux2024beyond,
  title={Beyond Autoregression: Fast {LLM}s via Self-Distillation Through Time},
  author={Deschenaux, Justin and Gulcehre, Caglar},
  booktitle={Proceedings of the 13th International Conference on Learning Representations},
  year={2025},
}

@inproceedings{xu2024energy,
  title={Energy-Based Diffusion Language Models for Text Generation},
  author={Xu, Minkai and Geffner, Tomas and Kreis, Karsten and Nie, Weili and Xu, Yilun and Leskovec, Jure and Ermon, Stefano and Vahdat, Arash},
  booktitle={Proceedings of the 13th International Conference on Learning Representations},
  year={2025}
}

@misc{gokaslan2019openwebtext,  
	title={OpenWebText Corpus},
	author={Aaron Gokaslan and Vanya Cohen},
	howpublished={\url{http://Skylion007.github.io/OpenWebTextCorpus}}, 
	year={2019}
}

@inproceedings{sahoo2024simple,
    title={Simple and Effective Masked Diffusion Language Models},
    author={Sahoo, Subham Sekhar and Arriola, Marianne and Schiff, Yair and Gokaslan, Aaron and Marroquin, Edgar and Chiu, Justin T and Rush, Alexander and Kuleshov, Volodymyr},
    booktitle={Advances in Neural Information Processing Systems},
    volume={36},
    year={2024}
}

@article{radford2019language,
  title={Language models are unsupervised multitask learners},
  author={Radford, Alec and Wu, Jeffrey and Child, Rewon and Luan, David and Amodei, Dario and Sutskever, Ilya},
  journal={OpenAI blog},
  year={2019}
}

@inproceedings{comunita2024specmaskgit,
  title={Spec{MaskGIT}: Masked generative modeling of audio spectrograms for efficient audio synthesis and beyond},
  author={Comunit{\`a}, Marco and Zhong, Zhi and Takahashi, Akira and Yang, Shiqi and Zhao, Mengjie and Saito, Koichi and Ikemiya, Yukara and Shibuya, Takashi and Takahashi, Shusuke and Mitsufuji, Yuki},
  booktitle={Proceedings of the 25th International Society for Music Information Retrieval Conference},
  pages={420--428},
  year={2024}
}

@inproceedings{maddison2014sampling,
  title={A* sampling},
  author={Maddison, Chris J and Tarlow, Daniel and Minka, Tom},
  booktitle={Advances in Neural Information Processing Systems},
  volume={27},
  pages={3086--3094},
  year={2014}
}

@inproceedings{kool2019stochastic,
  title={Stochastic beams and where to find them: The gumbel-top-k trick for sampling sequences without replacement},
  author={Kool, Wouter and Van Hoof, Herke and Welling, Max},
  booktitle={Proceedings of the 36th International Conference on Machine Learning},
  pages={3499--3508},
  year={2019},
}

@inproceedings{shi2024simplified,
  title={Simplified and generalized masked diffusion for discrete data},
  author={Shi, Jiaxin and Han, Kehang and Wang, Zhe and Doucet, Arnaud and Titsias, Michalis},
  booktitle={Advances in Neural Information Processing Systems},
  volume={37},
  pages={103131--103167},
  year={2024}
}

@article{chen2025overcoming,
  title={Overcoming Dimensional Factorization Limits in Discrete Diffusion Models through Quantum Joint Distribution Learning},
  author={Chen, Chuangtao and Zhao, Qinglin and Zhou, MengChu and He, Zhimin and Situ, Haozhen},
  journal={arXiv preprint arXiv:2505.05151},
  year={2025}
}

@inproceedings{zheng2024masked,
  title={Masked diffusion models are secretly time-agnostic masked models and exploit inaccurate categorical sampling},
  author={Zheng, Kaiwen and Chen, Yongxin and Mao, Hanzi and Liu, Ming-Yu and Zhu, Jun and Zhang, Qinsheng},
  booktitle={Proceedings of the 13th International Conference on Learning Representations},
  year={2025}
}

@book{boucheron2013concentration,
  title={Concentration Inequalities: A Nonasymptotic Theory of Independence},
  author={Boucheron, St{\'e}phane and Lugosi, G{\'a}bor and Massart, Pascal},
  year={2013},
  publisher={Oxford University Press}
}

@inproceedings{zhengreparameterized,
  title={A Reparameterized Discrete Diffusion Model for Text Generation},
  author={Zheng, Lin and Yuan, Jianbo and Yu, Lei and Kong, Lingpeng},
  booktitle={Proceedings of the 1st Conference on Language Modeling},
  year={2024}
}

@inproceedings{hayakawa2025distillation,
  title={Distillation of Discrete Diffusion through Dimensional Correlations},
  author={Hayakawa, Satoshi and Takida, Yuhta and Imaizumi, Masaaki and Wakaki, Hiromi and Mitsufuji, Yuki},
  booktitle={Proceedings of the 42nd International Conference on Machine Learning},
  year={2025}
}

@inproceedings{lezama2022discrete,
  title={Discrete predictor-corrector diffusion models for image synthesis},
  author={Lezama, Jose and Salimans, Tim and Jiang, Lu and Chang, Huiwen and Ho, Jonathan and Essa, Irfan},
  booktitle={Proceedings of the The Eleventh International Conference on Learning Representations},
  year={2023}
}

@inproceedings{ben2025accelerated,
  title={Accelerated Sampling from Masked Diffusion Models via Entropy Bounded Unmasking},
  author={Ben-Hamu, Heli and Gat, Itai and Severo, Daniel and Nolte, Niklas and Karrer, Brian},
  booktitle={Advances in Neural Information Processing Systems},
  volume={38},
  year={2025}
}

@inproceedings{kim2025train,
  title={Train for the Worst, Plan for the Best: Understanding Token Ordering in Masked Diffusions},
  author={Kim, Jaeyeon and Shah, Kulin and Kontonis, Vasilis and Kakade, Sham and Chen, Sitan},
  booktitle={Proceedings of the 42nd International Conference on Machine Learning},
  year={2025}
}

@inproceedings{besnier2025halton,
  title={Halton scheduler for masked generative image transformer},
  author={Besnier, Victor and Chen, Mickael and Hurych, David and Valle, Eduardo and Cord, Matthieu},
  booktitle={Proceedings of the 13th International Conference on Learning Representations},
  year={2025}
}

@inproceedings{ren2025fast,
  title={Fast solvers for discrete diffusion models: Theory and applications of high-order algorithms},
  author={Ren, Yinuo and Chen, Haoxuan and Zhu, Yuchen and Guo, Wei and Chen, Yongxin and Rotskoff, Grant M and Tao, Molei and Ying, Lexing},
  booktitle={Advances in Neural Information Processing Systems},
  volume={38},
  year={2025}
}

@article{guo2025reviving,
  title={Reviving any-subset autoregressive models with principled parallel sampling and speculative decoding},
  author={Guo, Gabe and Ermon, Stefano},
  journal={arXiv preprint arXiv:2504.20456},
  year={2025}
}

@inproceedings{nie2025large,
  title={Large language diffusion models},
  author={Nie, Shen and Zhu, Fengqi and You, Zebin and Zhang, Xiaolu and Ou, Jingyang and Hu, Jun and Zhou, Jun and Lin, Yankai and Wen, Ji-Rong and Li, Chongxuan},
  booktitle={Advances in Neural Information Processing Systems},
  volume={38},
  year={2025}
}

@inproceedings{xiang2022mimt,
  title={{MIMT}: Masked image modeling transformer for video compression},
  author={Xiang, Jinxi and Tian, Kuan and Zhang, Jun},
  booktitle={Proceedings of the 11th International Conference on Learning Representations},
  year={2023}
}

@inproceedings{zhu2025di,
  title={Di$\mathtt{[M]}${O}: Distilling Masked Diffusion Models into One-step Generator},
  author={Zhu, Yuanzhi and Wang, Xi and Lathuili{\`e}re, St{\'e}phane and Kalogeiton, Vicky},
  booktitle={Proceedings of the IEEE/CVF International Conference on Computer Vision},
  pages={18606--18618},
  year={2025}
}

@article{halton1960efficiency,
  title={On the efficiency of certain quasi-random sequences of points in evaluating multi-dimensional integrals},
  author={Halton, John H},
  journal={Numerische Mathematik},
  volume={2},
  pages={84--90},
  year={1960},
  publisher={Springer}
}

@inproceedings{dhariwal2021diffusion,
  title={Diffusion models beat {GAN}s on image synthesis},
  author={Dhariwal, Prafulla and Nichol, Alexander},
  booktitle={Advances in Neural Information Processing Systems},
  volume={34},
  pages={8780--8794},
  year={2021}
}

@inproceedings{gat2024discrete,
  title={Discrete flow matching},
  author={Gat, Itai and Remez, Tal and Shaul, Neta and Kreuk, Felix and Chen, Ricky TQ and Synnaeve, Gabriel and Adi, Yossi and Lipman, Yaron},
  booktitle={Advances in Neural Information Processing Systems},
  volume={37},
  pages={133345--133385},
  year={2024}
}

@inproceedings{li2023mage,
  title={{MAGE}: Masked generative encoder to unify representation learning and image synthesis},
  author={Li, Tianhong and Chang, Huiwen and Mishra, Shlok and Zhang, Han and Katabi, Dina and Krishnan, Dilip},
  booktitle={Proceedings of the IEEE/CVF Conference on Computer Vision and Pattern Recognition},
  pages={2142--2152},
  year={2023}
}

@inproceedings{katharopoulos2020transformers,
  title={Transformers are rnns: Fast autoregressive transformers with linear attention},
  author={Katharopoulos, Angelos and Vyas, Apoorv and Pappas, Nikolaos and Fleuret, Fran{\c{c}}ois},
  booktitle={Proceedings of the 37th International conference on Machine Learning},
  pages={5156--5165},
  year={2020},
  organization={PMLR}
}

@inproceedings{ma2025dkv,
  title={d{KV}-cache: The cache for diffusion language models},
  author={Ma, Xinyin and Yu, Runpeng and Fang, Gongfan and Wang, Xinchao},
  booktitle={Advances in Neural Information Processing Systems},
  volume={38},
  year={2025}
}

@inproceedings{arriola2025block,
title={Block Diffusion: Interpolating Between Autoregressive and Diffusion Language Models},
author={Marianne Arriola and Subham Sekhar Sahoo and Aaron Gokaslan and Zhihan Yang and Zhixuan Qi and Jiaqi Han and Justin T Chiu and Volodymyr Kuleshov},
booktitle={Proceedings of the 13th International Conference on Learning Representations},
year={2025},
}

@article{peng2025path,
  title={Path planning for masked diffusion model sampling},
  author={Peng, Fred Zhangzhi and Bezemek, Zachary and Patel, Sawan and Rector-Brooks, Jarrid and Yao, Sherwood and Bose, Avishek Joey and Tong, Alexander and Chatterjee, Pranam},
  journal={arXiv preprint arXiv:2502.03540},
  year={2025}
}

@inproceedings{kong2021diffwave,
    title={Diff{W}ave: A Versatile Diffusion Model for Audio Synthesis},
    author={Zhifeng Kong and Wei Ping and Jiaji Huang and Kexin Zhao and Bryan Catanzaro},
    booktitle={Proceeding of the 9th International Conference on Learning Representations},
    year={2021},
}

@inproceedings{garcia2023vampnet,
  title={Vamp{N}et: Music Generation via Masked Acoustic Token Modeling},
  author={Garcia, Hugo F Flores and Seetharaman, Prem and Kumar, Rithesh and Pardo, Bryan},
  booktitle={ISMIR 2023 Hybrid Conference},
  year={2023}
}

@article{dubey2024llama,
  title={The {L}lama 3 herd of models},
  author={Dubey, Abhimanyu and Jauhri, Abhinav and Pandey, Abhinav and Kadian, Abhishek and Al-Dahle, Ahmad and Letman, Aiesha and Mathur, Akhil and Schelten, Alan and Yang, Amy and Fan, Angela and others},
  journal={arXiv preprint arXiv:2407.21783},
  year={2024}
}

@article{kulesza2012determinantal,
  title={Determinantal point processes for machine learning},
  author={Kulesza, Alex and Taskar, Ben and others},
  journal={Foundations and Trends{\textregistered} in Machine Learning},
  volume={5},
  number={2--3},
  pages={123--286},
  year={2012},
  publisher={Now Publishers, Inc.}
}

@inproceedings{zhang2025convergence,
  title={Convergence of score-based discrete diffusion models: A discrete-time analysis},
  author={Zhang, Zikun and Chen, Zixiang and Gu, Quanquan},
  booktitle={Proceedings of the 13th International Conference on Learning Representations},
  year={2025}
}

@inproceedings{ren2025discrete,
  title={How Discrete and Continuous Diffusion Meet: Comprehensive Analysis of Discrete Diffusion Models via a Stochastic Integral Framework},
  author={Ren, Yinuo and Chen, Haoxuan and Rotskoff, Grant M and Ying, Lexing},
  booktitle={Proceedings of the 13th International Conference on Learning Representations},
  year={2025}
}

@inproceedings{liang2025absorb,
  title={Absorb and Converge: Provable Convergence Guarantee for Absorbing Discrete Diffusion Models},
  author={Liang, Yuchen and Huang, Renxiang and Lai, Lifeng and Shroff, Ness and Liang, Yingbin},
  booktitle={Advances in Neural Information Processing Systems},
  volume={38},
  year={2025}
}

@article{huang2025almost,
  title={Almost linear convergence under minimal score assumptions: Quantized transition diffusion},
  author={Huang, Xunpeng and Lin, Yingyu and Kuang, Nikki Lijing and Dong, Hanze and Zou, Difan and Ma, Yian and Zhang, Tong},
  journal={arXiv preprint arXiv:2505.21892},
  year={2025}
}

@article{srikanth2025discrete,
  title={Discrete State Diffusion Models: A Sample Complexity Perspective},
  author={Srikanth, Aadithya and Gaur, Mudit and Aggarwal, Vaneet},
  journal={arXiv preprint arXiv:2510.10854},
  year={2025}
}

@article{wan2025error,
  title={Error analysis of discrete flow with generator matching},
  author={Wan, Zhengyan and Ouyang, Yidong and Yao, Qiang and Xie, Liyan and Fang, Fang and Zha, Hongyuan and Cheng, Guang},
  journal={arXiv preprint arXiv:2509.21906},
  year={2025}
}

@inproceedings{liang2025discrete,
  title={Discrete Diffusion Models: Novel Analysis and New Sampler Guarantees},
  author={Liang, Yuchen and Liang, Yingbin and Lai, Lifeng and Shroff, Ness},
  booktitle={Advances in Neural Information Processing Systems},
  volume={38},
  year={2025}
}

@article{liang2026sharp,
      title={Sharp Convergence Rates for Masked Diffusion Models}, 
      author={Yuchen Liang and Zhiheng Tan and Ness Shroff and Yingbin Liang},
      year={2026},
      journal={arXiv preprint arXiv:2602.22505},
}

@article{dmitriev2026efficient,
  title={Efficient Sampling with Discrete Diffusion Models: Sharp and Adaptive Guarantees},
  author={Dmitriev, Daniil and Huang, Zhihan and Wei, Yuting},
  journal={arXiv preprint arXiv:2602.15008},
  year={2026}
}

@article{conforti2025non,
  title={Non-Asymptotic Convergence of Discrete Diffusion Models: Masked and Random Walk dynamics},
  author={Conforti, Giovanni and Durmus, Alain and Pham, Le-Tuyet-Nhi and Raoul, Gael},
  journal={arXiv preprint arXiv:2512.00580},
  year={2025}
}

@article{huang2025complexity,
  title={On the $\epsilon$-Free Inference Complexity of Absorbing Discrete Diffusion},
  author={Huang, Xunpeng and Lin, Yingyu and Jain, Nishant and Wang, Kaibo and Zou, Difan and Ma, Yian and Zhang, Tong},
  journal={arXiv preprint arXiv:2509.21835},
  year={2025}
}

@inproceedings{wakasugi2025state,
  title={State Size Independent Statistical Error Bound for Discrete Diffusion Models},
  author={Wakasugi, Shintaro and Suzuki, Taiji},
  booktitle={Advances in Neural Information Processing Systems},
  year={2025}
}

@article{fu2025bits,
  title={From bits to rounds: Parallel decoding with exploration for diffusion language models},
  author={Fu, Hengyu and Huang, Baihe and Adams, Virginia and Wang, Charles and Srinivasan, Venkat and Jiao, Jiantao},
  journal={arXiv preprint arXiv:2511.21103},
  year={2025}
}

@article{wei2025accelerating,
  title={Accelerating diffusion large language models with slowfast sampling: The three golden principles},
  author={Wei, Qingyan and Zhang, Yaojie and Liu, Zhiyuan and Liu, Dongrui and Zhang, Linfeng},
  journal={arXiv preprint arXiv:2506.10848},
  year={2025}
}

@article{lee2025lookahead,
  title={Lookahead unmasking elicits accurate decoding in diffusion language models},
  author={Lee, Sanghyun and Kim, Seungryong and Park, Jongho and Park, Dongmin},
  journal={arXiv preprint arXiv:2511.05563},
  year={2025}
}
\bibliographystyle{tmlr}

\newpage

\appendix
\section{Algorithms}\label{app:algorithm}
In this section,
we itemize the algorithm pseudocodes of the MakGIT sampler (Algorithm~\ref{algo:maskgit-one-round}),
Moment Sampler (Algorithm~\ref{algo:moment-one-round}), and the general form of choose-then-sample methods (Algorithm~\ref{algo:choose-then-sample}).
{To demonstrate the actual implementation of the moment sampler and highlight its differences from MaskGIT, we give their example PyTorch implementations in Figures~\ref{fig:algo-maskgit} and~\ref{fig:algo-moment}.}
\begin{algorithm}[H]
\caption{One-round of MaskGIT sampler: $\text{OneRoundMaskGIT}((p_i)_{i\in I}, k, \alpha)$}
\label{algo:maskgit-one-round}
\begin{algorithmic}[1] 
\Require
    \Statex $(p_i)_{i\in I}$: Family of probability distributions over $\S$, with $\lvert I\rvert = N$
    \Statex $k\in [N]$: Number of indices to choose
    \Statex $\alpha>0$: Gumbel temperature

\Ensure 
    \Statex $(i_1, \ldots, i_k)\in I^k$: $k$ distinct indices
    \Statex $(x_{i_\ell})_{\ell=1}^k\in \S^k$: Sampled tokens

\State {Independently sample standard Gumbel noise $\xi_i$ for each $i\in I$}
\State {Independently sample $x_i \sim p_i$ for each $i\in I$}
\State $(i_1,\ldots,i_k) \leftarrow \argtop{k}_{i\in I}\{\log p_i(x_i) + \alpha\xi_i\}$
\State \Return $(i_1,\ldots,i_k)$, $(x_{i_\ell})_{\ell=1}^k$
\end{algorithmic}
\end{algorithm}

\begin{algorithm}[H]
\caption{One-round of Moment sampler: $\text{OneRoundMoment}((p_i)_{i\in I}, k, \alpha, \gamma)$}
\label{algo:moment-one-round}
\begin{algorithmic}[1] 
\Require
    \Statex $(p_i)_{i\in I}$: Family of probability distributions over $\S$, with $\lvert I\rvert = N$
    \Statex $k\in [N]$: Number of indices to choose
    \Statex $\alpha>0$: Gumbel temperature
    \Statex $\gamma>0$: Inverse sampling temperature (set $\gamma={1+1/\alpha}$ when approximating MaskGIT)

\Ensure 
    \Statex $(i_1, \ldots, i_k)\in I^k$: $k$ distinct indices
    \Statex $(x_{i_\ell})_{\ell=1}^k\in \S^k$: Sampled tokens

\State Independently sample standard Gumbel noise $\xi_i$ for each $i\in I$
\State $(i_1,\ldots,i_k) \leftarrow \argtop{k}_{i\in I}\{\log\bigl(\sum_{x\in\S}p_i(x)^{1+1/\alpha}\bigr) + \xi_i\}$
\State Independently sample $x_{i_\ell}\sim p_{i_\ell}^\gamma/
    \lVert p_{i_\ell}\rVert_\gamma^\gamma$ for each $\ell\in [k]$
\State \Return $(i_1,\ldots,i_k)$, $(x_{i_\ell})_{\ell=1}^k$
\end{algorithmic}
\end{algorithm}

\begin{algorithm}[h]
\caption{General choose-then-sample algorithm}
\label{algo:choose-then-sample}
\begin{algorithmic}[1] 
\Require
    \Statex $\pi(\cdot|I, \bm{x}_I)$: Distribution over nonempty subsets of $[D]\setminus I$
    conditioned by $I$ and $\bm{x}_I$
    \Statex $p_{j|I}(\cdot|\bm{x}_I)$: Dimension-wise denoising model for any $j, I , \bm{x}_I$
    \Statex $\gamma>0$: Inverse sampling temperature

\Ensure 
    \Statex $\bm{x}\in\S^D$: Generated sample

\State Initialize $I\leftarrow\emptyset$

\While{$I\subsetneq [D]$}
    
    \State Sample $J\sim\pi(\cdot|I, \bm{x}_I)$ \Comment{Ignore $\bm{x}_\emptyset=\emptyset$ when $I=\emptyset$}
    \State Sample $\displaystyle x_j\sim \frac{p_{j|I}(\cdot|\bm{x}_I)^\gamma}{
        \lVert p_{j|I}(\cdot|\bm{x}_I)\rVert^\gamma_\gamma}$ for each $j\in J$
        \Comment{$\bm{x}_{I\cup J}$ has been determined so far}
    \State $I \leftarrow I\cup J$
    
\EndWhile
\State \Return $\bm{x}$
\end{algorithmic}
\end{algorithm}

\newpage

\begin{figure}[H]
\centering
\footnotesize
\begin{lstlisting}[style=samplerpython,linebackgroundcolor=\maskgitlinebackground]
def maskgit_sampler(model, batch_size, seq_len, num_steps, mask_id, alpha):
    """
    MaskGIT sampler with multiple rounds
    
    Args:
        model: Prediction model
        batch_size: Batch size
        seq_len: Sequence length (D)
        num_steps: Number of sampling steps
        mask_id: Mask token ID in the vocabulary set
        alpha: Gumbel temperature for top-k selection
    """
    # Initialize: all tokens are masked
    x = torch.full((batch_size, seq_len), mask_id)
    # Compute k (assume num_steps is a divisor of seq_len)
    k = seq_len // num_steps
    
    for t in range(num_steps):
        # Get probability distribution p_i at each position
        logits = model(x)
        log_probs = torch.log_softmax(logits, dim=-1)  # log p_i

        # Step 1: Sample standard Gumbel noise for each i
        gumbel_noise = torch.distributions.Gumbel(0, 1).sample(
            (batch_size, seq_len)
        )
        
        # Step 2: Sample x_i from p_i for each i
        cat_dist = torch.distributions.Categorical(logits=logits)
        samples = cat_dist.sample()  # x_i's
        
        # Step 3: Select top-k indices (i_1, ..., i_k)
        # Get log p_i(x_i) for sampled tokens
        sampled_log_probs = torch.gather(
            log_probs, dim=-1, index=samples.unsqueeze(-1)
        ).squeeze(-1)  # log p_i(x_i)
        
        # Calculate confidence with Gumbel noise
        confidence = sampled_log_probs + alpha * gumbel_noise

        # Exclude already unmasked positions from top-k determination
        confidence[(x != mask_id)] = float('-inf')
        
        # Get top-k indices with highest confidence (to unmask)
        _, indices_to_unmask = torch.topk(
            confidence, k, largest=True
        )

        # Create boolean mask for indices to unmask
        unmask = torch.zeros_like(x, dtype=torch.bool)
        unmask.scatter_(1, indices_to_unmask, True)

        # Step 4: Update only the selected positions
        x[unmask] = samples[unmask]
        
    return x
\end{lstlisting}
    {
    \mfigcaption{A}{Example PyTorch implementation of MaskGIT sampler. For simplicity, we unmask the same number of positions at each step, and the Gumbel temperature is held constant throughout sampling iterations.
    The ``Steps'' in the comments correspond to those in Algorithm~\ref{algo:maskgit-one-round}, and the highlighted lines show the differences from Moment sampler in Figure~\ref{fig:algo-moment}.}
    \label{fig:algo-maskgit}
    }
\end{figure}

\newpage

\begin{figure}[H]
\centering
\footnotesize
\begin{lstlisting}[style=samplerpython,linebackgroundcolor=\momentlinebackground]
def moment_sampler(model, batch_size, seq_len,
                   num_steps, mask_id, alpha, gamma):
    """
    Moment sampler with multiple rounds
    
    Args:
        model, ..., alpha: Same as maskgit_sampler
                   (alpha is assumed positive for simplicity)
        gamma: Inverse sampling temperature
               (set gamma = 1 + 1/alpha when approximating MaskGIT)
    """
    # Initialize: all tokens are masked
    x = torch.full((batch_size, seq_len), mask_id)
    # Compute k (assume num_steps is a divisor of seq_len)
    k = seq_len // num_steps
    
    for t in range(num_steps):
        # Get probability distribution p_i at each position
        logits = model(x)
        log_probs = torch.log_softmax(logits, dim=-1)  # log p_i
        
        # Step 1: Sample standard Gumbel noise for each i
        gumbel_noise = torch.distributions.Gumbel(0, 1).sample(
            (batch_size, seq_len)
        )
        
        # Step 2: Select top-k indices (i_1, ..., i_k)
        # Compute log(sum_x p_i(x)^(1+1/alpha)) for each i
        moment_scores = torch.logsumexp(
            (1 + 1/alpha) * log_probs, dim=-1
        )
        
        # Calculate confidence with Gumbel noise
        confidence = moment_scores + gumbel_noise
        
        # Exclude already unmasked positions from top-k determination
        confidence[(x != mask_id)] = float('-inf')
        
        # Get top-k indices with highest confidence (to unmask)
        _, indices_to_unmask = torch.topk(
            confidence, k, largest=True
        )
        
        # Step 3: Sample x_{i_ell} from p_{i_ell}^gamma for each ell
        # Extract log-probs only at positions to unmask ([batch_size, k])
        batch_indices = torch.arange(batch_size).unsqueeze(1).expand(-1, k)
        selected_log_probs = log_probs[batch_indices, indices_to_unmask]
        
        # Sample from the distribution proportional to p_{i_ell}^gamma
        cat_dist = torch.distributions.Categorical(
            logits=gamma * selected_log_probs
        )
        samples = cat_dist.sample()
        
        # Step 4: Update only the selected positions
        x[batch_indices, indices_to_unmask] = samples
        
    return x
\end{lstlisting}
    {
    \mfigcaption{B}{Example PyTorch implementation of moment sampler. For simplicity, we unmask the same number of positions at each step, and the Gumbel temperature is held constant throughout sampling iterations.
    The ``Steps'' in the comments correspond to those in Algorithm~\ref{algo:moment-one-round}, and the highlighted lines show the differences from MaskGIT sampler in Figure~\ref{fig:algo-maskgit}.}
    \label{fig:algo-moment}
    }
\end{figure}

\newpage

\section{Formal and informal derivation of moment sampler}
{For notational simplicity in the proofs, let $\beta:=1+1/\alpha$ throughout this section.}
\subsection{Informal derivation of moment sampler}\label{sec:derivation-moment-maskgit}
In this section, we derive the moment sampler approximation \eqref{eq:maskgit-cond-prob}
from \eqref{eq:maskgit-topk-eq} and \eqref{eq:large-sample-approxiamtion}.
First, by applying the approximation \eqref{eq:large-sample-approxiamtion} to \eqref{eq:maskgit-topk-eq},
we have
\begin{equation}
    \P{i_t^* = i_t \mid i_1^*=i_1,\ldots,i_{t-1}^*=i_{t-1}, (x_i)_{i=1}^N}
    \approx \frac{p_{i_t}(x_{i_t})^{1/\alpha}}{\sum_{i\in[N]\setminus I_{t-1}}\lVert p_i\rVert^\beta_\beta},
\end{equation}
for each $t=1,\ldots, k$.
Then, by multiplying this for $t=1,\ldots,\ell-1$,
we obtain
\[
    \P{i_1^*=i_1,\ldots, i_{\ell-1}^* = i_{\ell-1} \mid (x_i)_{i=1}^N}
    \approx
    \prod_{t=1}^{\ell-1} \frac{p_{i_t}(x_{i_t})^{1/\alpha}}{\sum_{i\in[N]\setminus I_{t-1}}\lVert p_i\rVert^\beta_\beta}.
\]
Note that the right-hand side is independent of $(x_i)_{i\in [N]\setminus I_\ell}$,
so we can replace the conditioning on $(x_i)_{i=1}^N$ by $(x_i)_{i\in I_\ell}$,
By marginalizing out $x_i$ with $i\in I_{\ell-1}$ from this, we have
\begin{align*}
    \P{i_1^*=i_1,\ldots, i_\ell^* = i_\ell \mid x_{i_\ell}}
    &= \sum_{x_{i_1}}\cdots\sum_{x_{i_{\ell-1}}} \P{i_1^*=i_1,\ldots, i_\ell^* = i_\ell \mid (x_i)_{i\in I_\ell}}
    \P{(x_i)_{i\in I_{\ell-1}}\mid x_{i_\ell}}\\
    &\approx \left(\prod_{t=1}^{\ell-1} \frac{\sum_{x_{i_t}}p_{i_t}(x_{i_t})^{1/\alpha}\cdot p_{i_t}(x_{i_t})}{
        \sum_{i\in[N]\setminus I_{t-1}}\lVert p_i\rVert^\beta_\beta}\right)
        \frac{p_{i_\ell}(x_{i_\ell})^{1/\alpha}}{\sum_{i\in[N]\setminus I_{\ell-1}}\lVert p_i\rVert^\beta_\beta}\\
    &= \underbrace{\left(\prod_{t=1}^{\ell-1} \frac{\lVert p_{i_t}\rVert_\beta^\beta}{
        \sum_{i\in[N]\setminus I_{t-1}}\lVert p_i\rVert^\beta_\beta}\right)}_{=:C_{\ell-1}(i_1,\ldots,i_{\ell-1})}
        \frac{p_{i_\ell}(x_{i_\ell})^{1/\alpha}}{\sum_{i\in[N]\setminus I_{\ell-1}}\lVert p_i\rVert^\beta_\beta}
        .
\end{align*}
Therefore, we have
\begin{align*}
    &\P{x_{i_\ell}, i_\ell^*=i_\ell \mid i_1^*=i_1,\ldots, i_{\ell-1}^* = i_{\ell-1}}
    = \frac{\P{i_1^*=i_1,\ldots, i_\ell^* = i_\ell \mid x_{i_\ell}} \P{x_{i_\ell}}}{
        \P{i_1^*=i_1,\ldots, i_{\ell-1}^* = i_{\ell-1}}}\\
    &\qquad\qquad\approx
    \frac{C_{\ell-1}(i_1,\ldots,i_{\ell-1})}{\P{i_1^*=i_1,\ldots, i_{\ell-1}^* = i_{\ell-1}}}
    \frac{p_{i_\ell}(x_{i_\ell})^\beta}{\sum_{i\in[N]\setminus I_{\ell-1}}\lVert p_i\rVert^\beta_\beta}
    \propto \frac{p_{i_\ell}(x_{i_\ell})^\beta}{\sum_{i\in[N]\setminus I_{\ell-1}}\lVert p_i\rVert^\beta_\beta}.
\end{align*}
Since the final right-hand side is actually a normalized probability distribution over $\S\times([N]\setminus I_{\ell-1})$,
we have derived \eqref{eq:maskgit-cond-prob}.

\subsection{Strategy for formal proof of Theorem~\ref{thm:main}}\label{sec:proof-strategy}
Formally, we work under the following setting:
\begin{sett}[MaskGIT sampler]\label{assp:setting}
We are given $\alpha>0$ and $N$ probability distributions $p_1,\ldots,p_N$ over a finite set $\S$.
We sample $x_i\sim p_i$ and standard Gumbel noise $\xi_i$ independently for each $i\in[N]$,
and let
$(i^*_1, \ldots, i^*_k)=\argtop{k}_{i\in[N]}\{\log p_i(x_i) + \alpha\xi_i\}$
with $k\in[N]$.
\end{sett}
Let us start by applying Bernstein's inequality (see, e.g., \citealt[Corollary 2.11]{boucheron2013concentration})
to the sum of $p_i(x_i)^{1/\alpha} - \E{p_i(x_i)^{1/\alpha}}$.
Since the summand is always within $[-1,1]$,
for $t\ge0$, we have
\begin{equation}
    \P{\sum_{i=1}^N p_i(x_i)^{1/\alpha}-\sum_{i=1}^N\E{p_i(x_i)^{1/\alpha}} \le -t}
    \le \exp\left( -\frac{t^2}{2(\sigma^2 + t/3)}
    \right),
    \label{eq:app:bernstein}
\end{equation}
where $\sigma^2:=\sum_{i=1}^N\E{\left(p_i(x_i)^{1/\alpha} - \E{p_i(x_i)^{1/\alpha}}\right)^2}$.
Since $0\le p_i(x_i)^{1/\alpha}\le 1$, we have
\begin{equation}
    \sigma^2 \le \sum_{i=1}^N \E{p_i(x_i)^{1/\alpha}\cdot p_i(x_i)^{1/\alpha}}\le
    \sum_{i=1}^N \E{p_i(x_i)^{1/\alpha}}.
    \label{eq:app:1-upper-bound}
\end{equation}
Now, recall that $\E{p_i(x_i)^{1/\alpha}}=\lVert p_i\rVert_\beta^\beta$ for $\beta=1+1/\alpha$.
By letting $t=\ve\sum_{i=1}^N \lVert p_i\rVert_\beta^\beta$ for some $0\le\ve\le1$
and combining \eqref{eq:app:bernstein} and \eqref{eq:app:1-upper-bound},
we have
\begin{align}
    \P{\sum_{i=1}^N p_i(x_i)^{1/\alpha}
    \le (1-\ve)\sum_{i=1}^N \lVert p_i\rVert_\beta^\beta}
    &\le \exp\left(
        -\frac{\ve^2\left(\sum_{i=1}^N \lVert p_i\rVert_\beta^\beta\right)^2}{
            2(
                \sum_{i=1}^N \lVert p_i\rVert_\beta^\beta
                + \ve \sum_{i=1}^N \lVert p_i\rVert_\beta^\beta/3
            )
        }
    \right)\nonumber\\
    &= \exp\left(
        -\frac{\ve^2\left(\sum_{i=1}^N \lVert p_i\rVert_\beta^\beta\right)}{
            2(1+\ve/3)
        }
    \right) \nonumber \\
    &\le \exp\left(
        -\frac{3\ve^2}8\sum_{i=1}^N \lVert p_i\rVert_\beta^\beta
    \right).
    \label{eq:app:all-bernstein}
\end{align}

To utilize this estimate for our analysis of MaskGIT sampler,
we prove the following proposition.
The proof is given in Appendix~\ref{sec:proof-aux}.
\begin{prop}\label{prop:aux}
    Under Setting~\ref{assp:setting},
    let $[N]_{<k}$ denote the set of all the subsets of $[N]$ with cardinality less than $k$.
    Then, for $\ve\in[0,1]$, we have
    \begin{equation}
        \sum_{i\in [N]\setminus I} p_i(x_i)^{1/\alpha}
        > \left(1-\ve-\frac{(k-1)S^{1/\alpha}}{N-k+1}\right)
        \sum_{i\in [N]\setminus I} \lVert p_i\rVert_\beta^\beta
        \quad \text{for all $I\in[N]_{<k}$}
        \label{eq:app:prop-aux}
    \end{equation}
    with probability at least
    $1 - \exp\left(
        -\frac38 \ve^2 N\lvert \S\rvert^{-1/\alpha}
    \right)$.
\end{prop}

We use Proposition~\ref{prop:aux} to prove the following assertion,
whose proof is given in Appendix~\ref{sec:proof-maskgit-upper-bound}
\begin{prop}\label{prop:maskgit-upper-bound}
    Under Setting~\ref{assp:setting},
    assume $\ve+\frac{(k-1)\lvert \S\rvert^{1/\alpha}}{N-k+1}<1$ holds.
    For $\ve\in[0,1]$, let us define $\mathcal{Z}_\ve\subset\S^N$ as the set of
    $(z_i)_{i=1}^N \in \S^N$ that satisfies the following inequality:
    \begin{align*}
        &\P{i_1^*=i_1,\ldots, i_{k}^* = i_{k}
            \mid (x_i)_{i=1}^N = (z_i)_{i=1}^N}
        \\
        &\qquad\qquad <
        \left(1-\ve-\frac{(k-1)\lvert\S\rvert^{1/\alpha}}{N-k+1}\right)^{-k}
        \prod_{\ell=1}^{k} \frac{p_{i_\ell}(z_{i_\ell})^{1/\alpha}}{\sum_{i\in[N]\setminus
        I_{\ell-1}}\lVert p_i\rVert^\beta_\beta},
    \end{align*}
    where $I_{\ell-1}:=\{i_1,\ldots,i_{\ell-1}\}$.
    Then, we have $\P{(x_i)_{i=1}^N \in \mathcal{Z}_\ve}\ge 
        1 - \exp\left(
            -\frac38 \ve^2 N\lvert \S\rvert^{-1/\alpha}
        \right)$.
\end{prop}

We next compute the probability distribution of the moment sampler.

\begin{sett}[Moment sampler]\label{sett:moment}
    We are given $\alpha>0$ and $N$ probability distributions
    $p_1,\ldots,p_N$
    over a finite set $\S$.
    Let $\beta=1+1/\alpha$.
    We sample $y_i\sim p_i^\beta/\lVert p_i\rVert_{\beta}^\beta$
    and standard Gumbel noise $\eta_i$ independently for each $i\in[N]$,
    and let $(j_1^*, \ldots, j_k^*)=\argtop{k}_{i\in[N]}\{\log\lVert p_i\rVert_\beta^\beta + \eta_i\}$ with $k\in [N]$.
\end{sett}

\begin{prop}\label{prop:moment-exact-probabilities}
    Under Setting~\ref{sett:moment}, for each distinct indices $i_1,\ldots,i_k\in[N]$
    and (not necessarily distinct) $z_{i_1},\ldots,z_{i_k}\in\S$,
    we have
    \begin{equation}
        \P{j_1^*=i_1,\ldots,j_k^*=i_k,
            (y_{j_\ell^*})_{\ell=1}^k = (z_{i_\ell})_{\ell=1}^k}
        = \prod_{\ell=1}^{k} \frac{p_{i_\ell}(z_{i_\ell})^\beta}{\sum_{i\in[N]\setminus
        I_{\ell-1}}\lVert p_i\rVert^\beta_\beta}
        \label{eq:app:moment-joint}
    \end{equation}
    where $I_{\ell-1}:=\{i_1,\ldots,i_{\ell-1}\}$.
\end{prop}
The proof is given in Appendix~\ref{sec:proof-moment-exact}.

Let us finally prove Theorem~\ref{thm:main}.
The following is its restatement under the above settings.
\begin{thm}\label{thm:main-formal}
    Under Setting~\ref{assp:setting}~\&~\ref{sett:moment},
    for each $\bm{i}=(i_1,\ldots,i_k)\in [N]^k$
    and $(z_{i_\ell})_{\ell=1}^k\in \S^k$,
    let
    \begin{align*}
        p_\mathrm{MaskGIT}(\bm{i}, (z_{i_\ell})_{\ell=1}^k)&:=
        \P{i_1^*=i_1,\ldots,i_k^*=i_k,
            (x_{i_\ell^*})_{\ell=1}^k = (z_{i_\ell})_{\ell=1}^k},\\
        p_\mathrm{moment}(\bm{i}, (z_{i_\ell})_{\ell=1}^k)
        &:=
        \P{j_1^*=i_1,\ldots,j_k^*=i_k,
            (y_{j_\ell^*})_{\ell=1}^k = (z_{i_\ell})_{\ell=1}^k}.
    \end{align*}
    Then, as probability distributions over $[N]^k\times\S^k$,
    we have
    \begin{equation}
        d_\mathrm{TV}({p}_\mathrm{moment}, {p}_\mathrm{MaskGIT})
        \le 5\sqrt{\frac{k^2\lvert\S\rvert^{1/\alpha}}{N}}\left(
            1 + \sqrt{\log^+\!\left(\frac{N}{k^2\lvert\S\rvert^{1/\alpha}}\right)}\right),
        \nonumber
    \end{equation}
    where $\log^+(x):=\log(\max\{1, x\})$ for $x\in\R$.
\end{thm}
This main result is proven in Appendix~\ref{sec:proof-main}.

\section{Proofs}
\subsection{Proof of Proposition~\ref{prop:one-by-one-cts}}\label{sec:proof-cts}
\begin{proof}
    Let $I_0=\emptyset$ and inductively (and randomly) define
    \[
        \sigma_n\sim \pi(\cdot|\bm{x}_{I_{n-1}}),
        \qquad
        x_{\sigma_n}\sim p_{\sigma_n|I_{n-1}}(\cdot|\bm{x}_{I_{n-1}}),
        \qquad
        I_n:=I_{n-1}\cup\{\sigma_n\}
    \]
    for $n=1,\ldots,D$,
    where we abuse the notation to simplify $\{\sigma_n\}\sim \pi$ into $\sigma_n\sim \pi$.
    Note that $(\sigma_1,\ldots,\sigma_D$ is a random permutation of $(1,\ldots,D)$.
    Let us write $p(\bm{x}, \bm{\sigma})$ be the joint distribution of
    $\bm{x}$ and $\bm{\sigma}=(\sigma_n)_{n=1}^D$ (the latter is constrained to be a permutation).
    Then, we have
    \begin{align*}
        p(\bm{x}, \bm{\sigma})
        &= \prod_{n=1}^D \pi(\sigma_n|\bm{x}_{I_{n-1}})
            p_{\sigma_n|I_{n-1}}(x_{\sigma_n}|\bm{x}_{I_{n-1}})\\
        &= \left(\prod_{n=1}^D \pi(\sigma_n|\bm{x}_{I_{n-1}})\right)
        \left(\prod_{n=1}^D 
            p_{\sigma_n|I_{n-1}}(x_{\sigma_n}|\bm{x}_{I_{n-1}})\right)
        = q_\text{data}(\bm{x}) \prod_{n=1}^D \pi(\sigma_n|\bm{x}_{I_{n-1}})
    \end{align*}
    from the assumption $p_{j|I}=q_{j|I}$.
    Thus, it suffices to prove that $\sum_{\bm{\sigma}}\prod_{n=1}^D\pi(\sigma_n|\bm{x}_{I_{n-1}})=1$.
    To this end, let us prove the following for $k=1,\ldots,D$ by induction on $k$
    (the case $k=D$ is what we would like to prove ultimately):
    \begin{equation}
        \sum_{\sigma_{D-k+1},\ldots, \sigma_D}
        \prod_{n=D-k+1}^D\pi(\sigma_n|\bm{x}_{I_{n-1}})=1
        \qquad
        \text{for any $I_{D-k}$ with $\lvert I_{D-k}\rvert = D-k$
        and $\bm{x}_{I_{D-k}}$}.
    \end{equation}
    It is clearly true for $k=1$, since it is just the total probability sum
    of the probability distribution $\pi(\cdot|\bm{x}_{I_{D-1}})$.
    For $k\ge2$,
    from the induction hypothesis, we have
    \begin{align*}
        \sum_{\sigma_{D-k+1},\ldots, \sigma_D}
        \prod_{n=D-k+1}^D\pi(\sigma_n|\bm{x}_{I_{n-1}})
        &= \sum_{\sigma_{D-k+1}}\pi(\sigma_{D-k+1}|\bm{x}_{I_{D-k}})
            \underbrace{\sum_{\sigma_{D-k+2},\ldots, \sigma_D}
            \prod_{n=D-k+1}^D\pi(\sigma_n|\bm{x}_{I_{n-1}})}_{=1\ \text{by induction hypothesis}}\\
        &= \sum_{\sigma_{D-k+1}}\pi(\sigma_{D-k+1}|\bm{x}_{I_{D-k}})=1.
    \end{align*}
    Therefore, the proof has been completed.
\end{proof}

\subsection{Proof of Proposition~\ref{prop:hybrid}}\label{sec:proof-kl-decomposition}
\begin{proof}
    By using the chain rule of KL divergence \citep[Theorem~2.5.3]{cover2006elements},
    we have
    \begin{align}
        D_\mathrm{KL}(q\,\Vert\,p)
        =D_\mathrm{KL}(q_I\,\Vert\,p_I)
        + \E[\bm{x}_I\sim q_I]{D_\mathrm{KL}(q_{I^c|I}(\cdot|\bm{x}_I)\,\Vert\,p_{I^c|I}(\cdot|\bm{x}_I))},
        \label{eq:kl-dec-1}
    \end{align}
    which shows the first inequality in \eqref{eq:kl-decomposition}.
    Let us first consider the KL divergence between $q_I$ and $p_I$.
    First, we have
    \begin{align}
        D_\mathrm{KL}(q_I\,\Vert\,p_I)
        &=
        \E[\bm{x}_I\sim q_I]{- \log\Biggl(\prod_{i\in I}q_i(x_i)\Biggr)
        + \log q_I(\bm{x}_I)}\nonumber\\
        &=\sum_{i\in I}\E[x_i\sim q_i]{-\log q_i(x_i)}
        - \E[\bm{x}_I\sim q_I]{-\log q_I(\bm{x}_I)}\nonumber\\
        &= \sum_{i\in I}H(q_i) - \E[\bm{x}_I\sim q_I]{-\log q_I(\bm{x}_I)}.
        \label{eq:kl-dec-2}
    \end{align}
    Next, by using a permutation $(\sigma_1,\ldots,\sigma_k)$ of $I$,
    the remainder term of \eqref{eq:kl-dec-2} (entropy of $q_I$)
    can be rewritten as follows:
    \begin{align}
        \E[\bm{x}_I\sim q_I]{-\log q_I(\bm{x}_I)}
        &= \sum_{x_{\sigma_1},\ldots,x_{\sigma_k}}
        \prod_{j=1}^k q_{\sigma_j|I_{j-1}}(x_{\sigma_j}|\bm{x}_{I_{j-1}})
        \sum_{\ell=1}^k (-\log q_{\sigma_\ell|I_{\ell-1}}(x_{\sigma_\ell}|\bm{x}_{I_{\ell-1}}))
        \nonumber\\
        &=
        \sum_{\ell=1}^k
        \sum_{x_{\sigma_1},\ldots,x_{\sigma_\ell}}
        q_{I_{\ell-1}}(\bm{x}_{I_{\ell-1}})q_{\sigma_\ell|I_{\ell-1}}(x_{\sigma_\ell}|\bm{x}_{I_{\ell-1}})
        (-\log q_{\sigma_\ell|I_{\ell-1}}(x_{\sigma_\ell}|\bm{x}_{I_{\ell-1}}))\nonumber\\
        &=\sum_{\ell=1}^k\mathbb{E}_{\bm{x}_{I_{\ell-1}}\sim q_{I_{\ell-1}}}
        \E[x_{\sigma_\ell}\sim q_{\sigma_\ell|I_{\ell-1}}]{-\log q_{\sigma_\ell|I_{\ell-1}}(x_{\sigma_\ell}|\bm{x}_{I_{\ell-1}})}\nonumber\\
        &=\sum_{\ell=1}^k\E[\bm{x}_{I_{\ell-1}}\sim q_{I_{\ell-1}}]{H(q_{\sigma_\ell|I_{\ell-1}}(\cdot|\bm{x}_{I_{\ell-1}}))}.
        \label{eq:kl-dec-3}
    \end{align}
    Let us consider taking the average of the right-hand side over all the permutations.
    With the uniformly random permutation $(\sigma_1,\ldots,\sigma_k)$,
    for each $i\in I$, $\ell$ with $\sigma_\ell=i$ takes the uniformly distribution over $[k]$.
    $I_{\ell-1}$ (conditioned by $\ell$) then takes the uniform distribution over
    all the possible size-$(\ell-1)$ subsets of $I\setminus\{i\}$.
    Therefore, if we write such an $\ell$ as $\ell=\sigma^{-1}(i)$,
    the probability that $J\subset I\setminus\{i\}$ is chosen as $I_{\sigma^{-1}(i)-1}$
    can be computed as
    \[
        \underbrace{\frac1k}_{\P{\sigma^{-1}(i)-1=\lvert J\rvert}}
        \binom{k-1}{\lvert J\rvert}^{-1}
        =\phi(J|I\setminus\{i\}),
    \]
    where $\phi$ is the probability distribution defined just before \eqref{eq:kl-decomposition}.
    By applying this to \eqref{eq:kl-dec-3},
    we have
    \[
        \E[\bm{x}_I\sim q_I]{-\log q_I(\bm{x}_I)}
        = \sum_{i\in I}\mathbb{E}_{J\sim \phi(\cdot|I\setminus\{i\})}
        \E[\bm{x}_J\sim q_J]{H(q_{i|J}(\cdot|\bm{x}_J))}.
    \]
    Combining it with \eqref{eq:kl-dec-2},
    we obtain
    \begin{equation}
        D_\mathrm{KL}(q_I\,\Vert\,p_I)
        = \sum_{i\in I}H(q_i) - \sum_{i\in I}\mathbb{E}_{J\sim \phi(\cdot|I\setminus\{i\})}
        \E[\bm{x}_J\sim q_J]{H(q_{i|J}(\cdot|\bm{x}_J))}.
        \label{eq:kl-dec-4}
    \end{equation}
    
    Finally, for the remaining term,
    it suffices to prove
    \[
        \E[\bm{x}_I\sim q_I]{D_\mathrm{KL}(q_{I^c|I}(\cdot|\bm{x}_I)
        \,\Vert\,p_{I^c|I}(\cdot|\bm{x}_I))}
        \le \E[\bm{x}_I\sim q_I]{\sum_{i\in [N]\setminus I}H(q_{i|I}(\cdot|\bm{x}_I))}.
    \]
    This can be proven by using the positivity of entropy
    and modifying \eqref{eq:kl-dec-4},
    with $q$ replaced by $q_{\cdot|I}$ and $I$ replaced by $I^c$.
    Thus, the proof is completed.
\end{proof}

\subsection{Proof of Proposition~\ref{prop:aux}}\label{sec:proof-aux}
\begin{proof}
    Let $\Omega_\ve$ be the event under which we have
    $\sum_{i=1}^N p_i(x_i)^{1/\alpha}
        > (1-\ve)\sum_{i=1}^N \lVert p_i\rVert_\beta^\beta$.
    From \eqref{eq:app:all-bernstein}, we have
    $\P{\Omega_\ve}\ge 1 - \exp\bigl(-\frac{3\ve^2}8 \sum_{i=1}^N\lVert p_i\rVert_\beta^\beta\bigr)$.
    For any $I\in [N]_{<k}$, under $\Omega_\ve$, we have
    \begin{align}
        \sum_{i\in [N]\setminus I} p_i(x_i)^{1/\alpha}
        &= \sum_{i=1}^N p_i(x_i)^{1/\alpha} - \sum_{i\in I}p_i(x_i)^{1/\alpha} \nonumber \\
        &> (1-\ve) \sum_{i=1}^N \lVert p_i\rVert_\beta^\beta - \sum_{i\in I}p_i(x_i)^{1/\alpha} \nonumber\\
        &\ge (1-\ve)\sum_{i\in [N]\setminus I} \lVert p_i\rVert_\beta^\beta - \sum_{i\in I}p_i(x_i)^{1/\alpha}
        \label{eq:app:omega-tau-1}
    \end{align}
    Now we want to estimate the ratio between
    $\sum_{i\in I} p_i(x_i)^{1/\alpha}$ and $\sum_{i\in[N]\setminus I} \lVert p_i\rVert_\beta^\beta$.
    Let $S:=\lvert\S\rvert$.
    From H{\"o}lder's inequality, we have
    \begin{equation}
        \lVert p_i\rVert_\beta^\beta
        = S\cdot\frac1S\sum_{x\in\S}p_i(x)^\beta
        \ge S\left(\frac1S\sum_{x\in\S}p_i(x)\right)^\beta
        = S^{1-\beta} = S^{-1/\alpha}.
        \label{eq:app:lower-bound-moment}
    \end{equation}
    By using this, we obtain
    \begin{equation*}
        \frac{\sum_{i\in I}p_i(x_i)^{1/\alpha}}{\sum_{i\in [N]\setminus I} \lVert p_i\rVert_\beta^\beta}
        \le \frac{(k-1)}{(N-k+1)S^{-{1/\alpha}}}
        = \frac{(k-1)S^{1/\alpha}}{N-k+1}.
    \end{equation*}
    By applying this to \eqref{eq:app:omega-tau-1}, under $\Omega_\ve$, we have
    \begin{equation*}
        \sum_{i\in [N]\setminus I} p_i(x_i)^{1/\alpha}
        > \left(1-\ve-\frac{(k-1)S^{1/\alpha}}{N-k+1}\right)
        \sum_{i\in [N]\setminus I} \lVert p_i\rVert_\beta^\beta.
    \end{equation*}
    Moreover, by using \eqref{eq:app:lower-bound-moment},
    we have
    \[
        \P{\Omega_\ve}\ge 1 - \exp\left(-\frac{3\ve^2}8 \sum_{i=1}^N\lVert p_i\rVert_\beta^\beta\right)
        \ge 1 - \exp\left(-\frac{3\ve^2}8 NS^{-1/\alpha}\right),
    \]
    which completes the proof of the proposition.
\end{proof}

\subsection{Proof of Proposition~\ref{prop:maskgit-upper-bound}}\label{sec:proof-maskgit-upper-bound}
\begin{proof}
    Whether or not \eqref{eq:app:prop-aux} holds only depends on the actual values of $(x_i)_{i=1}^N$.
    Thus, we can define the set $\mathcal{Z}^\prime_\ve \subset \S^N$ such that
    \eqref{eq:app:prop-aux} is satisfied if and only if
    $(x_i)_{i=1}^N\in\mathcal{Z}^\prime_\ve$.
    Now, let $(z_i)_{i=1}^N \in \mathcal{Z}^\prime_\ve$.
    From \eqref{eq:maskgit-topk-eq} and \eqref{eq:app:prop-aux}, we have (recall $S=\lvert\S\rvert$)
    \begin{align}
        &\P{i_1^* = i_1,\ldots, i_k^*=i_k \mid (x_i)_{i=1}^N=(z_i)_{i=1}^N}
        \nonumber\\
        &=
        \prod_{\ell=1}^k
        \P{i_\ell^* = i_\ell \mid i_1^*=i_1,\ldots,i_{\ell-1}^*=i_{\ell-1}, (x_i)_{i=1}^N=(z_i)_{i=1}^N}
        \nonumber\\
        &= \prod_{\ell=1}^k
        \frac{p_{i_\ell}(x_{i_\ell})^{1/\alpha}}{\sum_{i\in [N]\setminus I_{\ell-1}} p_i(x_i)^{1/\alpha}}
        \nonumber\\
        &<\prod_{\ell=1}^k \frac{p_{i_\ell}(x_{i_\ell})^{1/\alpha}}{
            \left(1-\ve-\frac{(k-1)S^{1/\alpha}}{N-k+1}\right) \sum_{i\in[N]\setminus I_{\ell-1}}
        \lVert p_i\rVert_\beta^\beta}
        \nonumber\\
        &=
        \left(1-\ve-\frac{(k-1)S^{1/\alpha}}{N-k+1}\right)^{-k}
        \prod_{\ell=1}^k \frac{p_{i_\ell}(x_{i_\ell})^{1/\alpha}}{ \sum_{i\in[N]\setminus I_{\ell-1}}
        \lVert p_i\rVert_\beta^\beta}.
        \nonumber 
    \end{align}
    We thus have $z\in\mathcal{Z}_\ve$, which implies $\mathcal{Z}_\ve^\prime\subset\mathcal{Z}_\ve$.
    Therefore,
    \[
        \P{(x_i)_{i=1}^N \in \mathcal{Z}_\ve}\ge \P{(x_i)_{i=1}^N \in \mathcal{Z}^\prime_\ve}\ge 
        1 - \exp\left(
            -\frac38 \ve^2 N\lvert \S\rvert^{-1/\alpha}
        \right)
    \]
    follows from Proposition~\ref{prop:aux} and the definition of $\mathcal{Z}^\prime_\ve$.
\end{proof}

\subsection{Proof of Proposition~\ref{prop:moment-exact-probabilities}}\label{sec:proof-moment-exact}
\begin{proof}
    From the independence of $y_i$ and $(j_\ell^*)_{\ell=1}^k$ in Setting~\ref{sett:moment},
    we have
    \begin{align*}
        &\P{j_1^*=i_1,\ldots,j_k^*=i_k,
            (y_{j_\ell^*})_{\ell=1}^k = (z_{i_\ell})_{\ell=1}^k}\\
        &=\P{j_1^*=i_1,\ldots,j_k^*=i_k}\P{(y_{i_\ell})_{\ell=1}^k = (z_{i_\ell})_{\ell=1}^k \mid j_1^*=i_1,\ldots,j_k^*=i_k}
        \\
        &=\P{j_1^*=i_1,\ldots,j_k^*=i_k}\prod_{\ell=1}^k\P{y_{i_\ell}=z_{i_\ell}}\\
        &=\P{j_1^*=i_1,\ldots,j_k^*=i_k}\prod_{\ell=1}^k\frac{p_{i_\ell}(z_{i_\ell})^\beta}{\lVert p_{i_\ell}\rVert_\beta^\beta}.
    \end{align*}
    Now, we have
    \[
        \P{j_1^*=i_1,\ldots,j_k^*=i_k}
        = \prod_{\ell=1}^{k} \frac{\lVert p_{i_\ell}\rVert_\beta^\beta}{\sum_{i\in[N]\setminus
        I_{\ell-1}}\lVert p_i\rVert^\beta_\beta}
    \]
    by letting $\mu_i=\log\lVert \mu_i\rVert_\beta^\beta$ in Proposition~\ref{thm:gumbel-topk}.
    We obtain \eqref{eq:app:moment-joint} through these two identities.
\end{proof}

\subsection{Proof of Theorem~\ref{thm:main-formal}}\label{sec:proof-main}
\begin{proof}
    Let $\bm{z}\in\S^N$.
    By introducing extra variables, let us define
    \begin{align*}
        \tilde{p}_\mathrm{MaskGIT}(\bm{i}, \bm{z})&:=
        \P{i_1^*=i_1,\ldots,i_k^*=i_k,
            (x_i)_{i=1}^N=(z_i)_{i=1}^N
            },\\
        \tilde{p}_\mathrm{moment}(\bm{i}, \bm{z})
        &:=
        \P{j_1^*=i_1,\ldots,j_k^*=i_k,
            (y_{i_\ell})_{\ell=1}^k = (z_{i_\ell})_{\ell=1}^k,
            (x_i)_{i\in[N]\setminus I_k} = (z_i)_{i\in[N]\setminus I_k}
            },
    \end{align*}
    where we suppose $x_i$ and $y_i$ are sampled independently.
    These define probability distributions over $[N]^k\times\S^N$.
    From the independence and Proposition~\ref{prop:moment-exact-probabilities}, we have
    \begin{align}
        \tilde{p}_\mathrm{moment}(\bm{i}, \bm{z})
        &= p_\mathrm{moment}(\bm{i}, (z_{i_\ell})_{\ell=1}^k) \P{(x_i)_{i\in[N]\setminus I_k} = (z_i)_{i\in[N]\setminus I_k}}
        \nonumber\\
        &= \prod_{\ell=1}^{k} \frac{p_{i_\ell}(z_{i_\ell})^\beta}{\sum_{i\in[N]\setminus
        I_{\ell-1}}\lVert p_i\rVert^\beta_\beta}
        \prod_{i\in[N]\setminus I_k}p_i(z_i).
        \label{eq:app:moment-additional-var}
    \end{align}
    Suppose $0\le \ve\le 1$, $\ve+\frac{(k-1)S^{1/\alpha}}{N-k+1}<1$
    and recall the set $\mathcal{Z}_\ve$ defined in Proposition~\ref{prop:maskgit-upper-bound}.
    For $\bm{z}\in\mathcal{Z}_\ve$.
    From the definition of $\mathcal{Z}_\ve$, we have  (recall $S=\lvert\S\rvert$)
    \begin{align}
        \tilde{p}_\mathrm{MaskGIT}(\bm{i}, \bm{z})
        &= \P{i_1^*=i_1,\ldots, i_{k}^* = i_{k}
            \mid (x_i)_{i=1}^N = (z_i)_{i=1}^N}
            \P{(x_i)_{i=1}^N = (z_i)_{i=1}^N}\nonumber \\
        &< \left(1-\ve-\frac{(k-1)S^{1/\alpha}}{N-k+1}\right)^{-k}
        \prod_{\ell=1}^{k} \frac{p_{i_\ell}(z_{i_\ell})^{1/\alpha}}{\sum_{i\in[N]\setminus
        I_{\ell-1}}\lVert p_i\rVert^\beta_\beta}
        \prod_{i=1}^N p_i(z_i) \nonumber\\
        &= \left(1-\ve-\frac{(k-1)S^{1/\alpha}}{N-k+1}\right)^{-k}
        \prod_{\ell=1}^{k} \frac{p_{i_\ell}(z_{i_\ell})^{1/\alpha}\cdot p_{i_\ell}(z_{i_\ell})}{\sum_{i\in[N]\setminus
        I_{\ell-1}}\lVert p_i\rVert^\beta_\beta}
        \prod_{i\in[N]\setminus I_k} p_i(z_i)\nonumber\\
        &= \left(1-\ve-\frac{(k-1)S^{1/\alpha}}{N-k+1}\right)^{-k}
        \tilde{p}_\mathrm{moment}(\bm{i}, \bm{z}),
    \end{align}
    where we have used \eqref{eq:app:moment-additional-var} in the last equality.
    Let us next bound the total variation distance between $\tilde{p}_\mathrm{MaskGIT}$
    and $\tilde{p}_\mathrm{moment}$.
    Let us denote $(a)_+:=\max\{0, a\}$.
    In general, for probability distributions $p$ and $q$ over the same finite set $\X$,
    we have
    \[
        \sum_{x\in\X}(p(x)-q(x))_+ - \sum_{x\in\X}(q(x)-p(x))_+ =
        \sum_{x\in\X}(p(x)-q(x))=1-1=0.
    \]
    Thus, for the total variation distance, we have
    \begin{align*}
        d_\mathrm{TV}(p, q)
        &=\frac12\sum_{x\in\X}\lvert p(x) - q(x)\rvert\\
        &=\frac12\left(\sum_{x\in\X}(p(x) - q(x))_+ +
        \sum_{x\in\X}(q(x) - p(x))_+
        \right)
        = \sum_{x\in\X}(p(x) - q(x))_+.
    \end{align*}
    By using this, we have
    \begin{align}
        &d_\mathrm{TV}(\tilde{p}_\mathrm{moment}, \tilde{p}_\mathrm{MaskGIT}) \nonumber\\
        &=
        \sum_{\bm{i}\in [N]^k}\sum_{\bm{z}\in \S^N}
        (\tilde{p}_\mathrm{MaskGIT}(\bm{i}, \bm{z}) - \tilde{p}_\mathrm{moment}(\bm{i}, \bm{z}))_+ \nonumber\\
        &= \sum_{\bm{i}\in [N]^k}\sum_{\bm{z}\in \mathcal{Z}_\ve}
        (\tilde{p}_\mathrm{MaskGIT}(\bm{i}, \bm{z}) - \tilde{p}_\mathrm{moment}(\bm{i}, \bm{z}))_+
        + \sum_{\bm{i}\in [N]^k}\sum_{\bm{z}\not\in \mathcal{Z}_\ve}
        (\tilde{p}_\mathrm{MaskGIT}(\bm{i}, \bm{z}) - \tilde{p}_\mathrm{moment}(\bm{i}, \bm{z}))_+ \nonumber\\
        &< \sum_{\bm{i}\in [N]^k}\sum_{\bm{z}\in \mathcal{Z}_\ve}
        \left(\left(1-\ve-\frac{(k-1)S^{1/\alpha}}{N-k+1}\right)^{-k} -1\right)
        \tilde{p}_\mathrm{moment}(\bm{i}, \bm{z})
        + \sum_{\bm{i}\in [N]^k}\sum_{\bm{z}\not\in \mathcal{Z}_\ve} \tilde{p}_\mathrm{MaskGIT}(\bm{i}, \bm{z}) \nonumber\\
        &\le \left(\left(1-\ve-\frac{(k-1)S^{1/\alpha}}{N-k+1}\right)^{-k} - 1\right)
        + \P{(x_i)_{i=1}^N \not\in \mathcal{Z}_\ve}.
    \end{align}
    By applying Proposition~\ref{prop:maskgit-upper-bound},
    we obtain
    \[
        d_\mathrm{TV}(\tilde{p}_\mathrm{moment}, \tilde{p}_\mathrm{MaskGIT})
        \le
        \left(\left(1-\ve-\frac{(k-1)S^{1/\alpha}}{N-k+1}\right)^{-k}
            -1\right)
        + \exp\left(
            -\frac38 \ve^2 NS^{-1/\alpha}
        \right).
    \]
    For $0< \delta \le \frac1{2k}$,
    we have
    \[
        (1-\delta)^{-1}
        =1+\delta\sum_{n=0}^\infty\delta^n
        \le 1+\frac\delta{1-\frac1{2k}}
        =1 + \frac{2k\delta}{2k-1}.
    \]
    Since $(k-1)\frac{2k\delta}{2k-1}\le \frac{k-1}{2k-1}\le 1/2$,
    we have
    \begin{align*}
        (1-\delta)^{-k}
        \le \left(1 + \frac{2k\delta}{2k-1}\right)^k
        &= 1 + \sum_{n=1}^k \binom{k}{n} \left(\frac{2k\delta}{2k-1}\right)^n\\
        &\le 1 + k\cdot\frac{2k\delta}{2k-1}
        \sum_{n=1}^k \frac{(k-1)^{n-1}}{n!} \left(\frac{2k\delta}{2k-1}\right)^{n-1}\\
        &\le 1 + k\cdot\frac{2k\delta}{2k-1}
        \sum_{m=0}^\infty \frac1{m!} \left((k-1)\frac{2k\delta}{2k-1}\right)^m\\
        &= 1 + \frac{2k^2\sqrt{e}}{2k-1}\delta < 1+4k\delta.
    \end{align*}
    Thus, assuming $\ve+\frac{(k-1)S^{1/\alpha}}{N-k+1}<\frac1{4k}$,
    we have
    \begin{equation}
        d_\mathrm{TV}(\tilde{p}_\mathrm{moment}, \tilde{p}_\mathrm{MaskGIT})
        \le
        4k\left(\ve+\frac{(k-1)S^{1/\alpha}}{N-k+1}\right)
        + \exp\left(
            -\frac38 \ve^2 NS^{-1/\alpha}
        \right).
        \label{eq:ub-total-var}
    \end{equation}
    Since the total variation distance is always bounded by $1$,
    actually
    \eqref{eq:ub-total-var} holds without the posed assumptions on $\ve+\frac{(k-1)S^{1/\alpha}}{N-k+1}$.

    Note that, when $N<k^2\lvert\S\rvert^{1/\alpha}$,
    the upper bound of $d_\mathrm{TV}(p_\mathrm{moment}, p_\mathrm{MaskGIT})$ becomes larger than $1$,
    which holds trivially true since $d_\mathrm{TV}$ is bounded above by $1$.
    So, it suffices to prove the desired inequality with $\log$ instead of $\log^+$ under the assumption $N\ge k^2\lvert\S\rvert^{1/\alpha}$.
    Under this, by letting
    $\ve = \sqrt{\frac83\frac{S^{1/\alpha}}{N}\cdot\frac12\log\frac{N}{k^2S^{1/\alpha}}}$
    in \eqref{eq:ub-total-var},
    we have
    \begin{align}
        &d_\mathrm{TV}(\tilde{p}_\mathrm{moment}, \tilde{p}_\mathrm{MaskGIT})
        \nonumber\\
        &\le
        4k\sqrt{\frac43\frac{S^{1/\alpha}}{N}\log\frac{N}{k^2S^{1/\alpha}}}
        + 
        \frac{4k(k-1)S^{1/\alpha}}{N-k+1}
        + \sqrt{\frac{k^2S^{1/\alpha}}{N}}.
        \label{eq:tv-bound-mousukoshi}
    \end{align}
    We need to make sure $\ve\le1$, but the above bound is valid even when $\ve>1$,
    again because of the boundedness of $d_\mathrm{TV}$.
    From $N\ge k^2\lvert\S\rvert^{1/\alpha}\ge k^2$, we also have
    \[
        \frac{4k(k-1)S^{1/\alpha}}{N-k+1}
        \le 4S^{1/\alpha}\cdot\frac{k^2-k}{N-k}
        \le 4S^{1/\alpha}\cdot\frac{k^2}{N}
        =\frac{4k^2S^{1/\alpha}}N
        \le 4\sqrt{\frac{k^2S^{1/\alpha}}N}.
    \]
    By applying this to \eqref{eq:tv-bound-mousukoshi}, we obtain
    \begin{align}
        d_\mathrm{TV}(\tilde{p}_\mathrm{moment}, \tilde{p}_\mathrm{MaskGIT})
        &\le
        4k\sqrt{\frac43\frac{S^{1/\alpha}}{N}\log\frac{N}{k^2S^{1/\alpha}}}
        + 5\sqrt{\frac{k^2S^{1/\alpha}}{N}}\nonumber\\
        &=\sqrt{\frac{k^2S^{1/\alpha}}{N}}\left(
            5 + \sqrt{\frac{64}{3}\log\frac{N}{k^2S^{1/\alpha}}}\right)\nonumber\\
        &\le 5\sqrt{\frac{k^2S^{1/\alpha}}{N}}\left(
            1 + \sqrt{\log\frac{N}{k^2S^{1/\alpha}}}\right).
            \label{eq:tv-final-ub}
    \end{align}
    Finally, by denoting $I = \{i_1,\ldots,i_k\}$,
    we have
    \begin{align*}
        &d_\mathrm{TV}({p}_\mathrm{moment}, {p}_\mathrm{MaskGIT})\\
        &= \frac12\sum_{\bm{i}\in [N]^k}\sum_{(z_i)_{i\in I}\in \S^k}
        \lvert {p}_\mathrm{MaskGIT}(\bm{i}, (z_i)_{i\in I}) - {p}_\mathrm{moment}(\bm{i}, (z_i)_{i\in I})\rvert\\
        &= \frac12\sum_{\bm{i}\in [N]^k}\sum_{(z_i)_{i\in I}\in \S^k}
        \left\lvert \sum_{(z_j)_{j\not\in I}\in\S^{N-k}}
        ({p}_\mathrm{MaskGIT}(\bm{i}, (z_i)_{i\in I}, (z_j)_{j\not\in I})
        - {p}_\mathrm{moment}(\bm{i}, (z_i)_{i\in I}, (z_j)_{j\not\in I}))\right\rvert\\
        &\le\frac12\sum_{\bm{i}\in [N]^k}\sum_{(z_i)_{i\in I}\in \S^k} \sum_{(z_j)_{j\not\in I}\in\S^{N-k}}
        \lvert
        {p}_\mathrm{MaskGIT}(\bm{i}, (z_i)_{i\in I}, (z_j)_{j\not\in I})
        - {p}_\mathrm{moment}(\bm{i}, (z_i)_{i\in I}, (z_j)_{j\not\in I}))\rvert\\
        &=d_\mathrm{TV}(\tilde{p}_\mathrm{moment}, \tilde{p}_\mathrm{MaskGIT}).
    \end{align*}
    By combining this with \eqref{eq:tv-final-ub},
    we obtain the desired conclusion.
\end{proof}

\section{Additional experimental details}\label{app:experiment}
\subsection{Sampling schedule}\label{sec:schedule}
Let $D$ be the number of positions (so $\bm{x}\in\S^D$)
and $N$ be the number of total sampling steps.
Let $J_n\subset[D]$ be the (random) set of indices that are open after the $n$-th step,
i.e., $\emptyset =J_0\subset J_1\subset\cdots \subset J_N=[D]$.
Let $I_n:=J_n\setminus J_{n-1}$ for $n=1,\ldots,N$,
which is the (random) set of indices we unmask at the $n$-th step.
In all the experiments,
the cardinalities $\lvert J_n\rvert$ and $\lvert I_n\rvert\ (=\lvert J_n\rvert-\lvert J_{n-1}\rvert)$
are predetermined by the unmasking size schedule such as:
\begin{itemize}
    \item Cosine schedule: $\lvert J_n\rvert=\text{round}(\cos(\frac\pi2 D(1-\frac{n}N)))$.
    \item Uniform schedule: $\lvert J_n\rvert=\text{round}(D\cdot \frac{n}N)$.
\end{itemize}
Here, ``round'' means integer rounding.
We adopted the cosine schedule for image and the uniform schedule for language.

Let us now consider specifically the $n$-th sampling step out of $N$ steps.
Given the denoising model $p$,
we use the marginal distributions $(p_{i|J_{n-1}})_{i\in [D]\setminus J_{n-1}}$ for this step.
Let $k$ be the number of indices to unmask in this step, determined by the unmasking size schedule.
Then, for the sampler \bt{MaskGIT}, we use $\text{OneRoundMaskGIT}((p_{i|J_{n-1}})_{i\in [D]\setminus J_{n-1}}, k, \alpha_n)$ from Algorithm~\ref{algo:maskgit-one-round}
to determine $I_n$ and $\bm{x}_{I_n}$.
Here, $\alpha_n$ is the Gumbel temperature for the $n$-th step, which is scheduled as
$\alpha_n=\alpha(1-n/N)$ following \citep{chang2022maskgit}, where $\alpha$ is the temperature parameter of the method presented in the figures (e.g., Figure~\ref{fig:mage}).
We use the same $\alpha_n$ for \bt{Moment} (Algorithm~\ref{algo:moment-one-round}) and its variants
given the parameter $\alpha$.
Note that, in the final step ($n=N$) of \bt{Moment} or other temperature-sampling methods,
we omit the sampling temperature (or take $\alpha_N\to\infty$), in order that it corresponds to the final step of \bt{MaskGIT}.

\subsection{Partial caching}\label{app:caching}
In the partial caching algorithm we described in Section~\ref{sec:caching},
we have a degree of freedom in dividing the selected index set $I$ into $A$ and $B$
(where we have$A\cup B=I$ and $A\cap B=\emptyset$).

Let us explain our implementation.
Let us use the notation of $J_n$ and $I_n$ introduced in the previous section.
In the $n$-th step,
suppose we decompose $I_n$ into $A_n$ and $B_n$,
where $A_n$ is the set of indices unmasked in the intermediate step of partial caching.
If we let $J_{n-1/2}:=J_{n-1}\cup A_n$ for $n\ge1$,
then we adopted the canonical extension of the scheduler in Appendix~\ref{sec:schedule}:
\begin{itemize}
    \item Cosine schedule: $\lvert J_{n-1/2}\rvert=\text{round}(\cos(\frac\pi2 D(1-\frac{n-1/2}N)))$.
    \item Uniform schedule: $\lvert J_{n-1/2}\rvert=\text{round}(D\cdot \frac{n-1/2}N)$.
\end{itemize}
Thus, the cardinality of $A_n$ was determined by $\lvert A_n\rvert=\lvert J_{n-1/2}\rvert-\lvert J_{n-1}\rvert$, depending on the sampling schedule we use.
Since each of our sampling algorithm outputs an ordering of masked indices
(from which we determine $I_n$),
we simply determine $A_n$ as the top-$k$ of the ordered indices ($k=\lvert A_n\rvert = \lvert J_{n-1/2}\rvert-\lvert J_{n-1}\rvert$).

\subsubsection{Computational efficiency of partial caching}
\label{sec:caching-compute}
The total cost of {\it attention computation}
in partial caching is $1+\lvert I\rvert/D$ times the original full attention computation (in terms of the notations in Section~\ref{sec:caching}).
However, it costs more computation in caching the vectors and other CPU/GPU operations.
Indeed, while caching shows some performance gain with A6000 latency in the language experiment
(Figure~\ref{fig:sdtt}, Right),
it vanishes when we use H100 GPU (Figure~\ref{fig:sdtt-omitted}, Right).
It would be caused by the faster attention computation of H100, reducing
its weight among the overall computational cost and making other computational overhead apparent.

\subsection{Additional details on image modeling experiments}
The MAGE ViT-B model \citep{li2023mage}, which we used in the experiments,
can be regarded as a masked diffusion model
on the space of a pretrained VQGAN tokenizers \citep{esser2021taming}.
It was trained on the ImageNet $256\times256$ dataset \citep{deng2009imagenet}.
The codebook size is given by $\lvert\S\rvert=1024$,
and the length of each token sequence (corresponding to a single image) is $D=256$.
Each experiment with MAGE was conducted with a single A6000 GPU with a minibatch size of 64.
Based on 50000 unconditional images generated from each sampler,
we measured FID and IS against the ImageNet dataset
by using \bt{torch-fidelity}\footnote{\url{https://github.com/toshas/torch-fidelity}.},
following the description of the repository of MAGE\footnote{\url{https://github.com/LTH14/mage}.}.

\subsection{Additional details on language modeling experiments}
We used the SDTT model \citep{deschenaux2024beyond}\footnote{It was loaded by 
\texttt{load\_small\_student(loss="kld", round=7)}
in the repository \url{https://github.com/jdeschena/sdtt}.},
which is a masked diffusion model over a GPT-2 tokenizers \citep{radford2019language}.
It was trained on the OpenWebText dataset \citep{gokaslan2019openwebtext}.
The codebook size is $\lvert\S\rvert = 50257$
and the token sequence length is given by $D=1024$.
Most experiments were conducted on a single H100 GPU,
while the preliminary experiments, the ones in Appendix~\ref{sec:numer-prec},
and the latency computation in Figure~\ref{fig:latency-perf}(Right) were conducted on a single A6000 GPU.
Each plot was computed by 1024 samples
generated with a minibatch size of $16$, using the following performance metrics.

\noindent\textbf{Generative Perplexity.}
It was measured against the GPT-2 large model \citep{radford2019language}
and averaged over 1024 samples.
We used the implementation of \citet{deschenaux2024beyond}.

\noindent\textbf{Entropy.}
Following the existing work \citep{gat2024discrete,zheng2024masked},
we measured the sentence-wise entropy for checking the diversity
of generated sentences.
In our implementation (following the description of \citep{zheng2024masked}),
for a sequence of tokens $\bm{x}=(x_1,\ldots,x_D)$,
we define the sentence entropy as
\[
    -\sum_{s\in\S\cap\bm{x}}\frac{\#\{i\in[D]\mid x_i = s\}}D \log \frac{\#\{i\in[D]\mid x_i = s\}}D,
\]
where $\S\cap\bm{x}$ is the set of tokens appearing in $\bm{x}$.
Its average over 1024 samples was plotted.

\begin{figure}[t]
    \centering
    \subfigure{
        \includegraphics[width=0.475\linewidth]{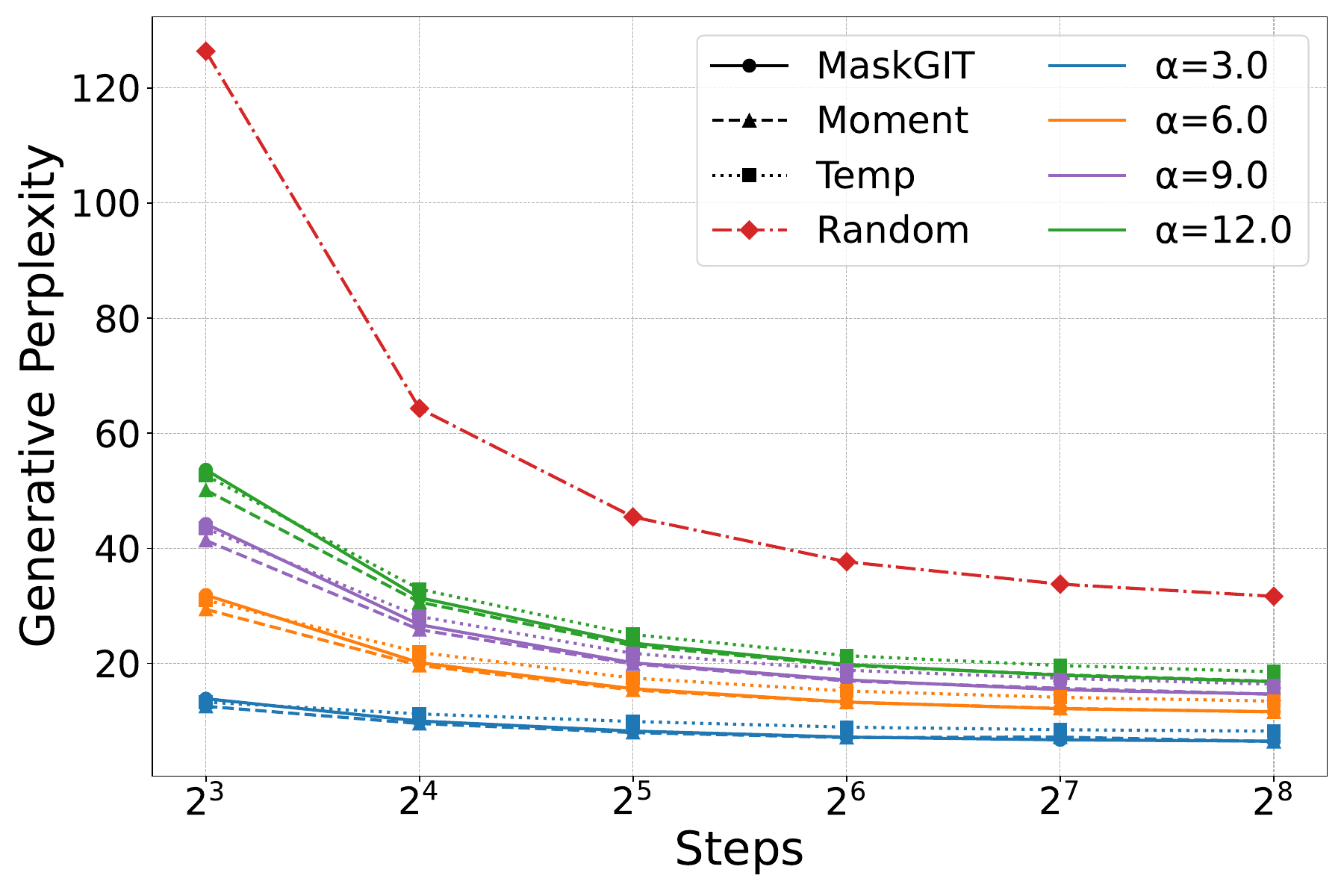}
    }
    \subfigure{
        \includegraphics[width=0.475\linewidth]{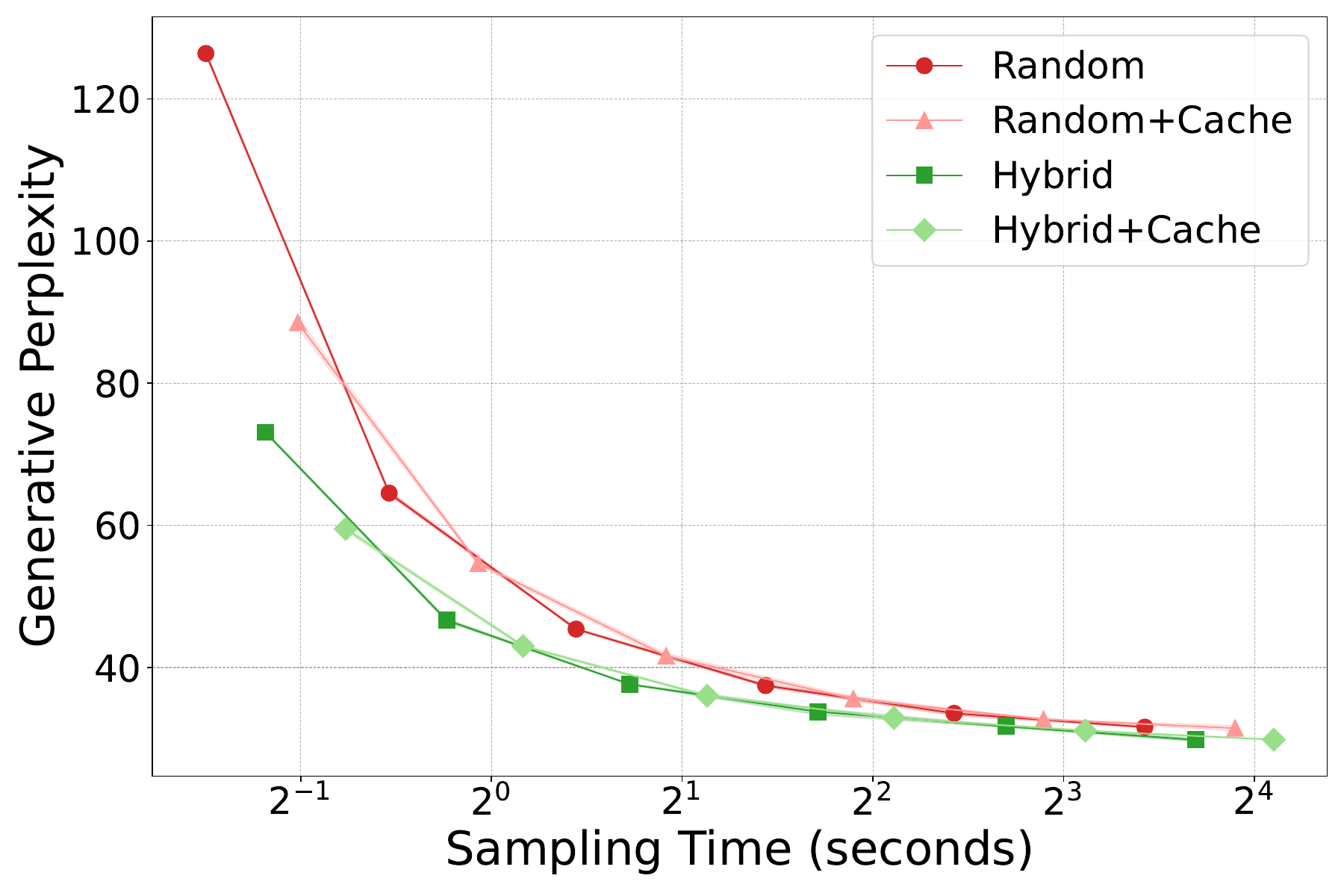}
    }
    \vspace{-4mm}
    \caption{
        Additional experimental results.
        (\textit{Left}) Generative Perplexity of various samplers with temperature sampling.
        (\textit{Right}) Generative Perplexity of our proposed samplers against sampling time per batch on H100 GPU.
    }
    \label{fig:sdtt-omitted}
\end{figure}

Figure~\ref{fig:sdtt-omitted}(Left) shows the omitted results for Generative Perplexity.
While the methods with lower temperature apparently attain better the generation quality,
lowering the temperature extremely harms the diversity in reality (Figure~\ref{fig:sdtt}, Left).

\subsubsection{Numerical precision}\label{sec:numer-prec}
\citet{zheng2024masked} pointed out that the sampling from masked diffusion models
with low numerical precision (32-bit)
can lead to errors in categorical sampling,
showing lower (better) Generative Perplexity at the cost of lower (worse) Entropy.
Because of this, they suggest using 64-bit computation for a fair evaluation of masked diffusion models.

However, as they also note in Section~J.2.2 of their paper,
it is primarily because the lower numerical precision results in biased positional selection
rather than the sampling distribution shift at each position.
In our case,
since we fix the token positions in most samplers (except \bt{MaskGIT}),
we do not suffer from this problem though we use 32-bit precision.
Indeed, Table~\ref{tab:precision} shows that,
while the vanilla sampler in discrete diffusion suffer from the difference in numerical precision,
\bt{Fixed}, which corresponds to \bt{Random} in the main text, exhibits similar results
in both precision settings.

\begin{table}[h]
    \centering
    \caption{Comparison of different numerical precision in \bt{Vanilla}
    and \bt{Fixed} samplers.
    \bt{Vanilla} is a standard sampler for discrete diffusion,
    where it independently determines whether or not unmasking a certain position.
    \bt{Fixed} is a sampler that pre-determines the number of unmasked positions at each step,
    and it determines which positions to unmask uniformly at random.
    Both follows the uniform schedule (Appendix~\ref{sec:schedule}) in expectation.
    } 
    \label{tab:precision}
    \vspace{2mm}
    \begin{NiceTabular}{cccccccc} 
    \toprule 
    \Block{2-1}{Sampler} & \Block{2-1}{Precision} & \multicolumn{2}{c}{8 steps}  & \multicolumn{2}{c}{32 steps}
    & \multicolumn{2}{c}{128 steps} \\ 
    \cmidrule(lr){3-4} \cmidrule(lr){5-6}
    \cmidrule(lr){7-8}
     & & Gen.~PPL  & Entropy & Gen.~PPL  & Entropy & Gen.~PPL & Entropy  \\
    \midrule
    \multirow{2}{*}{\bt{Vanilla}} & 32-bit
    & 125.62 & 5.40 & 41.91 & 5.31 & 27.63 & 5.17 \\
    & 64-bit & 137.95 & 5.42 & 46.57 & 5.35 & 33.80 & 5.28 \\
    \midrule
    \bt{Fixed} & 32-bit & 131.01 & 5.41 & 45.10 & 5.32 & 33.15 & 5.26  \\
    (= \bt{Random})& 64-bit & 130.76 & 5.41 & 46.66 & 5.35 & 34.29 & 5.29  \\
    \bottomrule
    \end{NiceTabular}
\end{table}

\subsubsection{\bt{Hybrid} algorithm}\label{sec:hybrid-details}
Let us explain the details of the \bt{Hybrid} sampler in the language experiments,
where we merged the \bt{Halton} and \bt{U-Moment} samplers.
Let us consider the $n$-th sampling step out of $N$ total steps
and let $J_{n-1}$ be the set of indices already unmasked at this stage.
The ordering of first $k = \lvert I_n\rvert$ (where $I_n=J_n\setminus J_{n-1}$ is from Appendix~\ref{sec:schedule})
positions from each sampler is given as follows:
\begin{itemize}
    \item \bt{Halton}:
    We consider the one-dimensional Halton sequence of indices (with base $2$),
    i.e., rearrangement of $[D]$, and let
    $\bm{i}=(i_1, \ldots, i_k)$ be its first $k$ entries that are also in $[D]\setminus J_{n-1}$.
    \item \bt{U-Moment}:
    As in Algorithm~\ref{algo:moment-one-round}, we define the ordering $\bm{j} = (j_1,\ldots,j_k)$ by
    \[
        \bm{j}=\argtop{k}_{j\in[D]\setminus J_{n-1}}\left\{\log\sum_{x\in\S}p_{j|J_{n-1}}(x|\bm{x}_{J_{n-1}})^\beta
        +\xi_j\right\},
    \]
    where the exponent $\beta=1+1/\alpha$ is determined by the temperature parameter $\alpha$,
    and $\xi_j$ is an independently sampled standard Gumbel noise for each $j\in [D]\setminus J_{n-1}$. 
\end{itemize}
We then merge $\bm{i}$ and $\bm{j}$ into $\bm{k}$ as described in Section~\ref{sec:explo} to obtain
an ordering for \bt{Hybrid},
where merging parameter $m=m_n$, controlling how many indices we take from $\bm{i}$,
is scheduled as $m_n = \mathrm{round}((1 - n/N)\lvert I_n\rvert)$.
Intuitively, it means that we basically start from \bt{Halton},
whose exploration works at the initial stages,
and gradually move to \bt{U-Moment}, which conducts exploitation-based index selection.
In the implementation of \bt{Hybrid+Cache},
we just apply the caching procedure as explained in Appendix~\ref{app:caching} to the above merged ordering $\bm{k}$.


\section{Further experimental results}
\subsection{Unconditional image generation with MAGE}
In addition to the raw data presented in Figure~\ref{fig:mage},
we give the detailed FID values of \bt{Random}, \bt{MaskGIT}, \bt{Moment}, \bt{Temp}
at each temperature in Table~\ref{tab:fid-mage}.
Based on these numbers, we also present the relative differences between \bt{MaskGIT}
and the other samplers in Table~\ref{tab:error-mage} in the main body, which validates our theory on
\bt{Moment} approximating \bt{MaskGIT} with the same temperature.
\begin{table}[h]
\centering
{
\mtabcaption{B}{ImageNet FIDs for various methods in unconditional experiment with MAGE.}
\vspace{2mm}
\small
\begin{NiceTabular}{cccccccccccccc}
\toprule
\Block{2-1}{Steps} & \Block{2-1}{\bt{Random}} & \multicolumn{3}{c}{\bt{MaskGIT}} & \multicolumn{3}{c}{\bt{Moment}} & \multicolumn{3}{c}{\bt{Temp}} & \multicolumn{3}{c}{\rev{\bt{U-Moment}}} \\
\cmidrule(lr){3-5} \cmidrule(lr){6-8} \cmidrule(lr){9-11} \cmidrule(lr){12-14}
 & & $\alpha=3.0$ & 6.0 & 12.0 & 3.0 & 6.0 & 12.0 & 3.0 & 6.0 & 12.0 & \rev{3.0} & \rev{6.0} & \rev{12.0} \\
\midrule
8 & 49.86 & 18.95 & 19.21 & 27.18 & 19.47 & 17.42 & 24.26 & 20.83 & 18.74 & 25.26 & \rev{48.42} & \rev{48.75} & \rev{48.92} \\
16 & 31.85 & 19.36 & 11.29 & 13.32 & 20.71 & 11.52 & 12.73 & 19.86 & 12.09 & 13.25 & \rev{29.41} & \rev{30.21} & \rev{30.99} \\
32 & 22.55 & 24.61 & 11.45 & 9.84 & 25.76 & 11.77 & 9.73 & 23.83 & 11.73 & 10.00 & \rev{20.71} & \rev{21.53} & \rev{22.24} \\
64 & 18.40 & 27.23 & 12.36 & 9.03 & 27.80 & 12.77 & 9.13 & 25.32 & 12.33 & 9.27 & \rev{16.79} & \rev{17.52} & \rev{18.29} \\
\bottomrule
\end{NiceTabular}
\label{tab:fid-mage}
}
\end{table}

\subsection{Class-conditional image generation with MaskGIT-PyTorch}
To confirm the approximation ability of \bt{Moment} in conditional image generation,
we used the MaskGIT-PyTorch \citep{besnier2023pytorch} model pretrained on ImageNet $256\times256$.
We adopted the arccos sampling scheduler (see also Appendix~\ref{sec:schedule}) for unmasking size,
and the Gumbel temperature schedule followed the default implementation (same as that in Appendix~\ref{sec:schedule}).
The classifier-free guidance coefficient was set to $3.0$ throughout the experiment.

\begin{figure}[htbp]
    \centering
    \includegraphics[width=0.6\linewidth]{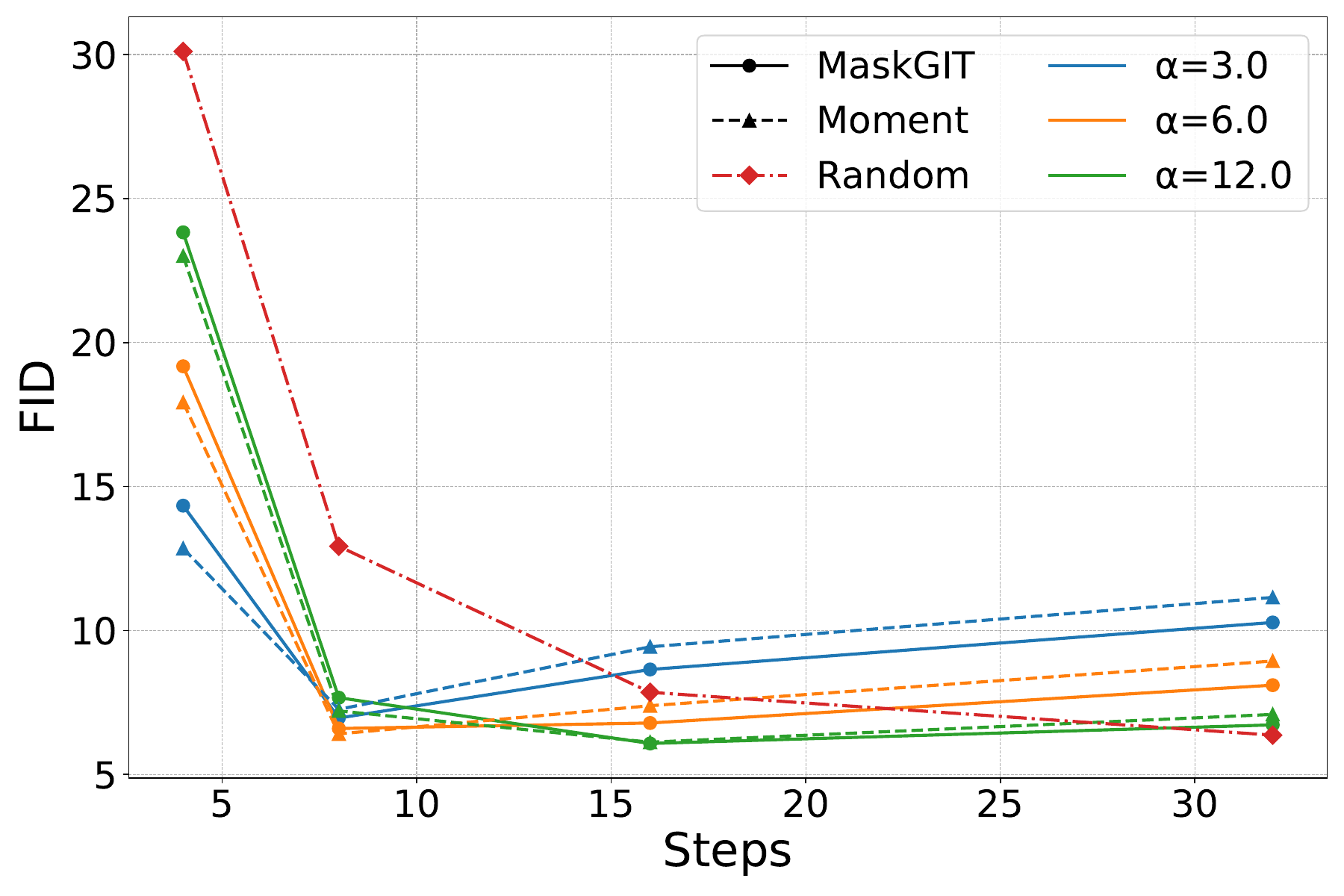}
    \vspace{-4mm}
    \mfigcaption{C}{FID against number of steps for various samplers with MaskGIT-PyTorch.
    FID was computed with 50,000 (50 per class) class-conditional generation results.}
    \label{fig:maskgit}
    \vspace{2mm}
    \subfigure{
        \includegraphics[width=0.475\linewidth]{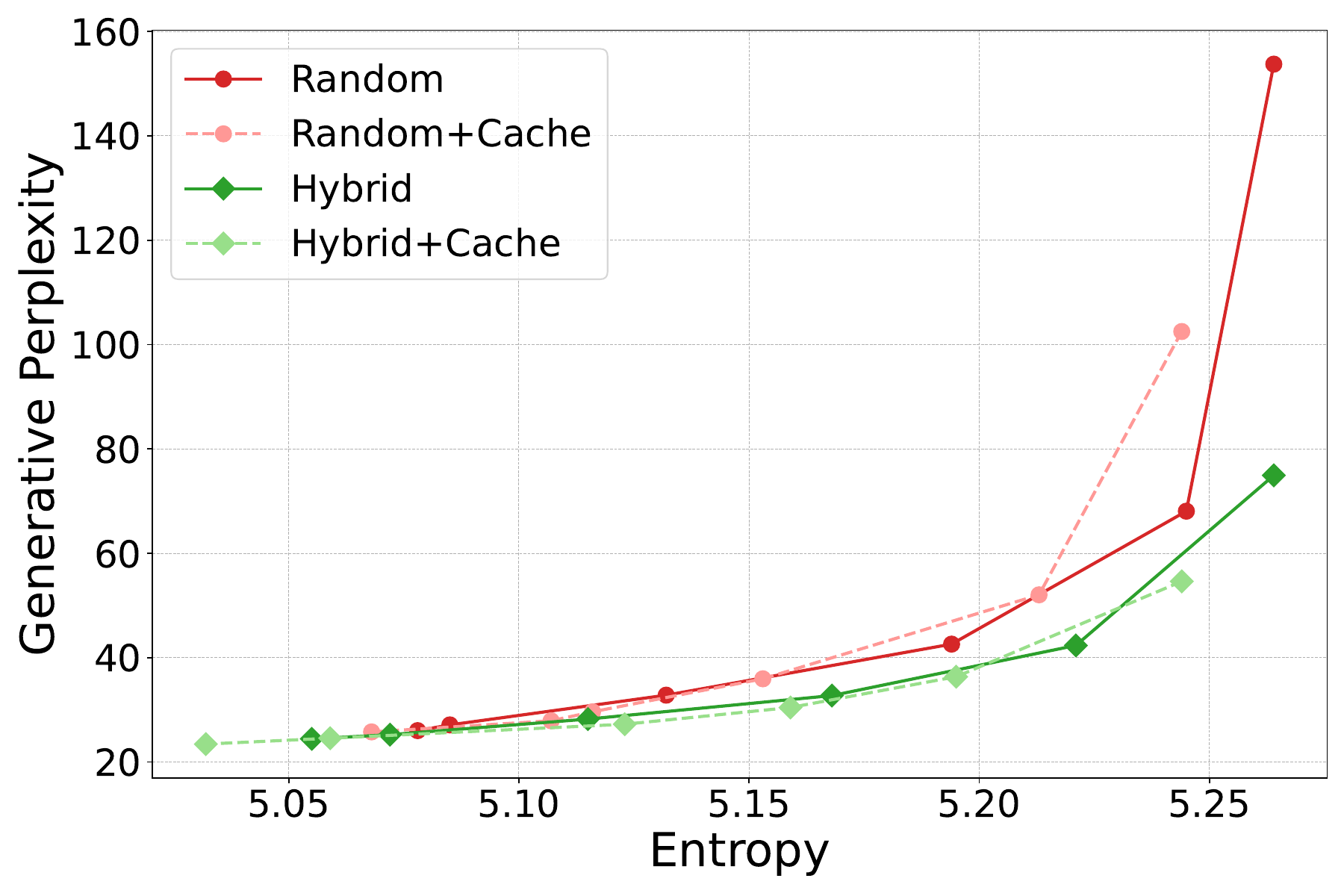}
    }
    \subfigure{
        \includegraphics[width=0.475\linewidth]{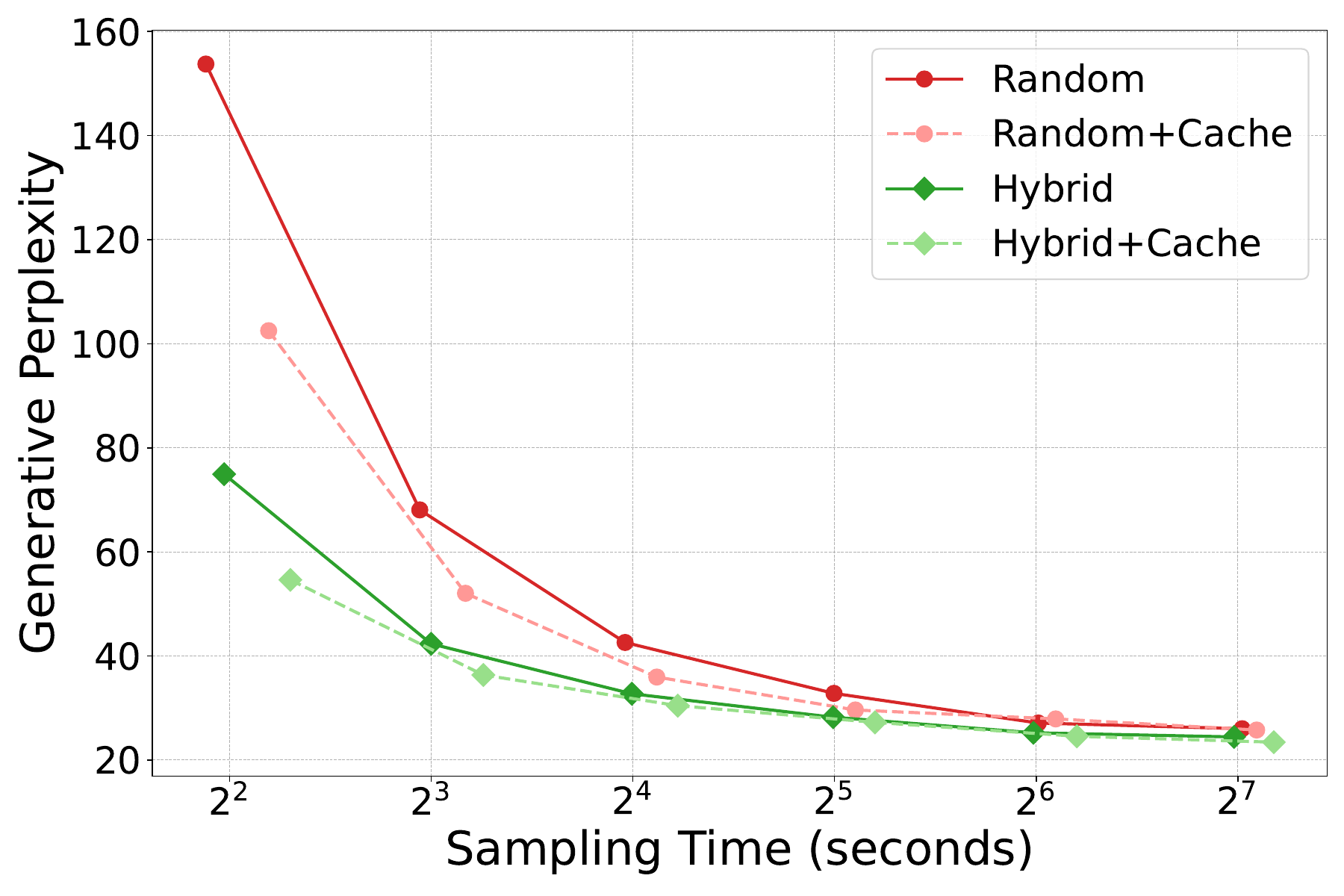}
    }
    \vspace{-4mm}
    \mfigcaption{D}{
        Experiments with SDTT-large. Generative Perplexity was measured by Llama3-8B.
    (\textit{Left}) Trade-off between Generative Perplexity and Entropy, 
    (\textit{Right}) Generative Perplexity against Sampling Time on A6000.
    }
    \label{fig:large}
    \vspace{2mm}
    \centering
    \includegraphics[width=0.6\linewidth]{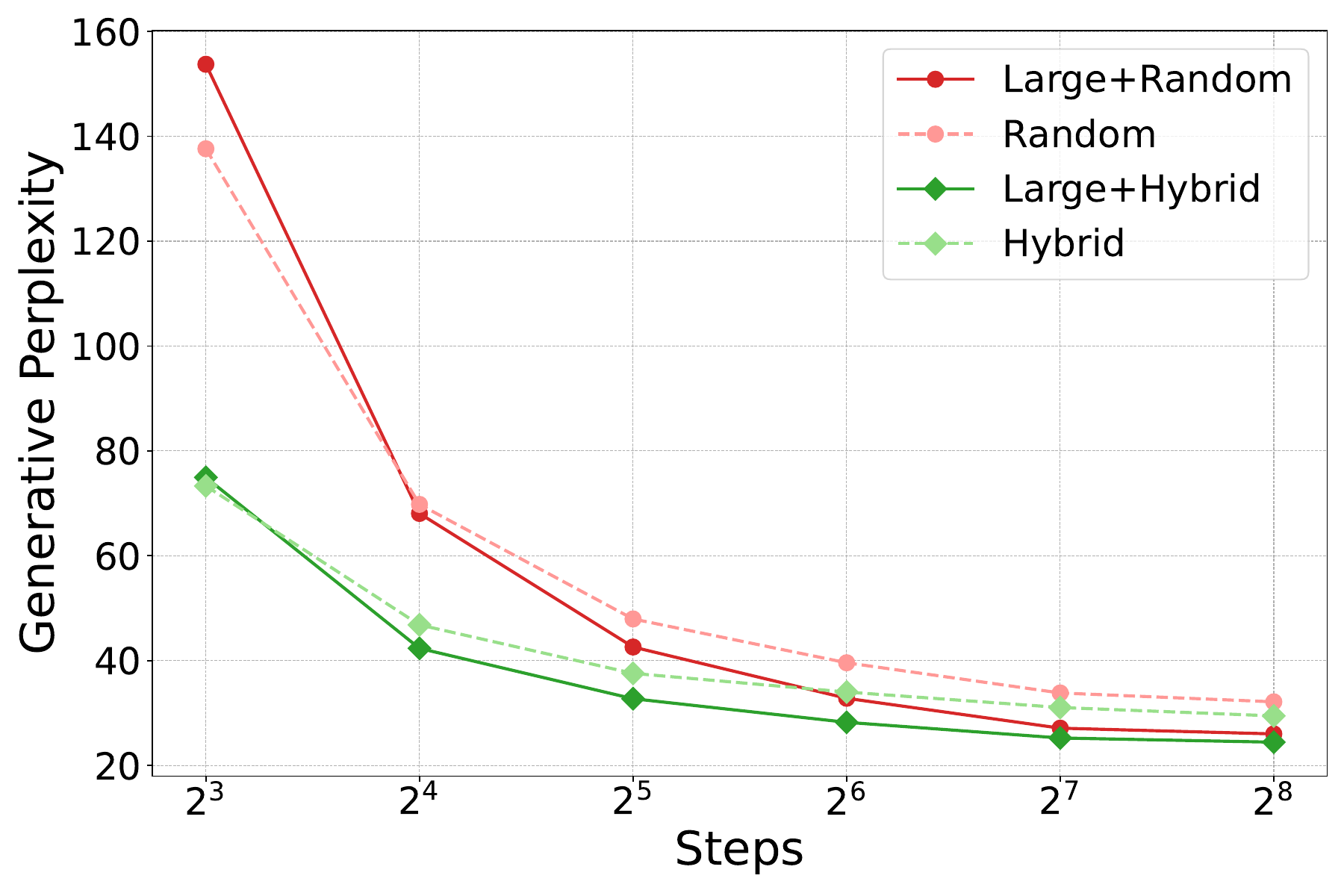}
    \vspace{-4mm}
    \mfigcaption{E}{Efficiency comparison of SDTT-large and SDTT with \bt{Random} and \bt{Hybrid} samplers.
    `Large+' means SDTT-large; otherwise it shows the results of SDTT.
    Generative Perplexity was measured by Llama3-8B for both models.}
    \label{fig:large-small-sdtt}
\end{figure}

Table~\ref{tab:fid-cond-maskgit} shows the FID values of several samplers and Table~\ref{tab:error-cond-maskgit}
shows their relative differences to \bt{MaskGIT} samplers
(corresponding to Table~\ref{tab:fid-mage} and Table~\ref{tab:error-mage}
in the unconditional experiment, respectively).
We can confirm again that \bt{Moment} samplers consistently achieve the FID within 10\%
with \bt{MaskGIT} of corresponding temperature.
This approximation can also be visibly confirmed by the plot in Figure~\ref{fig:maskgit}.

\rev{In the experiment with MaskGIT-PyTorch, while \bt{U-Moment} improves when increasing steps, it does not necessarily outperforms \bt{Random} at fewer steps, in contrast to the experiment with MAGE Table~\ref{tab:fid-mage}.
We attribute this to the use of classifier-free guidance, which extrapolates logits between conditional and unconditional predictions; since this extrapolation does not yield normalized probabilities, the assumptions underlying Proposition~\ref{prop:one-by-one-cts} no longer hold in this setting.}


\begin{table}[htbp]
\centering
{
\mtabcaption{C}{FIDs for various methods and steps in class-conditional experiment with MaskGIT-PyTorch.
}
\vspace{2mm}
\small
\begin{NiceTabular}{cccccccccccccc}
\toprule
\Block{2-1}{Steps} & \Block{2-1}{\bt{Random}} & \multicolumn{3}{c}{\bt{MaskGIT}} & \multicolumn{3}{c}{\bt{Moment}} & \multicolumn{3}{c}{\bt{Temp}} & \multicolumn{3}{c}{\rev{\bt{U-Moment}}} \\
\cmidrule(lr){3-5} \cmidrule(lr){6-8} \cmidrule(lr){9-11} \cmidrule(lr){12-14}
 & & $\alpha=3.0$ & 6.0 & 12.0 & 3.0 & 6.0 & 12.0 & 3.0 & 6.0 & 12.0 & \rev{3.0} & \rev{6.0} & \rev{12.0} \\
\midrule
4 & 30.11 & 14.34 & 19.17 & 23.83 & 12.85 & 17.92 & 23.01 & 17.65 & 22.82 & 25.97 & \rev{31.96} & \rev{31.89} & \rev{31.65} \\
8 & 12.92 & 6.96 & 6.60 & 7.67 & 7.26 & 6.41 & 7.21 & 7.02 & 8.20 & 9.89 & \rev{13.53} & \rev{13.54} & \rev{13.40} \\
16 & 7.85 & 8.65 & 6.79 & 6.08 & 9.43 & 7.39 & 6.12 & 7.24 & 6.18 & 6.41 & \rev{7.90} & \rev{7.84} & \rev{7.87} \\
32 & 6.37 & 10.28 & 8.10 & 6.72 & 11.15 & 8.94 & 7.09 & 8.43 & 6.45 & 5.91 & \rev{6.32} & \rev{6.26} & \rev{6.26} \\
\bottomrule
\end{NiceTabular}
\label{tab:fid-cond-maskgit}
}
\end{table}

\begin{table}[htbp]
\centering
{
\mtabcaption{D}{Mean relative difference (\%) of ImgeNet FIDs against reference MaskGIT samplers across 4, 8, 16, 32 steps in class-conditional experiment with MaskGIT-PyTorch.
Mean relative difference was computed in the same way as Table~\ref{tab:error-mage}.
Bolded under $10\%$.
\bt{Moment} consistently approximates \bt{MaskGIT} with the same temperature.
}
\vspace{2mm}
\begin{NiceTabular}
{ccccccccccc}
\toprule
\Block{2-1}{Ref. \bt{MaskGIT} \\ temperature} & \Block{2-1}{\bt{Random}} & \multicolumn{3}{c}{\bt{MaskGIT}} & \multicolumn{3}{c}{\bt{Moment}} & \multicolumn{3}{c}{\bt{Temp}} \\
\cmidrule(lr){3-5} \cmidrule(lr){6-8} \cmidrule(lr){9-11}
 & & $\alpha=3.0$ & 6.0 & 12.0 & 3.0 & 6.0 & 12.0 & 3.0 & 6.0 & 12.0 \\
\midrule
$\alpha=3.0$ & 60.7 & - & 20.4 & 35.2 & \textbf{8.1} & 15.1 & 31.1 & 14.5 & 35.7 & 47.9 \\
$\alpha=6.0$ & 47.5 & 21.3 & - & 17.0 & 29.9 & \textbf{7.1} & 12.9 & \textbf{6.3} & 18.2 & 29.5 \\
$\alpha=12.0$ & 32.4 & 36.1 & 16.4 & - & 43.1 & 23.9 & \textbf{3.9} & 19.7 & \textbf{4.2} & 13.9 \\
\bottomrule
\end{NiceTabular}
\label{tab:error-cond-maskgit}
}
\end{table}


\subsection{Language experiments with different models}\label{sec:language-different-models}
To obtain further empirical evidence of the efficiency of our methods in Section~\ref{sec:method},
we additionally tested them against a larger model \citep[which we refer to as SDTT-large]{deschenaux2024beyond}
and a model with a modified architecture \citep[which we refer to as Di4C]{hayakawa2025distillation}.
Each datapoint is based on 256 generated sequences of 1024 tokens, with 8 to 256 sampling steps.
In each of these two experiments, \bt{Hybrid} used Gumbel coefficient of $\alpha=6.0$ as in the main body.

\rev{We also conduct experiments with the LLaDA-8B base model \citep{nie2025large} to validate our findings in an even larger scale in Section~\ref{sec:llada}.}

\subsubsection{SDTT-large model with Hybrid and Partial caching}
In the experiment with a larger model, we used SDTT-large, a pretrained transformer model scaled to 863M parameters,
in contrast to SDTT with 169M parameters used in Section~\ref{sec:exp-language}.
Since SDTT-large is larger than the GPT-2 large model (774M parameters) in scale,
we used Llama3-8B \citep{dubey2024llama} for computing generative perplexities.

Figure~\ref{fig:large} shows the results for SDTT-large.
We can see a similar trade-off improvement of \bt{Hybrid} over \bt{Random}
and consistent speedups of \bt{Hybrid} and \bt{Random+Cache} over \bt{Random}.
We also plotted a comparison of SDTT-large and SDTT with \bt{Random} and \bt{Hybrid} samplers in Figure~\ref{fig:large-small-sdtt}.
While the model difference becomes dominant as we increase sampling steps,
the use of \bt{Hybrid} is effective for moth models, especially in the few-step regime.


\subsubsection{Hybrid sampler with mixture modeling}
\begin{figure}[htbp]
    \centering
    \includegraphics[width=0.6\linewidth]{figs/ls_perplexity_vs_step_log.pdf}
    \vspace{-4mm}
    \mfigcaption{E}{Efficiency comparison of SDTT-large and SDTT with \bt{Random} and \bt{Hybrid} samplers.
    `Large+' means SDTT-large; otherwise it shows the results of SDTT.
    Generative Perplexity was measured by Llama3-8B for both models.}
    \label{fig:large-small-sdtt}
    \vspace{2mm}
    \subfigure{
        \includegraphics[width=0.475\linewidth]{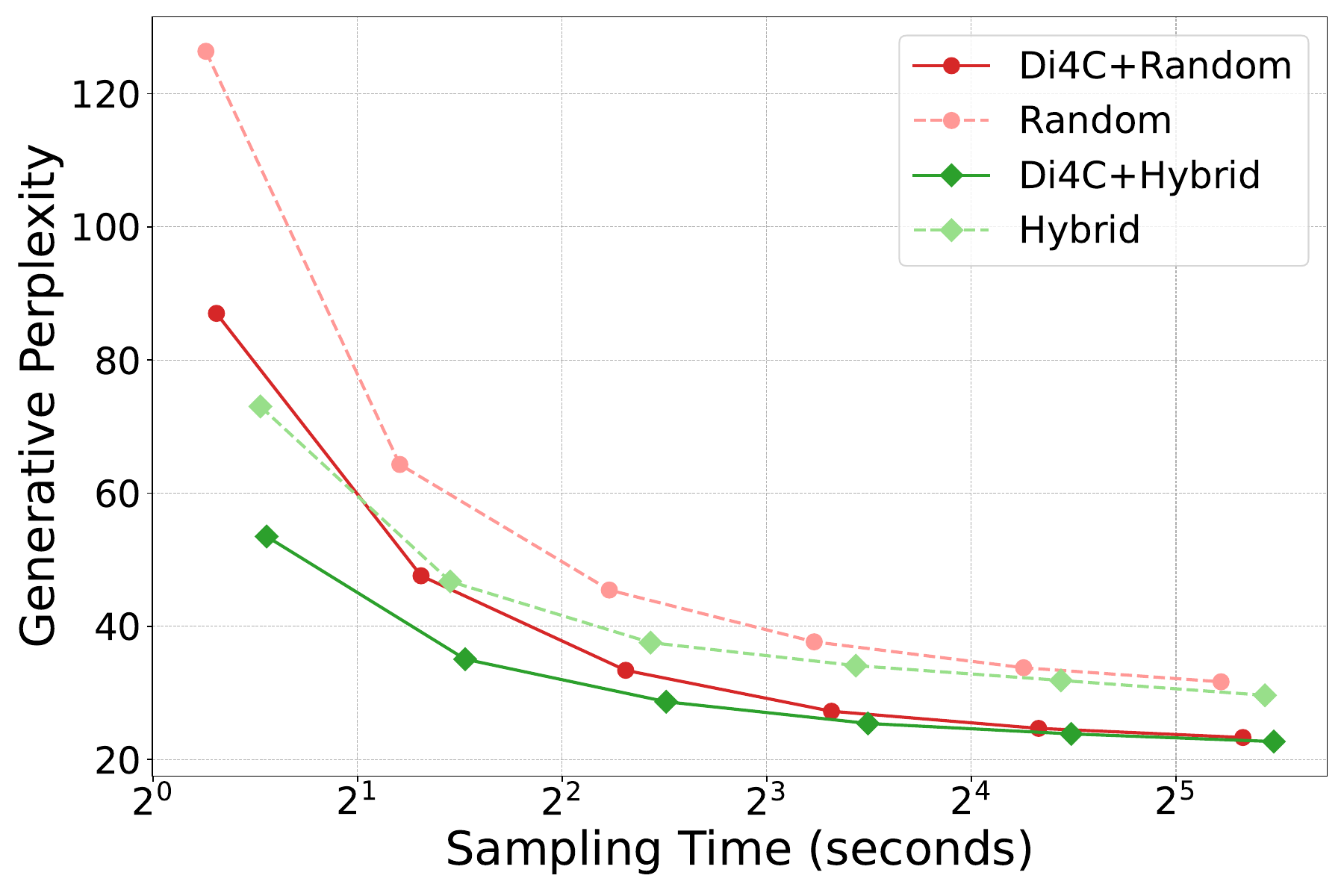}
    }
    \subfigure{
        \includegraphics[width=0.475\linewidth]{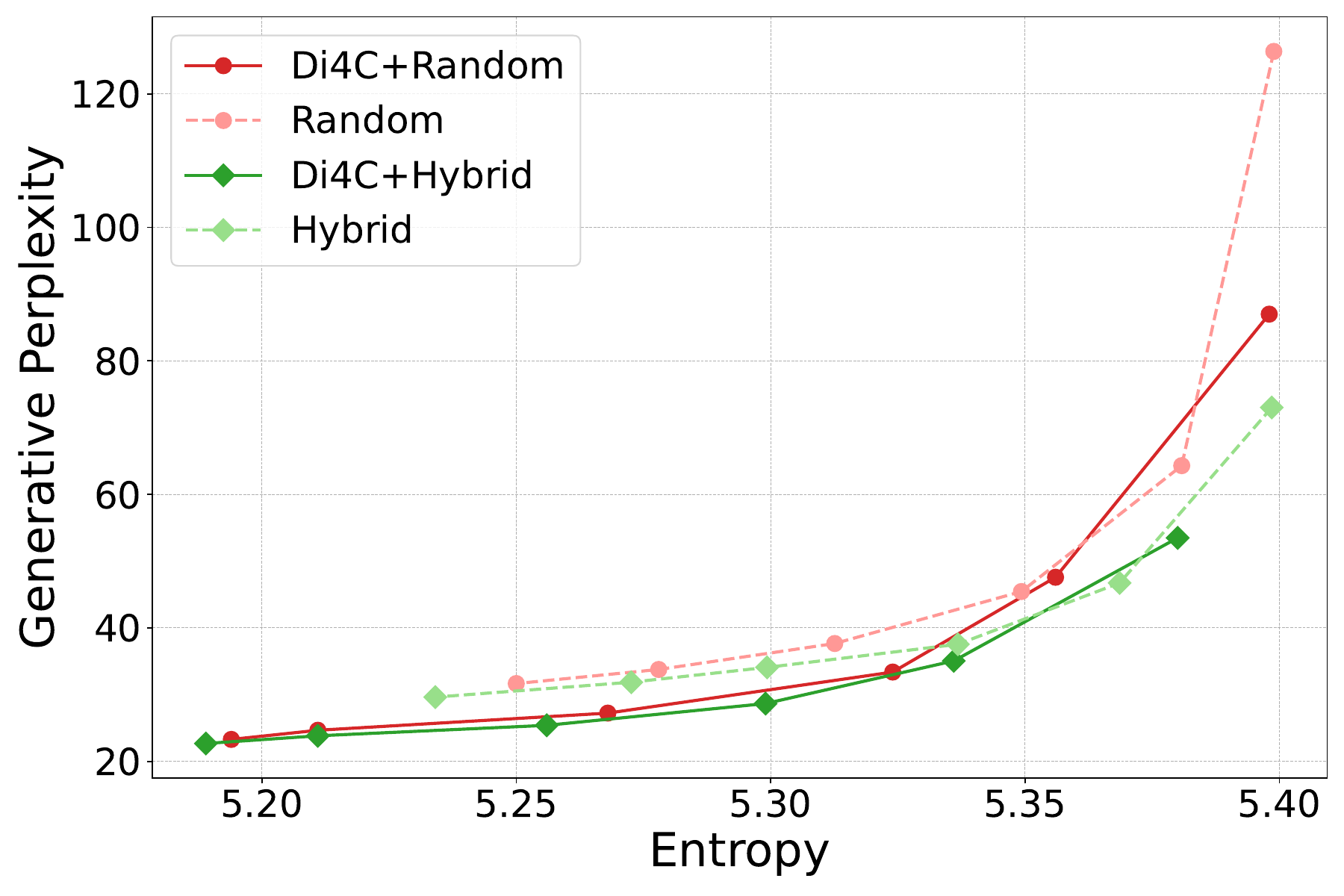}
    }
    \vspace{-4mm}
    \mfigcaption{F}{
        Efficiency comparison of Di4C and SDTT with \bt{Random} and \bt{Hybrid} samplers.
        Plots without `Di4C+' show the results of SDTT.
        Generative Perplexity was measured by GPT-2 large for both models.
    }
    \label{fig:di4c}
    \vspace{2mm}
    \subfigure{
        \includegraphics[width=0.475\linewidth]{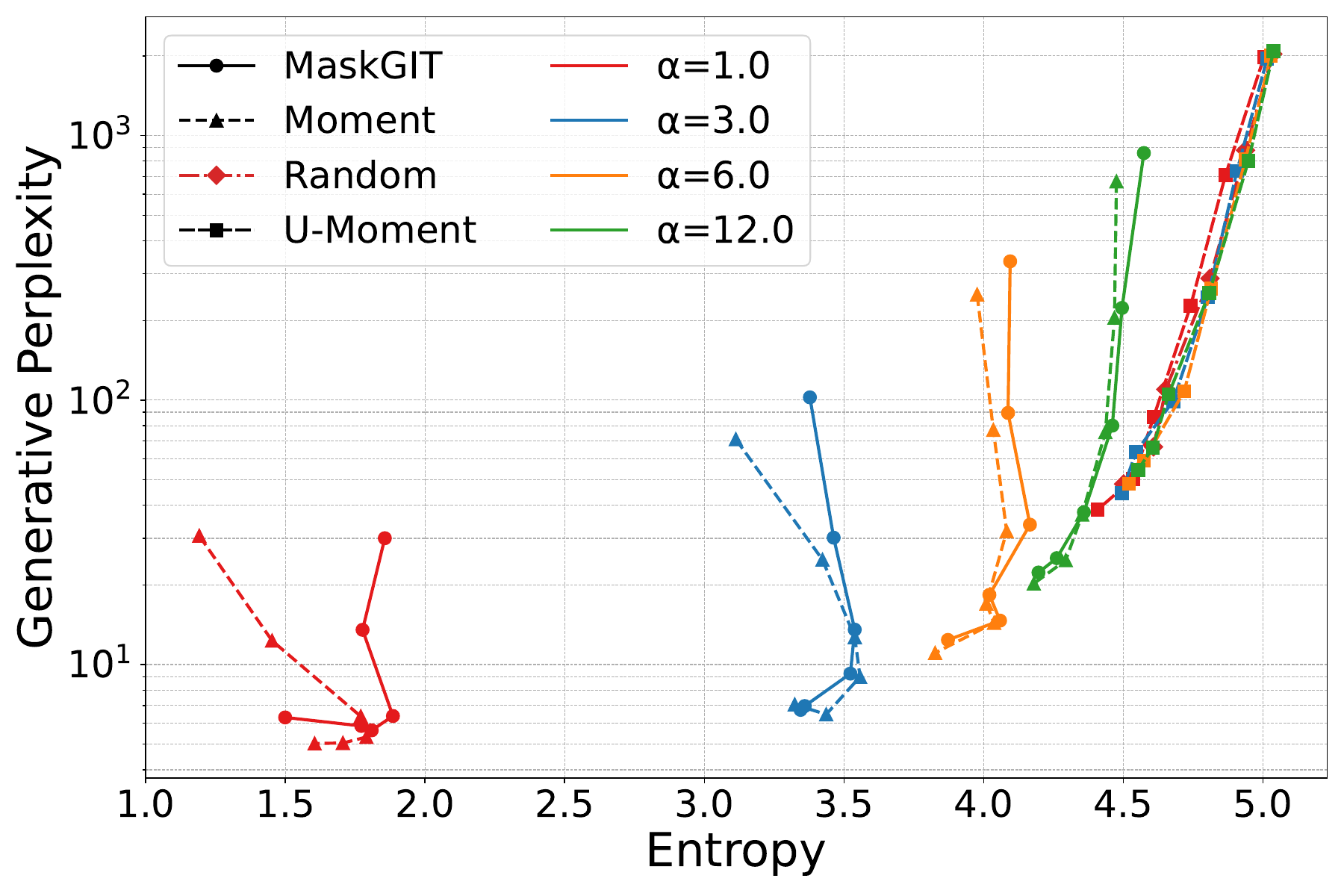}
    }
    \subfigure{
        \includegraphics[width=0.475\linewidth]{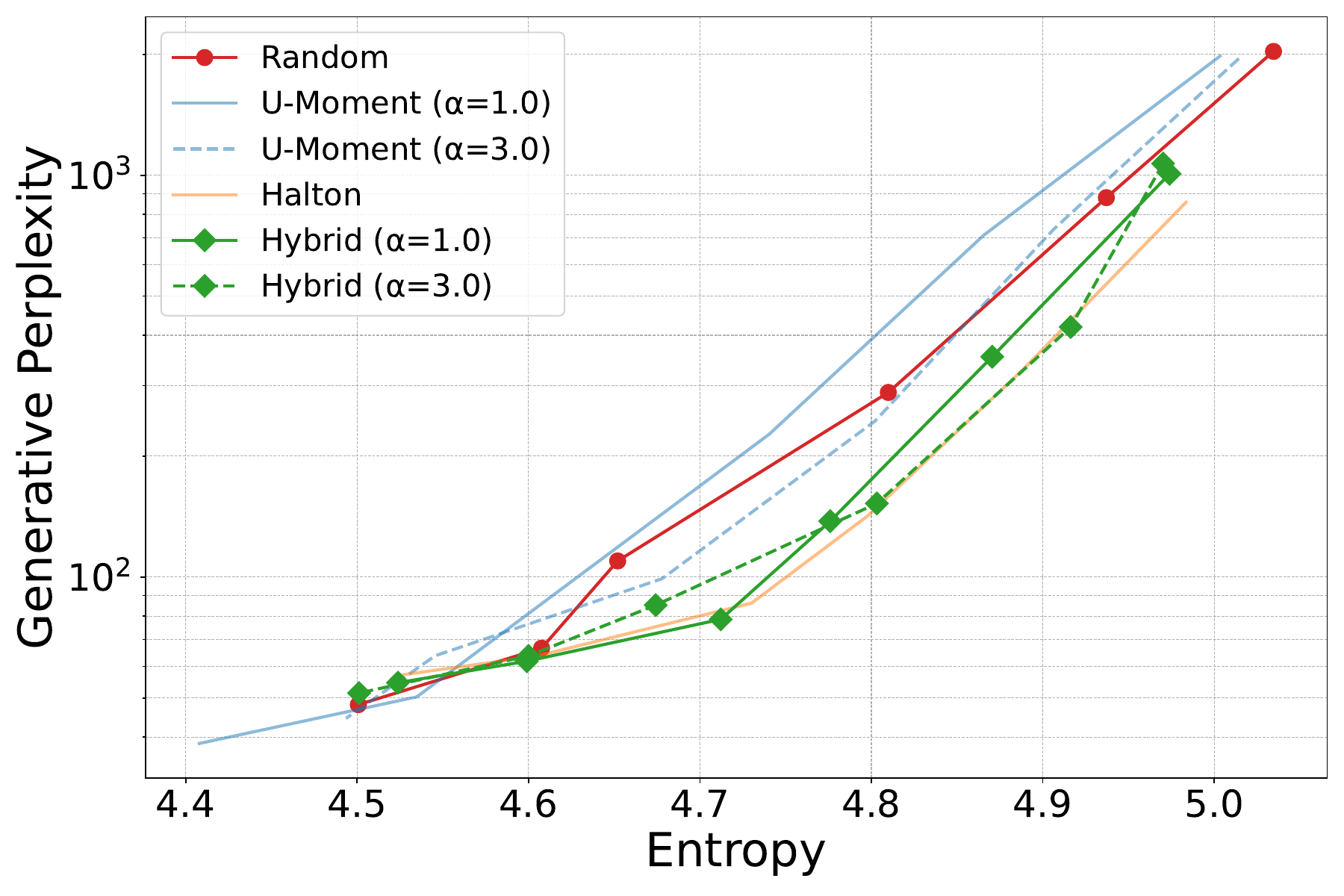}
    }
    \vspace{-4mm}
    \rev{\nfigcaption{A}{Experiments with LLaDA-8B. Each plot was computed by 256 generated sentences with 256 tokens with
    number of steps in $\{8, 16, 32, 64, 128, 256\}$.
    (\textit{Left}) Generative Perplexities ($\uparrow$) against Entropy for temperature-based methods.
    We can again confirm that temperature sampling (\bt{MaskGIT}, \bt{Moment}) significantly harms Entropy, while their unbiased counterpart (\bt{U-Moment} keep a moderate entropy).
    (\textit{Right}) Trade-off between Generative Perplexity and Entropy.
    \bt{Hybrid} samplers show improvement over \bt{Random} similarly to Figure~\ref{fig:sdtt}(Right), while \bt{Halton} is also competitive in this experimental regime.
    }}
    \label{fig:llada}
\end{figure}

We applied \bt{Hybrid} to Di4C, a model obtained from SDTT after two rounds of Di4C-finetuing \citep{hayakawa2025distillation}\footnote{{{\ttfamily sdtt7-di4c2.ckpt} in \url{https://zenodo.org/records/15124163}}}.
This Di4C finetuning includes a slight model architecture modification,
which adds a random latent variable to capture dimensional correlations between different positions/tokens
(caching is not applicable because of this modified modeling).
Figure~\ref{fig:di4c} shows the comparison including SDTT results.
Combining \bt{Hybrid} and the Di4C finetuning (\bt{Di4C+Hybrid}) sums up individual performance gains of \bt{Hybrid} and Di4C
from the original result of \bt{Random}; this is especially remarkable in the few-step regime.

\rev{
\subsubsection{Experiments with an 8B-scale diffusion language model}\label{sec:llada}
In the experiment with LLaDA-8B \citep{nie2025large},
we tested its unconditional generation of 256-token sequences, where we kept the Entropy and Generative Perplexity measured by Llama3-8B \citep{dubey2024llama} as primary evaluation metrics.
While large models like LLaDA are conventionally evaluated by conditional-generation metrics, but those metrics usually just measure the alignment between query and answer, rather than the distributions of output. Our results show that, when low-temperature MaskGIT samplers are applied to LLaDA, sentence entropy in unconditional generation particularly captures the output's diversity loss, which is not measured in conditional tasks.


Figure~\ref{fig:llada} shows the results for LLaDA-8B.
We can see that the MaskGIT's implicit low-temperature sampling can be approximated by the moment sampler and leads to extremetly reduced Entropy (Figure~\ref{fig:llada}, Left). In Figure~\ref{fig:llada}(Right), the trade-off of exploration-exploitation is shifted a bit in favor of exploration, while the Hybrid sampler still interpolates between the two samplers and perform better than the vanilla sampler in terms of both trade-off and efficiency.
}

\subsection{Further ablation of Hybrid modeling}
\subsubsection{Asymmetry of Hybrid modeling}
\bt{Hybrid} is asymmetric regarding the order of two merged methods as detailed in Appendix~\ref{sec:hybrid-details},
where \bt{Halton} determines the indices to unmask and then \bt{U-Moment} determines the rest of unmasking.
Thus, we can also consider \bt{s-Hybrid}, where we just swap the order of two methods in each step,
while the number of indices allocated to each method is unchanged.
We show the results of \bt{s-Hybrid} (256 samples per datapoint)
with SDTT among other samplers in Table~\ref{tab:hybrid-swapped}.
While \bt{s-Hybrid} shows slightly worse performance compared to \bt{Hybrid},
the overall behavior is similar, especially in the few-step regime.
The potential source of the difference is that, since the Halton sequence is deterministic,
its low-discrepancy effect can be reduced when we already select some positions.
To mitigate this effect, adaptive methods such as determinantal point processes \citep{kulesza2012determinantal}
as a replacement for \bt{Halton} can be promising in future work.

\begin{table}[htbp]
\centering
{
    \mtabcaption{E}{Generative Perplexity and Entropy of \bt{s-Hybrid} and other samplers applied to SDTT.}
    \label{tab:hybrid-swapped}
    \vspace{2mm}
    \begin{tabular}{ccccccccc}
    \toprule
    \multirow{2}{*}{Metric} & \multirow{2}{*}{Method} &
    \multicolumn{6}{c}{Steps} \\
    & & 8 & 16 & 32 & 64 & 128 & 256 \\
    \midrule
    \multirow{5}{*}{Gen. PPL}
    & \bt{Random} & 126.36 & 64.31 & 45.45 & 37.67 & 33.78 & 31.67 \\
    & \bt{U-Moment} & 107.04 & 55.36 & 38.95 & 31.96 & 28.40 & 26.62 \\
    & \bt{Hybrid} & 73.02 & 46.73 & 37.55 & 34.09 & 31.87 & 29.64 \\
    & \bt{s-Hybrid} & 73.46 & 48.76 & 40.25 & 35.30 & 31.97 & 30.84 \\
    & \bt{Halton} & 55.09 & 42.53 & 37.41 & 35.76 & 34.32 & 33.13 \\
    \midrule
    \multirow{5}{*}{Entropy}
    & \bt{Random} & 5.399 & 5.381 & 5.349 & 5.313 & 5.278 & 5.250 \\
    & \bt{U-Moment} & 5.354 & 5.333 & 5.298 & 5.263 & 5.228 & 5.199 \\
    & \bt{Hybrid} & 5.399 & 5.369 & 5.337 & 5.299 & 5.273 & 5.234 \\
    & \bt{s-Hybrid} & 5.382 & 5.344 & 5.318 & 5.278 & 5.229 & 5.198 \\
    & \bt{Halton} & 5.426 & 5.385 & 5.351 & 5.320 & 5.293 & 5.263 \\
    \bottomrule
    \end{tabular}
}
\end{table}

\subsubsection{\rev{Comparison of various schedules in Hybrid modeling}}
\begin{figure}[htbp]
    {
    \centering
    \includegraphics[width=0.6\linewidth]{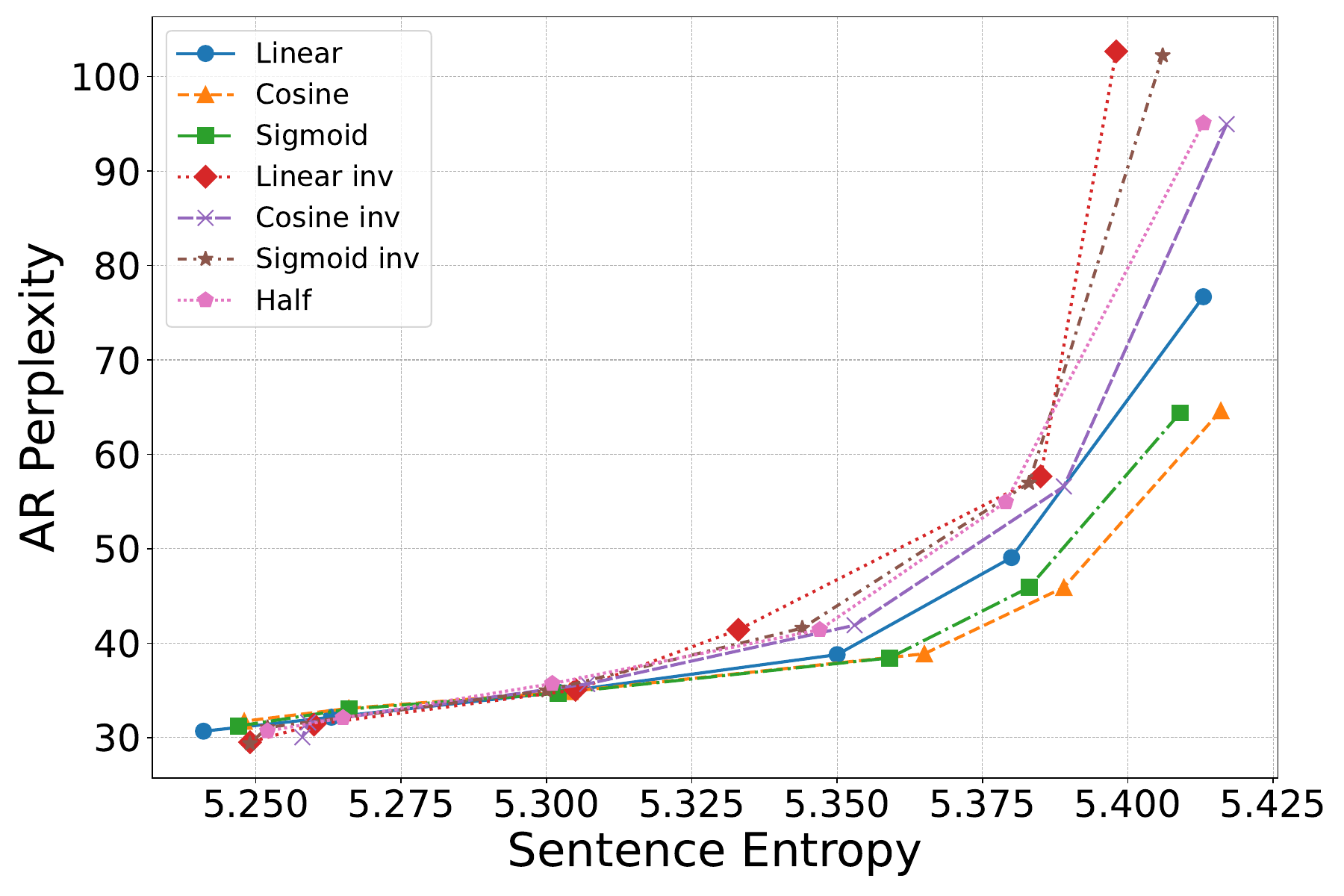}
    \vspace{-4mm}
    \rev{\nfigcaption{B}{
        Comparison of various merging schedulers applied to \bt{Hybrid} sampler of the SDTT model.
    }}
    \label{fig:hybrid-schedule-comparison}
    }
\end{figure}

\rev{
As an ablation study regarding the merging schedule in the Hybrid sampler,
we compared the following schedulers in the SDTT experiment,
where $r_{\text{halton}}$ determines the ratio of indices allocated to \bt{Halton} in the merging rule in Section~\ref{sec:hybrid-details}
in the $n$-th sampling step:
\begin{itemize}
\begin{minipage}[t]{0.48\linewidth}
    \item \bt{Linear}: $r_{\text{halton}} = 1 - \frac{n}{N}$.
    \item \bt{Cosine}: $r_{\text{halton}} = \cos\left(\frac{\pi n}{2N}\right)$.
    \item \bt{Sigmoid}: $r_{\text{halton}} = 1 - \sigma\left(12\left(\frac{n}{N} - \frac{1}{2}\right)\right)$.
    \item \bt{Half}: $r_{\text{halton}} = \frac{1}{2}$.
\end{minipage}
\hfill
\begin{minipage}[t]{0.48\linewidth}
    \item \bt{Linear inv}: $r_{\text{halton}} = \frac{n}{N}$.
    \item \bt{Cosine inv}: $r_{\text{halton}} = \sin\left(\frac{\pi n}{2N}\right)$.
    \item \bt{Sigmoid inv}: $r_{\text{halton}} = \sigma\left(12\left(\frac{n}{N} - \frac{1}{2}\right)\right)$.
\end{minipage}
\end{itemize}
Here, $\sigma = \frac1{1+e^{-x}}$ is the sigmoid function.
The actual number of indices for \bt{Halton} in the experiment is computed as
$m_{\text{halton}} = \mathop\mathrm{round}(\frac{D}{N}r_{\text{halton}})$.
Figure~\ref{fig:hybrid-schedule-comparison} shows the results, where each datapoint was computed using 256 samples.
We can see that our default scheduler (\bt{Linear}) is overall a reasonable choice,
but the optimal merging strategy appears to highly depend on the number of steps, and so ultimately we should adaptively change the merging schedule depending on the total sampling steps or equivalently the budget for generating samples.
}

\subsection{Generated samples}
\subsubsection{MAGE with {MaskGIT} and {Moment} samplers}
\begin{figure}
    \centering
{
    \subfigure[\bt{MaskGIT}, $\alpha=6.0$, 16 steps (FID: 11.64).]{
        \includegraphics[width=0.45\textwidth]{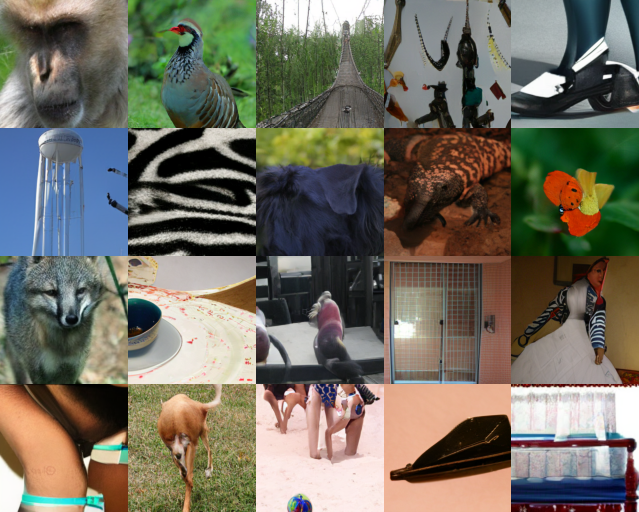}
    }
    \subfigure[\bt{Moment}, $\alpha=6.0$, 16 steps (FID: 11.69).]{
        \includegraphics[width=0.45\textwidth]{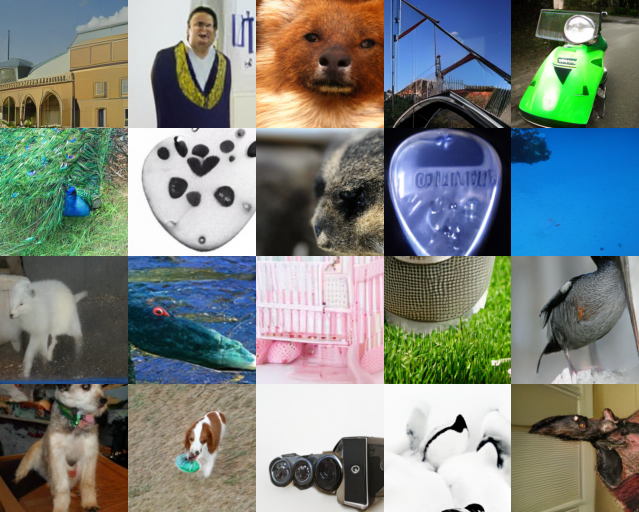}
    }
    \subfigure[\bt{MaskGIT}, $\alpha=12.0$, 64 steps (FID: 9.12).]{
        \includegraphics[width=0.45\textwidth]{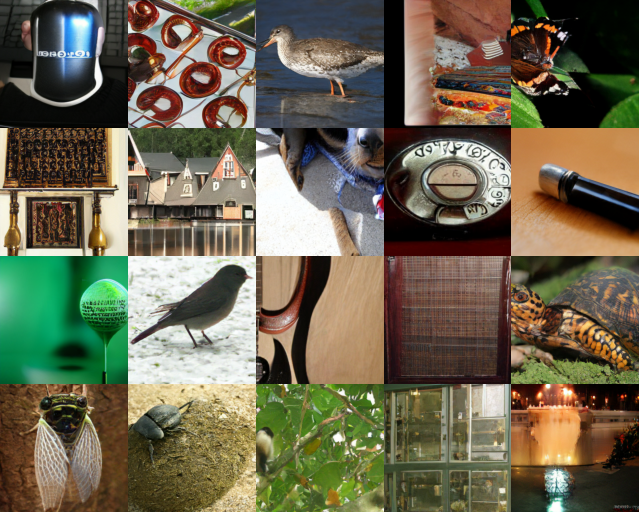}
    }
    \subfigure[\bt{Moment}, $\alpha=12.0$, 64 steps (FID: 9.13).]{
        \includegraphics[width=0.45\textwidth]{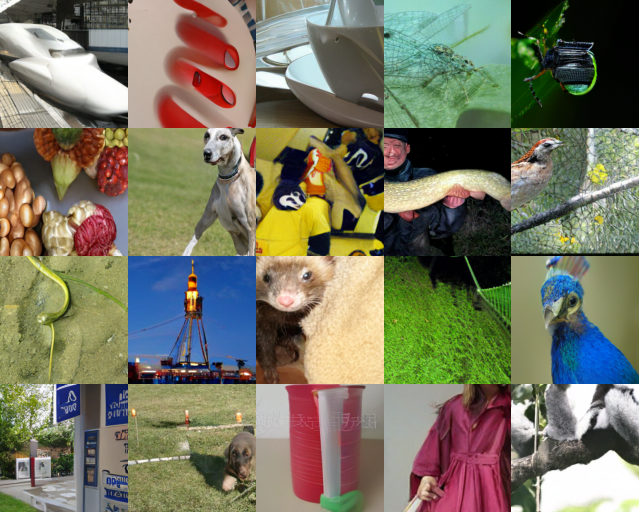}
    }
    \vspace{-2mm}
    \mfigcaption{G}{Comparison of generated samples by MAGE.}
    \label{fig:mage-samples}
}
\end{figure}

We plot the generated samples of MAGE using these samplers with 16 steps and 64 steps in Figure~\ref{fig:mage-samples}.
Note that, although they are generated from the same seed, samples do not correspond to each other between
\bt{MaskGIT} and \bt{Moment} because of the different uses of random numbers as in Figures~\ref{fig:algo-maskgit} and~\ref{fig:algo-moment}.

\subsubsection{SDTT with {MaskGIT} and {Hybrid} samplers}\label{sec:samples-sdtt}
To see the actual examples of entropy reduction and its mitigation by our methods,
we put the 8-step generation results of \bt{MaskGIT} and \bt{Hybrid} with $\alpha=3.0, 6.0$.
We can see the repetitive use of the same characters in \bt{MaskGIT} samplers, which harms the Entropy,
while \bt{Hybrid} results look more natural with diverse use of words, even with higher (worse) Generative Perplexities.

\rev{
\subsubsection{LLaDA-8B with MaskGIT and Hybrid samplers}
Similarly to the previous section, we put 256-token unconditional generation results of LLaDA-8B in Figures~\ref{fig:llada-maskgit}~\&~\ref{fig:llada-hybrid}.
We can observe that the implicit low-entropy sampling reduces the diversity in the \bt{MaskGIT} sampler, while our \bt{Hybrid} counterparts does not suffer from it.
}

\begin{figure}[htbp]
\centering
\begin{tcolorbox}[width=\textwidth, colback=white, colframe=black]
    	\eot iz and for free speech. And to I. however, the other night, and to ‘indicate’ his comments, he argued that I. of the.e. did not say, and to prove to the woman he did not speak the following, and that it is not true.
	
	I. he said that, that and that. and that I. and I.” and, that “I have. and, I. and, in the case of I. and. and. and her,.. and and. and her,,” and he reports that the public has been framed in a manner that is reported to him.
	
I. he connected I. to the communication of the.e to the.e, and to the case of the.e. and. with the case of I. in the. and. I. and I. II, and, I., and,, that was the case of the.e. and, did not decide, I. and. and I. and. and. in the case of I. in the. and. I.” he about a, that I. I, and. I, and I., and I
\end{tcolorbox}
\vspace{-2mm}
{\mfigcaption{H}{Generated by SDTT with \bt{MaskGIT}, $\alpha=3.0$ (Gen.~PPL: 13.88, Entropy: 3.320). }}
\label{fig:maskgit-3}
\end{figure}

\begin{figure}[htbp]
\centering
\begin{tcolorbox}[width=\textwidth, colback=white, colframe=black]
\eot\ and get to show off your help. Don’t worry about that side of the screen, the beta version gets released on June 5, 2015.

[U+2620]:

What’s available here is a free download set up for a variety of machines, naturally; specifically to vDB, MooseX, and the UE6 Server. It is prided itself on the recent release of a new GUI server, and the game is well over there, as there are yet to play too.

Unofficial demos, which focuses on the development, does not release on CODAs as Saturday 3, 2015.

The the current release is the server release. The New Descripten was created late and released in 2012, and several developers have since been involved, working on various different games over a year or so.

In addition, the community has already begun out beta testing on the server and client in an attempt to polish them and ability to show them through the client publicly. Once that can play out, potential releases kaizen out this summer, and beyond this fall, it is enough time for the human capital of release to remain in place. To ensure the delay, however, a group of senior officials, Republicans, and
\end{tcolorbox}
\vspace{-2mm}
{\mfigcaption{I}{Generated by SDTT with \bt{Hybrid}, $\alpha=3.0$ (Gen.~PPL: 67.98, Entropy: 5.360). [U+2620] is actually a single Unicode character of crossbones.}}
\label{fig:hybrid-3}
\end{figure}

\begin{figure}[htbp]
\centering
\begin{tcolorbox}[width=\textwidth, colback=white, colframe=black]
	\eot\ and unworthy of free speech. According to Judge Sitton, the FBI, had listened to respond to said individuals’ their comments, and argued against it. But the judge,..t., refused to respond to the motion and did not respond, on the grounds that it is not confidentiality.
	
	“The FBI does not, and doesn.. have a right stopping this,” he wrote. “This matter, therefore, can’t be mentioned in the reported briefs.”
	
	However,..t. wrote that “there are a number of reports that appear to have been contained in the case that turns up. .
	
	.
	
	So, perhaps considered the possibility of taking action on. or on that, and reviewing it.
	
While this was in time with the.A. and.A. it occurred to me that Mr.. A. v. A. is an important part of the history of the FBI and, until now, not very much known. I had talked to Mr. Sullivan, and in a conversation with “. Barry, the AFR,” he about a year ago, indicated today, and today is about 10:30 to 11:
\end{tcolorbox}
\vspace{-2mm}
{\mfigcaption{J}{Generated by SDTT with \bt{MaskGIT}, $\alpha=6.0$ (Gen.~PPL: 31.89, Entropy: 4.465).}}
\label{fig:maskgit-6}
\end{figure}

\begin{figure}[htbp]
\centering
\begin{tcolorbox}[width=\textwidth, colback=white, colframe=black]
\eot\ and get a show of your game. Don’t worry about right side of the screen, the physical version gets certified on April 5, 2015.

Now read:

We’ve read here about a game download set to ship a perfect three months, offering various changes to game quality in Moosehead, by the time PMabouts. Microsoft has Guided on the most recent release of the new development game, and that product is well too available, as we are excited to test some.

The gameplay demos will be shown on Steam, which does not release on CMA1 as Saturday, October 12.

The the ship will be the first release of The Invaders, Raiden, the Ocean and Star Fox games, and will ship on the same date, but on a different console about a year after arrival.

So far, six games have already been confirmed for testing on this ship, and we will attempt to get begin to publicly ship with them during the summer (because the year can be a difficult time to try them out). Please thank you for your game, as soon as it arrives in our capital, bear in mind each other. To get the latest information and set your sights on at least, you can find
\end{tcolorbox}
\vspace{-2mm}
{\mfigcaption{K}{Generated by SDTT with \bt{Hybrid}, $\alpha=6.0$ (Gen.~PPL: 73.02, Entropy: 5.399).}}
\label{fig:hybrid-6}
\end{figure}

\begin{figure}[htbp]
\centering
\begin{tcolorbox}[width=\textwidth, colback=white, colframe=black]
\underline{\textbf{32 steps:}} 0000 and 1000000

The the values, all the value differences closest to 10000.

Here is a breakdown of the solution steps:

1. Calculate the absolute values of x and 10000:
   a. For 100000 < x < 10000: 100000 - 10000 < x < 100000 = 10000
   b. For 10000 < x < 1000000: 100000 - 10000 < x < 100000 = 10000

2. Calculate the differences:
   a. For 10000 < 1000000: 10000 = 10000
   b. For 10000 < 1000000: 10000 - 2000 = 2000

3. Determine the final

\underline{\textbf{256 steps}}:
00000000000000000000000000000000000000000\\
0111111111111111111000000000000000 0000000000000000000000000000000000000\\
0111111111111111111000000000000000 0000000000000000000000000000000000000\\
00000000000000000000000000000000000000000 00000000000000000000000000
\end{tcolorbox}
\vspace{-2mm}
\rev{\nfigcaption{C}{Generated by LLaDA-8B with \bt{MaskGIT}, $\alpha=1.0$, with 32-step and 256-step sampling.}}
\label{fig:llada-maskgit}
\end{figure}

\begin{figure}[htbp]
\centering
\begin{tcolorbox}[width=\textwidth, colback=white, colframe=black]
\underline{\textbf{32 steps:}}\small
\begin{verbatim}
                
            //---------------------------------------
                            // retreive information
            //---------------------------------------
            SaveString the  string  myQuery, ' ' ;
            SaveString strDir , null;
    
                //--------------------------------------------
            //  Fetch the ODBCQL sub request
            //--------------------------------------
            String myQuery;
            String QueryList;
            InitQueryCaret();

            //---------- 1. Check  Initial  Directory --------------------------
     // <-> check the root directory exists for time profile ATUMS <<<-------------- */
   
            if ( CheckQuery(hroot.myDir, myQuery) )
        return();   //Query found 
        ////----------------------------------------------------------------//
/*  //-----  ------------------

            if ( CheckQuery(strDir,myDir, myQuery.) ):
            .
                //refresh
            writer.write('Treeftp[3 characters] ', strDir  , ', '   '>' .myQuery. '<\n');
            intNextDir := false;         
                //--------------------- Set procedure ---------------------

               
    
\end{verbatim}
\normalsize
\underline{\textbf{256 steps:}}
adiham in rhythm between 1891 and 1899.[1] Although the Allitt building was demolished in the 1970s, a reproduction of its function as The Southern Cross appears in Four Winds \& Compass, the museum in Fylde Way as at 2013.[1] As of 2013, the widow recovers [21][1][1] a replica first mate's uniform, worn by the 19th sourdough captain of the Southwind,[1] on display at Fylde Deco Hotel.[1]  

Arts \\
--------------

In 1898, the year of the appearance of the widow, the Needlecraft Hall was just built; now - some believe - was one of the first bonds between classes represented, said to have been formed.[1]  

Marriage \\
--------------------

Wilson at work, at the G\_gensoe textile factory in 1902

The wife initially apports, from the reproduction of seahorses shot, in order to buy his own smithy.[1] Wilson, who had auctioned a number of goods based on the experience of Dutchmen, had his erstwhile
\end{tcolorbox}
\vspace{-2mm}
\rev{\nfigcaption{D}{Generated by LLaDA-8B with \bt{Hybrid}, $\alpha=1.0$, with 32-step and 256-step sampling. In the result of 32-step generation, we used the \texttt{verbatim} package of \LaTeX\ to properly display the spacings. \texttt{[3~characters]} are the Unicode characters [U+6389][U+5802][U+5802].}}
\label{fig:llada-hybrid}
\end{figure}

\end{document}